%% file: RLPaper.tex
\documentclass[final,3p,times,twocolumn]{elsarticle}  

\usepackage{amssymb}
\usepackage[english]{babel}
\usepackage{blindtext}
\usepackage{booktabs}
\usepackage{url}
\usepackage{hyperref}
\hypersetup{pdfborder={0 0 0}}
\usepackage{amsthm}
\usepackage{amsmath}
\usepackage{tikz}						
\usepackage{circuitikz}
\usetikzlibrary{trees}
\usetikzlibrary{calc}				
\usetikzlibrary{plotmarks}
\usepgflibrary{arrows}
\usetikzlibrary{patterns}
\usetikzlibrary{positioning}
\usetikzlibrary{shapes.arrows}
\usetikzlibrary{positioning,calc,shapes,arrows,shapes.multipart}
\tikzset{
	papDecision/.style = {
		diamond,
		draw, 
		text width = 18 mm, 
		align = center, 
		text badly centered,
		inner sep = 1 pt,
		font=\footnotesize,
		minimum width = 25mm,
		minimum height = 7mm,
	},
	papStart/.style = {
		rectangle,
		draw, 
		align = center, 
		text width = 3.3cm, 
		text badly centered,
		inner sep = 4 pt,
		rounded corners=10pt,
		font=\footnotesize,
		minimum width = 30mm,
		minimum height = 7mm,
	},
	papEnd/.style = {
		rectangle,
		draw, 
		align = center, 
		text width = 3.3cm, 
		text badly centered,
		inner sep = 4 pt,
		rounded corners=10pt,
		font=\footnotesize,
		minimum width = 30mm,
		minimum height = 7mm,
	},
	papData/.style = {
		trapezium,
		draw, 
		align = center, 
		text width = 20 mm, 
		text badly centered,
		inner sep = 4 pt,
		trapezium left angle=70,
		trapezium right angle=110,
		font=\footnotesize,
		minimum width = 30mm,
		minimum height = 7mm,
	},
	papPredProc/.style = {
		draw,
		rectangle split,
		rectangle split horizontal,
		rectangle split parts = 3,
		rectangle split empty part width=-8pt,
		align = center, 
		text badly centered,
		font=\footnotesize,
		minimum width = 30mm,
		minimum height = 7mm,
	},
	papProcess/.style = {
		rectangle,
		draw,
		align = center, 
		text width = 3.3cm, 
		text badly centered,
		font=\footnotesize,
		minimum width = 30mm,
		minimum height = 7mm,
	},
	papLine/.style = {
		draw,
		-stealth,
		font=\footnotesize,
	},
}

\usepackage{pgfplots}
\pgfplotsset{compat=1.14}
\usepgfplotslibrary{fillbetween}

\definecolor{mmpBlack}{rgb}{0, 0, 0}
\definecolor{mmpDarkBlue}{rgb}{0, 0.3294, 0.6235}
\definecolor{mmpMiddleBlue}{rgb}{0.251, 0.498, 0.7176}
\definecolor{mmpLightBlue}{rgb}{0.5569, 0.7294, 0.898}
\definecolor{mmpRed}{rgb}{0.8, 0.0275, 0.1176}
\definecolor{mmpGreen}{rgb}{0.3412, 0.6706, 0.1529}
\definecolor{mmpOrange}{rgb}{0.9647, 0.6588, 0}
\definecolor{mmpDarkGray}{rgb}{0.75, 0.75, 0.75}
\definecolor{mmpMiddleDarkGray}{rgb}{0.5, 0.5, 0.5}
\definecolor{mmpMiddleGray}{rgb}{0.25, 0.25, 0.25}
\definecolor{mmpLightGray}{rgb}{0.1, 0.1, 0.1}

\usepackage{glossaries}
\input{Layout/Lay_Glossaries}
\input{Layout/symbols}
\input{Layout/Abbreviations}

\newlength{\PlotWidth}
\newlength{\PlotHeight}	
\newlength{\DeltaHeight}
\newlength{\DeltaHeightDashed}

\newlength{\DeltaWidth}
\newlength{\Ypos}	
\newlength{\Xpos}	
\newlength{\XEndPos}

\newlength{\YposTop}
\newlength{\YposButtom}

\newcommand{\mySymbol}[2]{#1\textsubscript{#2}}

\newcommand{\pzyl}{\mySymbol{\gls{rp}}{\gls{iZyl}}}

\newcommand{\rkl}{\mySymbol{\gls{rr}}{\gls{iKlasse},\gls{iRichtung}}}

\usepackage[locale = DE]{siunitx}		
\DeclareSIUnit\bar{bar}
\newcommand{\degreeKW}{\ensuremath{^\circ\text{CA}}}
\newcommand{\degreeKWnZOT}{\ensuremath{^\circ\text{CAaTDC}}}
\DeclareSIUnit{\degreeCrankAngle}{\degreeKW} 
\DeclareSIUnit{\degreeCrankAngleaTDC}{\degreeKWnZOT}
\DeclareSIUnit{\barperdegree}{\bar\per\degreeKW}

\usepackage{natbib}

\begin{document}

\begin{frontmatter}



\title{Safe Reinforcement Learning for Real-World Engine Control}

 \affiliation[label1]{organization={Teaching and Research Area Mechatronics in Mobile Propulsion},
            addressline={RWTH Aachen University},             city={Aachen},
            country={Germany}}


\author[label1]{Julian Bedei}
\author[label1]{Lucas Koch}
\author[label1]{Kevin Badalian}
\author[label1]{Alexander Winkler}
\author[label1]{Patrick Schaber}
\author[label1]{Jakob Andert}

\begin{abstract}
\input{Chapters/Abstract.tex}

\end{abstract}




\end{frontmatter}


\input{Chapters/IntroductionNewFocusRL}

\input{Chapters/RLFundamentals}
\input{Chapters/ExperimentalSetup}

\input{Chapters/Fundamentals}

\input{Chapters/Validation.tex}

\input{Chapters/Conclusion.tex}

\section*{Acknowledgements}
This research was performed as part of the research unit 2401 (FOR2401) “Optimization based Multiscale Control for Low Temperature Combustion Engines” funded by the Deutsche Forschungsgemeinschaft (DFG, German Research Foundation) - 277012063. This support is gratefully acknowledged.

\section*{Data Availability}
The data and scripts supporting this study are openly available in Zenodo at https://doi.org/10.5281/zenodo.14499423





\bibliographystyle{elsarticle-harv} 

\bibliography{Bibtex/RLPaper}





\end{document}

%% file: Layout/Lay_Glossaries.tex
\newglossary[slg]{symver}{syi}{syg}{Symbolverzeichnis}

\newglossarystyle{symstyle}{%
	\renewenvironment{theglossary}%
	{\begin{longtable}{@{}p{.25\textwidth}@{}p{.75\textwidth}@{}}
			\endfirsthead
			\endhead
			\endfoot
			\endlastfoot}{\end{longtable}}%
	\renewcommand*{\glsgroupheading}[1]{\ifstrequal{A}{##1}{\relax}{\rlap{\large\textbf{\glsgetgrouptitle{##1}}} \\[-8pt]}}
	}

\newglossary[alg]{abkver}{acr}{acn}{Abkürzungsverzeichnis}

\newglossarystyle{abkstyle}{%
	{\begin{longtable}{@{}p{.25\textwidth}@{}p{.75\textwidth}@{}}
			\endfirsthead
			\endhead
			\endfoot
			\endlastfoot}{\end{longtable}}%
	\renewcommand*{\glsgroupheading}[1]{\ifstrequal{A}{##1}{\relax}{\rlap{\large\textbf{\glsgetgrouptitle{##1}}} \\[-8pt]}}
	}

%% file: Layout/symbols.tex
\newglossaryentry{aNabla}{type=symver,name={\ensuremath{\nabla}},description={Nabla Operator},sort={aNabla}}

\newglossaryentry{rAK}{type=symver,name={\ensuremath{A_\text{K}}},description={Kolbenquerschnittsfläche},sort={iAK}}
\newglossaryentry{rAW}{type=symver,name={\ensuremath{A_\text{W}}},description={Für den Wandwämeübergang wirksame Wandfläche},sort={iAW}}
\newglossaryentry{roNeuron}{type=symver,name={\ensuremath{a}},description={Aktivierungszustand eines künstlichen Neurons},sort={iaNeuron}}

\newglossaryentry{rb}{type=symver,name={\ensuremath{b}},description={Bias-Wert},sort={ib}}

\newglossaryentry{rcv}{type=symver,name={\ensuremath{c_\text{v}}},description={Isochore spezifische Wärmekapazität},sort={icv}}
\newglossaryentry{rCK}{type=symver,name={\ensuremath{K}},description={Menge aller Klassen},sort={iKKlasse}}
\newglossaryentry{rCapacity}{type=symver,name={\ensuremath{C}},description={Elektrische Kapazität},sort={iCapacity}}

\newglossaryentry{rd}{type=symver,name={\ensuremath{d}},description={Abstand},sort={id}}
\newglossaryentry{rdRL}{type=symver,name={\ensuremath{d}},description={Boolescher Indikator für das Ende einer Episode},sort={idaRL}}

\newglossaryentry{rDPMAX}{type=symver,name={\ensuremath{\mathrm{d}\gls{rp}_{\gls{imax}}}},description={Maximaler Druckgradient},sort={iDPMAX}}

\newglossaryentry{rDPMAXi}{type=symver,name={\ensuremath{\mathrm{d}\gls{rp}_{\gls{imax},i}}},description={Maximaler Druckgradient},sort={iDPMAXi}}
\newglossaryentry{rDPMAXi1}{type=symver,name={\ensuremath{\mathrm{d}\gls{rp}_{\gls{imax},i-1}}},description={Maximaler Druckgradient},sort={iDPMAXi1}}
\newglossaryentry{rDPMAXLim}{type=symver,name={\ensuremath{\mathrm{d}\gls{rp}_{\gls{imax},\gls{iLim}}}},description={Maximaler Druckgradient},sort={iDPMAX}}

\newglossaryentry{rDUIONMAX}{type=symver,name={\ensuremath{\mathrm{d}\gls{rUIon}_{,\gls{imax}}}},description={Maximaler Gradient des Ionenstromsignals},sort={iDUIONMAX}}
\newglossaryentry{rEEnergie}{type=symver,name={\ensuremath{E}},description={Energie},sort={iEAEnergie}}
\newglossaryentry{rE}{type=symver,name={\ensuremath{E}},description={Fehler},sort={iE}}
\newglossaryentry{rErr}{type=symver,name={\ensuremath{\mathcal{L}}},description={Fehlerfunktion},sort={iLErr}}
\newglossaryentry{rERL}{type=symver,name={\ensuremath{\mathcal{E}}},description={Episode},sort={iERL}}

\newglossaryentry{rG}{type=symver,name={\ensuremath{G}},description={Ertrag (engl.: Return)},sort={iG}}

\newglossaryentry{rHU}{type=symver,name={\ensuremath{\text{LCV}}},description={Unterer Heizwert},sort={iHU}}
\newglossaryentry{rHUKr}{type=symver,name={\ensuremath{\text{LCV}_{\gls{iKr}}}},description={Unterer Heizwert Kraftstoff},sort={iHU}}
\newglossaryentry{rHUEth}{type=symver,name={\ensuremath{\text{LCV}_{\gls{iEth}}}},description={Unterer Heizwert Ethanol},sort={iHU}}

\newglossaryentry{rHEin}{type=symver,name={\ensuremath{H_{\text{ein}}}},description={Enthalpie des Eingangsmassenstroms},sort={iHEin}}
\newglossaryentry{rHAus}{type=symver,name={\ensuremath{H_{\text{aus}}}},description={Enthalpie des Ausgangsmassenstroms},sort={iHAus}}
\newglossaryentry{rHv}{type=symver,name={\ensuremath{H_{\text{v}}}},description={Verdampfungsenthalpie},sort={iHv}}

\newglossaryentry{rIIon}{type=symver,name={\ensuremath{i_\text{Ion}}},description={Ionenstrom},sort={iIIon}}

\newglossaryentry{rIonMax}{type=symver,name={\ensuremath{\ensuremath{U_{\text{Ion},\gls{imax}}}}},description={Maximum des Ionenstromsignals},sort={iUIonMax}}
\newglossaryentry{rIonMaxi}{type=symver,name={\ensuremath{\ensuremath{U_{\text{Ion},\gls{imax},i}}}},description={Maximum des Ionenstromsignals},sort={iUIonMax}}
\newglossaryentry{rIonMaxi1}{type=symver,name={\ensuremath{\ensuremath{U_{\text{Ion},\gls{imax},i-1}}}},description={Maximum des Ionenstromsignals},sort={iUIonMax}}
\newglossaryentry{rIonIntegral}{type=symver,name={\ensuremath{I_{U_{\text{Ion}}}}},description={Integral des Ionenstroms},sort={iIonIntegral}}
\newglossaryentry{rIonIntegrali}{type=symver,name={\ensuremath{I_{U_{\text{Ion}},i}}},description={Integral des Ionenstroms},sort={iIonIntegral}i}
\newglossaryentry{rIonIntegrali1}{type=symver,name={\ensuremath{I_{U_{\text{Ion}},i-1}}},description={Integral des Ionenstroms},sort={iIonIntegral}i1}

\newglossaryentry{rI}{type=symver,name={\ensuremath{\ensuremath{I}}},description={Relative Wichtigkeit eines Eingangsmerkmals},sort={iI}}

\newglossaryentry{rKP}{type=symver,name={\ensuremath{K_\text{P}}},description={Proportionalanteil eines Reglers},sort={iKP}}

\newglossaryentry{rm}{type=symver,name={\ensuremath{m}},description={Masse},sort={im}}

\newglossaryentry{rnumber}{type=symver,name={\ensuremath{n}},description={Anzahl},sort={inb}}
\newglossaryentry{rNMot}{type=symver,name={\ensuremath{n_\text{Mot}}},description={Motordrehzahl},sort={inMot}}
\newglossaryentry{rni}{type=symver,name={\ensuremath{n_i}},description={Stoffmenge einer Komponente $i$},sort={ini}}

\newglossaryentry{rNG}{type=symver,name={\ensuremath{\mathcal{N}}},description={Gaußsches Rauschen},sort={iNA}}

\newglossaryentry{rO}{type=symver,name={\ensuremath{\boldsymbol{O}}},description={Orientierungsmatrix des Vermessungsalgorithmus},sort={iO}}
\newglossaryentry{ro}{type=symver,name={\ensuremath{o}},description={Element der Orientierungsmatrix},sort={iO}}

\newglossaryentry{rp}{type=symver,name={\ensuremath{p}},description={Druck},sort={ip}}

\newglossaryentry{rpmi}{type=symver,name={\ensuremath{\text{IMEP}}},description={Indizierter Mitteldruck},sort={ipmi}}
\newglossaryentry{rpmii}
{type=symver,name={\ensuremath{\text{IMEP}_{i}}},description={Indizierter Mitteldruck},sort={ipmii}}
\newglossaryentry{rpmii1}
{type=symver,name={\ensuremath{\text{IMEP}_{i-1}}},description={Indizierter Mitteldruck},sort={ipmii1}}
\newglossaryentry{rpmiSoll}
{type=symver,name={\ensuremath{\text{IMEP}_{\gls{iSoll}}}},description={Indizierter Mitteldruck},sort={ipmiSoll}}
\newglossaryentry{rpmiiSoll}
{type=symver,name={\ensuremath{\text{IMEP}_{\gls{iSoll},i}}},description={Indizierter Mitteldruck},sort={ipmiiSoll}}
\newglossaryentry{rpmii1Soll}
{type=symver,name={\ensuremath{\text{IMEP}_{\gls{iSoll},i-1}}},description={Indizierter Mitteldruck},sort={ipmii1Soll}}

\newglossaryentry{rP}{type=symver,name={\ensuremath{P}},description={Wahrscheinlichkeit},sort={iP}}

\newglossaryentry{rQ}{type=symver,name={\ensuremath{Q}},description={Wärmefreisetzung durch Verbrennung},sort={iQW}}
\newglossaryentry{rQW}{type=symver,name={\ensuremath{Q_\text{W}}},description={Wandwärmeverluste},sort={iQZ}}
\newglossaryentry{rQfunc}{type=symver,name={\ensuremath{Q}},description={Q-Funktion},sort={iQA}}

\newglossaryentry{rRGas}{type=symver,name={\ensuremath{R}},description={Universelle Gaskonstante},sort={iRU}}
\newglossaryentry{rRShunt}{type=symver,name={\ensuremath{R_\text{Shunt}}},description={Einstellbarer Widerstand zur Messung des Ionenstroms},sort={iRE}}
\newglossaryentry{rr}{type=symver,name={\ensuremath{R}},description={Radius des Vermessungsalgorithmus},sort={iRR}}
\newglossaryentry{rRVermessung}{type=symver,name={\ensuremath{R}},description={Positionsmatrix des Vermessungsalgorithmus},sort={iRM}}

\newglossaryentry{rRReward}{type=symver,name={\ensuremath{r}},description={Belohnung (engl.: Reward)},sort={irzb}}
\newglossaryentry{rrDeltaExpl}{type=symver,name={\ensuremath{\Delta r_\text{Expl}}},description={Schrittweite des Vermessungsalgorithmus in radialer Richtung},sort={irzz}}
\newglossaryentry{rrCor}{type=symver,name={\ensuremath{r}},description={Korrelationskoeffizient},sort={irzk}}

\newglossaryentry{rRMSE}{type=symver,name={\text{RMSE}},description={Root Mean Square Error},sort={rRMSE}}

\newglossaryentry{rs}{type=symver,name={\ensuremath{s}},description={Iterationsschritt},sort={is}}

\newglossaryentry{rT}{type=symver,name={\ensuremath{T}},description={Temperatur},sort={iTA}}
\newglossaryentry{rTRL}{type=symver,name={\ensuremath{T}},description={Tupel einer  Erfahrung beim Reinforcement Learning},sort={iTB}}

\newglossaryentry{rt}{type=symver,name={\ensuremath{t}},description={Zeit},sort={itB}}
\newglossaryentry{rTWand}{type=symver,name={\ensuremath{T_\text{W}}},description={Wandtemperatur},sort={iTWand}}

\newglossaryentry{rURaum}{type=symver,name={\ensuremath{\mathcal{A}}},description={Aktionsraum},sort={iU}}

\newglossaryentry{rU}{type=symver,name={\ensuremath{U}},description={Innere Energie},sort={iUA}}

\newglossaryentry{ru}{type=symver,name={\ensuremath{u}},description={spezifische innere Energie},sort={iua}}

\newglossaryentry{ruCtrl}{type=symver,name={\ensuremath{a}},description={Stellgröße},sort={iub}}

\newglossaryentry{ruCtrlVec}{type=symver,name={\ensuremath{\vec{a}}},description={Vektor der Stellgrößen},sort={iuc}}

\newglossaryentry{rUDC}{type=symver,name={\ensuremath{U_\text{dc}}},description={Angelegte Gleichspannung zur Bestimmung des Ionenstroms},sort={iud}}

\newglossaryentry{rUIon}{type=symver,name={\ensuremath{U_\text{Ion}}},description={Ionenstrom gemessen als Spannungsabfall},sort={iue}}

\newglossaryentry{rV}{type=symver,name={\ensuremath{V}},description={Volumen},sort={iVA}}
\newglossaryentry{rVRL}{type=symver,name={\ensuremath{V}},description={Wertefunktion},sort={iVB}}

\newglossaryentry{rRichtungsvektor}{type=symver,name={\ensuremath{\vec{v}}},description={Richtungsvektor des Vermessungsalgorithmus},sort={ivC}}

\newglossaryentry{rVC}{type=symver,name={\ensuremath{V_{\mathrm{C}}}},description={kompressionsvolumen},sort={iVD}}
\newglossaryentry{rVH}{type=symver,name={\ensuremath{V_{\mathrm{H}}}},description={Hubvolumen},sort={iVE}}
\newglossaryentry{rvkm}{type=symver,name={\ensuremath{v_\text{K,m}}},description={Mittlere Kolbengeschwindigkeit},sort={ivF}}

\newglossaryentry{rWK}{type=symver,name={\ensuremath{W_{\mathrm{K}}}},description={Volumenarbeit am Kolben},sort={iWc}}
\newglossaryentry{rW}{type=symver,name={\ensuremath{W}},description={Arbeit},sort={iWA}}
\newglossaryentry{rw}{type=symver,name={\ensuremath{w}},description={Gewichtung},sort={iwb}}

\newglossaryentry{rX}{type=symver,name={\ensuremath{\mathcal{S}}},description={Zustandsraum},sort={iX}}

\newglossaryentry{rxAnteil}{type=symver,name={\ensuremath{x}},description={Anteil},sort={ixA}}

\newglossaryentry{rxState}{type=symver,name={\ensuremath{s}},description={Zustandsgröße},sort={ixb}}

\newglossaryentry{rxStateVec}{type=symver,name={\ensuremath{\vec{s}}},description={Vektor der Zustandsgrößen},sort={ixc}}

\newglossaryentry{rxAbg}{type=symver,name={\ensuremath{x_{\mathrm{abg}}}},description={Abgasrückführrate},sort={ixd}}

\newglossaryentry{rZ}{type=symver,name={\ensuremath{\boldsymbol{Z}}},description={Zählermatrix des Vermessungsalgorithmus},sort={iZA}}

\newglossaryentry{rz}{type=symver,name={\ensuremath{z}},description={Element der Zählermatrix},sort={izb}}

\newglossaryentry{rnet}{type=symver,name={\ensuremath{z}},description={Nettoeingabewert eines künstlichen Neurons},sort={izc}}

\newglossaryentry{gAlpha}{type=symver,name={\ensuremath{\alpha}},description={Kurbelwinkel},sort={ga}}
\newglossaryentry{gAlpha50}{type=symver,name={\ensuremath{\alpha_{50}}},description={Verbrennungsschwerpunktlage},sort={ga50}}
\newglossaryentry{gAlpha50i}{type=symver,name={\ensuremath{\alpha_{50,i}}},description={Verbrennungsschwerpunktlage},sort={ga50i}}
\newglossaryentry{gAlpha50i1}{type=symver,name={\ensuremath{\alpha_{50,i-1}}},description={Verbrennungsschwerpunktlage},sort={ga50i1}}
\newglossaryentry{gAlpha50Soll}{type=symver,name={\ensuremath{\alpha_{50,\gls{iSoll}}}},description={Verbrennungsschwerpunktlage},sort={ga50Soll}}

\newglossaryentry{gAlphaIon50}{type=symver,name={\ensuremath{\alpha_{\text{Ion},50}}},description={Winkel bei dem 50 \% des Ionenstromintegrals erreicht sind},sort={ga50Ion}}
\newglossaryentry{gAlphaK}{type=symver,name={\ensuremath{\alpha_\text{K}}},description={Durchströmzahl},sort={gak}}
\newglossaryentry{gAlphaW}{type=symver,name={\ensuremath{\alpha_\text{W}}},description={Wärmeübergangskoeffizient},sort={gaw}}
\newglossaryentry{ggamma}{type=symver,name={\ensuremath{\gamma}},description={Diskontierungsfaktor},sort={gc}}
\newglossaryentry{gepsilon}{type=symver,name={\ensuremath{\varepsilon}},description={Verdichtungsverhältnis},sort={ge}}
\newglossaryentry{geta}{type=symver,name={\ensuremath{\eta_\text{i}}},description={Indizierter Wirkungsgrad},sort={gg}}
\newglossaryentry{gtheta}{type=symver,name={\ensuremath{\theta}},description={Schwellenwert einer Aktivierungsfunktion},sort={ghSchwellenwert}}
\newglossaryentry{gthetaParam}{type=symver,name={\ensuremath{\theta}},description={Parameter des Parametersatzes \gls{gTheta}},sort={ghParam}}
\newglossaryentry{gTheta}{type=symver,name={\ensuremath{\Theta}},description={Parametersatz eines neuronalen Netzes},sort={gH}}
\newglossaryentry{gKappa}{type=symver,name={\ensuremath{\kappa}},description={Isentropenexponent},sort={gj}}
\newglossaryentry{gLambda}{type=symver,name={\ensuremath{\lambda_{\text{L}}}},description={Kraftstoff-Luft-Verhältnis},sort={gkbLambda}}
\newglossaryentry{gLambdaRL}{type=symver,name={\ensuremath{\lambda}},description={Abklingfaktor},sort={gkaLambdaRL}}

\newglossaryentry{gm}{type=symver,name={\ensuremath{\mu}},description={Mittelwert einer Verteilung},sort={glaM}}
\newglossaryentry{gmRL}{type=symver,name={\ensuremath{\mu}},description={Deterministische Strategie eines Reinforcement Learning-Agenten},sort={glaD}}
\newglossaryentry{gmi}{type=symver,name={\ensuremath{\mu_i}},description={Chemisches Potenzial einer Komponente $i$},sort={glb}}

\newglossaryentry{grho}{type=symver,name={\ensuremath{\rho}},description={Dichte},sort={gq}}
\newglossaryentry{grhoPoly}{type=symver,name={\ensuremath{\rho}},description={Faktor für die Mittelung nach Polyak},sort={gqPoly}}
\newglossaryentry{gsigma}{type=symver,name={\ensuremath{\sigma}},description={Empirische Standardabweichung},sort={gr}}
\newglossaryentry{gsigmaP}{type=symver,name={\ensuremath{\sigma}},description={Propagierungsfunktion},sort={grP}}
\newglossaryentry{gphi}{type=symver,name={\ensuremath{\varphi}},description={Aktivierungsfunktion},sort={gu}}
\newglossaryentry{gPhi}{type=symver,name={\ensuremath{\varPhi}},description={Wahrscheinlichkeitsdichte},sort={gU}}

\newglossaryentry{gpi}{type=symver,name={\ensuremath{\pi}},description={Stochastische Strategie eines Reinforcement Learning-Agenten},sort={gp}}

\newglossaryentry{gpsi}{type=symver,name={\ensuremath{\psi}},description={Durchflussfunktion},sort={gw}}

\newglossaryentry{gxi}{type=symver,name={\ensuremath{\xi}},description={Lernrate},sort={gn}}

\newglossaryentry{iA}{type=symver,name={\ensuremath{\text{A}}},description={Ausgabeschicht eines künstlichen neuronalen Netzes},sort={rA}}
\newglossaryentry{iAbg}{type=symver,name={\ensuremath{\text{Abg}}},description={Abgas},sort={rAbg}}
\newglossaryentry{iBatch}{type=symver,name={\ensuremath{b}},description={\gls{iBatch}-tes Element eines Mini-Batches},sort={rBatch}}

\newglossaryentry{iDMV}{type=symver,name={\ensuremath{\text{DMV}}},description={Größe der dynamischen Multiskalenvermessung},sort={rDMV}}

\newglossaryentry{iEth}{type=symver,name={\ensuremath{\text{Eth}}},description={Ethanol},sort={rEth}}
\newglossaryentry{iEnde}{type=symver,name={\ensuremath{\text{Ende}}},description={Ende eines Intervalls},sort={rEnde}}
\newglossaryentry{iZyklus}{type=symver,name={\ensuremath{i}},description={\gls{iZyklus}-ter Verbrennungszyklus},sort={rib}}

\newglossaryentry{iiNeuron}{type=symver,name={\ensuremath{i}},description={\gls{iiNeuron}-tes Eingabeneuron},sort={ria}}

\newglossaryentry{iInj}{type=symver,name={\ensuremath{\text{Inj}}},description={Einspritzung},sort={ric}}

\newglossaryentry{ij}{type=symver,name={\ensuremath{j}},description={\gls{ij}-te Stellgröße},sort={rjb}}
\newglossaryentry{ijNeuron}{type=symver,name={\ensuremath{j}},description={\gls{ijNeuron}-tes Ausgabeneuron},sort={rja}}

\newglossaryentry{ijParam}{type=symver,name={\ensuremath{j}},description={\gls{ijParam}-ter Paramter der Fehlerfunktion \gls{rErr}},sort={rjParam}}
\newglossaryentry{iK}{type=symver,name={\ensuremath{\text{K}}},description={Kühlwasser},sort={rK}}
\newglossaryentry{iKr}{type=symver,name={\ensuremath{\text{Gas}}},description={Kraftstoff},sort={rKr}}
\newglossaryentry{iKlasse}{type=symver,name={\ensuremath{k}},description={\gls{iKlasse}-te Klasse des Vermessungsalgorithmus},sort={rk}}
\newglossaryentry{iL}{type=symver,name={\ensuremath{\text{L}}},description={Luft},sort={rL}}
\newglossaryentry{iLW}{type=symver,name={\ensuremath{\text{LW}}},description={Ladungswechsel},sort={rLW}}
\newglossaryentry{iLim}{type=symver,name={\ensuremath{\text{Lim}}},description={Limit},sort={rLim}}
\newglossaryentry{iVZyklus}{type=symver,name={\ensuremath{m}},description={\gls{iVZyklus}-ter Vermessungszyklus},sort={rm}}
\newglossaryentry{ilayer}{type=symver,name={\ensuremath{l}},description={\gls{ilayer}-te Schicht eines künstlichen neuronalen Netzes},sort={rlayer}}

\newglossaryentry{iRichtung}{type=symver,name={\ensuremath{l}},description={\gls{iRichtung}-te Richtung der Vermessung},sort={rl}}
\newglossaryentry{imax}{type=symver,name={\ensuremath{\text{Max}}},description={Maximum},sort={rmax}}
\newglossaryentry{imin}{type=symver,name={\ensuremath{\text{Min}}},description={Minimum},sort={rmin}}
\newglossaryentry{iNorm}{type=symver,name={\ensuremath{\text{Norm}}},description={Normalisierte Größe},sort={rNorm}}
\newglossaryentry{iN}{type=symver,name={\ensuremath{\text{N}}},description={Nachbarn},sort={rN}}
\newglossaryentry{iNVO}{type=symver,name={\ensuremath{\text{NVO}}},description={Negative Ventilüberschneidung (engl.: negative Valve Overlap)},sort={rNVO}}

\newglossaryentry{iHD}{type=symver,name={\ensuremath{\text{HD}}},description={Hochdruckprozess},sort={rHD}}

\newglossaryentry{ikomp}{type=symver,name={\ensuremath{\text{Komp}}},description={Kompensation},sort={rkomp}}

\newglossaryentry{iKNN}{type=symver,name={\ensuremath{\text{KNN}}},description={Künstliches neuronales Netz},sort={rKNN}}

\newglossaryentry{iOl}{type=symver,name={\ensuremath{\text{Öl}}},description={Öl},sort={rOl}}
\newglossaryentry{iSoll}{type=symver,name={\ensuremath{\text{Set}}},description={Sollgröße},sort={rSoll}}
\newglossaryentry{iSchw}{type=symver,name={\ensuremath{\text{Schw}}},description={Aktivierungsschwelle},sort={rSchw}}
\newglossaryentry{iSau}{type=symver,name={\ensuremath{\text{Sau}}},description={Saugrohr},sort={rSau}}
\newglossaryentry{iStartInt}{type=symver,name={\ensuremath{\text{Start}}},description={Start eines Intervalls},sort={rStartInt}}
\newglossaryentry{iStart}{type=symver,name={\ensuremath{\text{Start}}},description={Startpunkt der Vermessung},sort={rStart}}
\newglossaryentry{iU}{type=symver,name={\ensuremath{\text{U}}},description={Umgebung},sort={rU}}
\newglossaryentry{iUnsicher}{type=symver,name={\ensuremath{\text{Unsafe}}},description={Unsafe state},sort={rUnsicher}}
\newglossaryentry{iSafetyFilter}{type=symver,name={\ensuremath{\text{Safety}}},description={Safety},sort={rUberwachung}}

\newglossaryentry{iRL}{type=symver,name={\ensuremath{\text{RL}}},description={Größe des Reinforcement Learning-Agenten},sort={rRL}}
\newglossaryentry{iRoh}{type=symver,name={\ensuremath{\text{Raw}}},description={Rohwert},sort={rRoh}}
\newglossaryentry{iSicher}{type=symver,name={\ensuremath{\text{Safe}}},description={Safe State},sort={rSicher}}
\newglossaryentry{iTol}{type=symver,name={\ensuremath{\text{Tol}}},description={Toleranz},sort={rTol}}
\newglossaryentry{iTest}{type=symver,name={\ensuremath{\text{Test}}},description={Testdatensatz},sort={rTest}}
\newglossaryentry{iTrain}{type=symver,name={\ensuremath{\text{Train}}},description={Trainingsdatensatz},sort={rTrain}}

\newglossaryentry{iverbrannt}{type=symver,name={\ensuremath{\text{v}}},description={Verbrannt},sort={rvverbrannt}}

\newglossaryentry{iW}{type=symver,name={\ensuremath{\text{W}}},description={Wasser},sort={rW}}
\newglossaryentry{iZyl}{type=symver,name={\ensuremath{\text{cyl}}},description={Zylinder},sort={rZyl}}
\newglossaryentry{iZK}{type=symver,name={\ensuremath{\text{ZK}}},description={Zwischenkompression},sort={rZK}}

%% file: Layout/Abbreviations.tex
\newglossaryentry{abez}{type=abkver,name={bez.},description={bezüglich},sort={abez}}
\newglossaryentry{abzw}{type=abkver,name={bzw.},description={beziehungsweise},sort={abzw}}
\newglossaryentry{adh}{type=abkver,name={d.\,h.},description={das heißt},sort={adh}}
\newglossaryentry{aengl}{type=abkver,name={engl.},description={englisch},sort={aengl}}
\newglossaryentry{aevtl}{type=abkver,name={evtl.},description={eventuell},sort={aevtl}}
\newglossaryentry{aia}{type=abkver,name={i.\,Allg.},description={im Allgemeinen},sort={aia}}
\newglossaryentry{aua}{type=abkver,name={u.\,a.},description={unter anderem},sort={aua}}
\newglossaryentry{ausw}{type=abkver,name={usw.},description={und so weiter},sort={ausw}}
\newglossaryentry{auU}{type=abkver,name={u.\,U.},description={unter Umständen},sort={auU}}
\newglossaryentry{azzt}{type=abkver,name={zzt.},description={zurzeit},sort={azzt}}
\newglossaryentry{azZt}{type=abkver,name={z.\,Zt.},description={zur Zeit},sort={azZt}}

\newglossaryentry{kAO}{type=abkver,name={\text{AÖ}},description={Auslass öffnet},sort={kAO}}
\newglossaryentry{kAS}{type=abkver,name={\text{AS}},description={Auslass schließt},sort={kAS}}
\newglossaryentry{kBRR}{type=abkver,name={BRR},description={Brennraumrückführung},sort={kBRR}}
\newglossaryentry{kCLB}{type=abkver,name={CLB},description={Konfigurierbare Logikblöcke (engl.: configurable Logic Block)},sort={kCLB}}
\newglossaryentry{kCO2}{type=abkver,name={$\text{CO}_2$},description={Kohlenstoffdioxid},sort={kCO2}}
\newglossaryentry{kCFD}{type=abkver,name={CFD},description={Computational-Fluid-Dynamics},sort={kCFD}}

\newglossaryentry{kDDPG}{type=abkver,name={DDPG},description={Deep Deterministic Policy Gradient},sort={kDDPG}}
\newglossaryentry{kDQN}{type=abkver,name={DQN},description={Deep-Q-Network},sort={kDQN}}

\newglossaryentry{kDMV}{type=abkver,name={DMV},description={Dynamische Multiskalenvermessung},sort={kDMV}}
\newglossaryentry{kDFG}{type=abkver,name={DFG},description={Deutsche Forschungsgemeinschaft},sort={kDFG}}

\newglossaryentry{kEO}{type=abkver,name={\text{EÖ}},description={Einlass öffnet},sort={kEO}}
\newglossaryentry{kES}{type=abkver,name={\text{ES}},description={Einlass schließt},sort={kES}}
\newglossaryentry{kEMVT}{type=abkver,name={EMVT},description={Elektromechanischer Ventiltrieb},sort={kEMVT}}
\newglossaryentry{kEKR}{type=abkver,name={EKR},description={Einlasskanalrückführung},sort={kEKR}}

\newglossaryentry{kFF}{type=abkver,name={FF},description={Flip-Flop},sort={kFF}}
\newglossaryentry{kFPGA}{type=abkver,name={FPGA},description={Field-Programmable Gate Array},sort={kFPGA}}
\newglossaryentry{kfAKR}{type=abkver,name={fAKR},description={Frühe Auslasskanalrückführung},sort={kfAKR}}

\newglossaryentry{kHCCI}{type=abkver,name={HCCI},description={Homogene Kompressionszündung (engl.: Homogeneous Charge Compression Ignition)},sort={kHCCI}}

\newglossaryentry{kINZ}{type=abkver,name={INZ},description={In-Zyklus},sort={kINZ}}

\newglossaryentry{kKNN}{type=abkver,name={KNN},description={Künstliches neuronales Netz},sort={kKNN}}

\newglossaryentry{kNVO}{type=abkver,name={NVO},description={Negative Ventilüberschneidung (engl.: negative Valve Overlap)},sort={kNVO}}

\newglossaryentry{kLUT}{type=abkver,name={LUT},description={Lookup-Tabellen},sort={kLUT}}
\newglossaryentry{kLin}{type=abkver,name={Lin},description={Linear},sort={kLin}}
\newglossaryentry{kLEXCI}{type=abkver,name={LExCI},description={Learning and Experiencing Cycle Interface},sort={kLEXCI}}
\newglossaryentry{kLOT}{type=abkver,name={LOT},description={Obere Totpunkt des Ladungswechsels},sort={kLOT}}

\newglossaryentry{kMABX}{type=abkver,name={MABX},description={dSPACE MicroAutoBox},sort={kMABX}}
\newglossaryentry{kMPC}{type=abkver,name={MPC},description={Modellprädiktive Regelung (engl.: Model Predictive Control)},sort={kMPC}}
\newglossaryentry{kMIMO}{type=abkver,name={MIMO},description={Mehrgrößensystem (engl.: Multiple Input, Multiple Output)},sort={kMIMO}}
\newglossaryentry{kMux}{type=abkver,name={Mux},description={Multiplexer},sort={kMux}}
\newglossaryentry{kMSE}{type=abkver,name={MSE},description={Mittlerer quadratischer Fehler (engl.: Mean Square Error)},sort={kMSE}}
\newglossaryentry{kMS}{type=abkver,name={MS},description={Multiskalen},sort={kMS}}
\newglossaryentry{kMEP}{type=abkver,name={MEP},description={Markov-Entscheidungsproblem},sort={kMEP}}
\newglossaryentry{kNOX}{type=abkver,name={\ensuremath{\text{NO}_{\text{x}}}},description={Stickoxid},sort={kNOX}}

\newglossaryentry{kRL}{type=abkver,name={RL},description={Reinforcement Learning},sort={kRL}}
\newglossaryentry{kRPI}{type=abkver,name={RPI},description={Raspberry Pi},sort={kRPI}}
\newglossaryentry{kRelu}{type=abkver,name={ReLu},description={Rectified Linear Unit},sort={kRelu}}
\newglossaryentry{kPRBS}{type=abkver,name={PRBS},description={Pseudozufällige Binärsequenzen (engl.: Pseudorandom Binary Sequence)},sort={kPRBS}}
\newglossaryentry{kPPO}{type=abkver,name={PPO},description={Proximal Policy Optimization},sort={kPPO}}
\newglossaryentry{kpAKR}{type=abkver,name={pAKR},description={Auslasskanalrückführung mit parallelem Ansaugen},sort={kpAKR}}
\newglossaryentry{kPID}{type=abkver,name={PID},description={Proportional-Integral-Differential-Regler},sort={kPID}}

\newglossaryentry{kSISO}{type=abkver,name={SISO},description={Eingrößensystem (engl.: Single Input, Single Output)},sort={kSISO}}
\newglossaryentry{ksAKR}{type=abkver,name={sAKR},description={Späte Auslasskanalrückführung},sort={ksAKR}}
\newglossaryentry{kSPCCI}{type=abkver,name={SPCCI},description={Funkengesteuerte Kompressionszündung (engl.: Spark Controlled Compression Ignition)},sort={kSPCCI}}

\newglossaryentry{kTD3}{type=abkver,name={TD3},description={Twin Delayed Deep Deterministic Policy Gradient},sort={kTD3}}

\newglossaryentry{kUDP}{type=abkver,name={UDP},description={User-Datagramm-Protokoll},sort={kUDP}}
\newglossaryentry{kUT}{type=abkver,name={UT},description={Unterer Totpunkt},sort={kUT}}

\newglossaryentry{kVHDL}{type=abkver,name={VHDL},description={Textbasierte Hardwarebeschreibungssprache (engl.: Very High Speed Integrated Circuit Hardware Description Language) },sort={kVHDL}}

\newglossaryentry{kZ2Z}{type=abkver,name={Z2Z},description={Zyklus-zu-Zyklus},sort={kZ2Z}}
\newglossaryentry{kZOT}{type=abkver,name={ZOT},description={Zündung oberer Totpunkt},sort={kZOT}}

%% file: Chapters/Abstract.tex
This work introduces a toolchain for applying Reinforcement Learning (RL), specifically the Deep Deterministic Policy Gradient (DDPG) algorithm, in safety-critical real-world environments. As an exemplary application, transient load control is demonstrated on a single-cylinder internal combustion engine testbench in Homogeneous Charge Compression Ignition (HCCI) mode, that offers high thermal efficiency and low emissions. However, HCCI poses challenges for traditional control methods due to its nonlinear, autoregressive, and stochastic nature.

RL provides a viable solution, however, safety concerns --~such as excessive pressure rise rates~-- must be addressed when applying to HCCI. A single unsuitable control input can severely damage the engine or cause misfiring and shut down. Additionally, operating limits are not known a priori and must be determined experimentally. To mitigate these risks, real-time safety monitoring based on the k-nearest neighbor algorithm is implemented, enabling safe interaction with the testbench.

The feasibility of this approach is demonstrated as the RL agent learns a control policy through interaction with the testbench. A root mean square error of $0.1374\,\si{\bar}$ is achieved for the indicated mean effective pressure, comparable to neural network-based controllers from the literature. The toolchain's flexibility is further demonstrated  by adapting the agent's policy to increase ethanol energy shares, promoting renewable fuel use while maintaining safety.

This RL approach addresses the longstanding challenge of applying RL to safety-critical real-world environments. The developed toolchain, with its adaptability and safety mechanisms, paves the way for future applicability of RL in engine testbenches and other safety-critical settings.

%% file: Chapters/IntroductionNewFocusRL.tex
\section{Introduction and Motivation}
\label{chap:Introduction}

Reinforcement Learning (RL) is a powerful Machine Learning (ML) paradigm which offers distinct advantages over traditional methods in the context of adaptive control. 

Its model-free algorithms eliminate the need for explicit system modeling, thus significantly reducing engineering effort, and its policies --~often represented as artificial neural networks (ANNs)~-- enable rapid execution, making them suitable for real-time applications. Furthermore, RL agents can uncover hidden patterns in the environment, potentially surpassing domain experts' knowledge~\citep{Badalian2024, Picerno2023}. 

Central to RL’s effectiveness is its learning mechanism based on interactions with the environment. By evaluating and refining its policy based on feedback from the environment, an RL agent can adapt to dynamic, high-dimensional systems without relying o.n precise models. These characteristics make RL a compelling solution for complex control problems.

However, applying RL in real-world settings, particularly in safety-critical systems, remains challenging. RL’s reliance on exploration to optimize behavior inherently involves the risk of unsafe or suboptimal actions during the learning process, which can compromise safety by destabilizing the system, causing mechanical damage or even threatening human health. These safety concerns  represent a major barrier to deploying RL in real-world environments~\citep{DulacArnold.2021} and making it an active area of research~\citep{Kwon.2023}.

This challenge is particularly evident in internal combustion engine control, where RL has already demonstrated its potential for automated function development~\citep{Koch2023}, boost pressure control~\citep{Hu2019}, and emission reduction~\citep{Picerno2023}. However, these RL applications are often limited to virtual environments due to the aforementioned concerns. To address this, additional measures must be implemented to monitor the agent's actions to guarantee safety. In \citep{Norouzi.2023}, a safe RL approach for emission control in diesel engines using the Deep Deterministic Policy Gradient (DDPG) algorithm is proposed, where actions are constrained through a quadratic programming solver to prevent unsafe actions. Nevertheless, this approach remains confined to simulation. 

In contrast, real-world applications of RL for engine control remain exceedingly rare. In \citep{Maldonado.2024}, a control policy for adjusting fuel injection is learned using Q-Learning, though the action space is limited to a single action with a narrow range, eliminating the need for additional safety mechanisms. In \citep{Hu2021}, a Deep Q Network (DQN) is employed for boost control of a real-world diesel engine using a safety shield presented in~\citep{alshiekh2018safe}.  However, employing DQN limits their method do discrete action spaces. To the best of the authors' knowledge, there have been no RL applications for real-world combustion engines involving multiple actions and continuous action spaces.

To overcome these limitations, in this work, we introduce DDPG --~suited for continuous state and action spaces~-- with a multi-dimensional safety monitoring to preemptively identify unsafe actions in real-time, preventing them from being applied. As an exemplary application, we consider transient load control of an engine operating in Homogeneous Charge Compression Ignition (HCCI) mode. HCCI is a promising low-temperature combustion technique that achieves both high thermal efficiency and low emissions~\citep{Li.2001, Kulzer.2009}. Unlike conventional spark-ignition and compression-ignition engines, HCCI utilizes the auto-ignition of a homogeneously mixed charge of air, gasoline and residual gas. The latter is trapped in the cylinder by negative valve overlap (NVO) and transferred into the next combustion cycle, raising the mixture temperature. This results in rapid, low-temperature combustion, reducing nitrogen oxide and particulate matter emissions~\citep{Yao.2009, Brassat.2013, Wick.2018} while increasing efficiency. 

However, controlling HCCI is challenging due to nonlinearities and high cyclic variability~\citep{Hellstrom.2012}, which arise from autoregressive coupling through transfer of residual gas from cycle to cycle. This can lead to stochastic outlier cycles, characterized by incomplete combustion or misfires. As a result, traditional control methods, such as rule-based or model-free controllers~\citep{Wick.2018, Wick.2019, Gordon.2019}, often encounter difficulties in maintaining operational stability and efficiency under varying loads and conditions. Although model predictive control (MPC) has shown potential for HCCI control~\citep{Albin.2015, Bengtsson.10420061062006, Ebrahimi.2018, Nuss.2019, Chen.2023}, the key challenge is to  identify an accurate model. Due to real-time constraints, these models often need to be simplified, compromising their precision.

In response to these challenges, the HCCI research field has increasingly adopted learning-based approaches. Given the high-dimensional, multiple-input, multiple-output behavior and nonlinear characteristics of HCCI, data-driven methods, particularly those employing ANNs, have proven to be promising solutions. These controllers often rely on cycle-integral values to characterize combustion, such as the total heat released~\gls{rQ}, the combustion phasing~\gls{gAlpha50} --~defined as the crank angle where $50\,\%$ of the fuel has burned~-- and the indicated mean effective pressure (IMEP), representing the engine's load. Successful control implementations with ANNs utilize Extreme Learning Machines~\citep{Vaughan.14.10.2013} and the inversion of the system dynamics~\citep{Wick.2020, Bedei.2023, Bedei.2023b}. Recent advancements have further enabled the integration of recurrent neural networks (RNNs) with long short-term memory (LSTM) into nonlinear MPC frameworks, significantly enhancing control performance in HCCI applications~\citep{Gordon.2024}.

RL extends these traditional learning-based methods by enabling agents to learn directly through interaction with the environment. Unlike other data-driven paradigms, RL combines data generation with the learning of an optimized control policy, allowing controllers to adapt effectively to real-time variations in HCCI combustion dynamics. The nonlinear, autoregressive, and stochastic nature of HCCI, combined with the lack of sufficiently accurate models, necessitates data generation through direct interaction with the engine~\citep{Wick.2020}. This real-world interaction is essential for capturing the complex cross-couplings between states and actions, highlighting RL's significant potential in this application. 

Moreover, RL's data generation capability facilitates transfer learning, allowing agents to adapt to system drifts, new boundary conditions, or changing objectives without starting the learning process from the beginning. For combustion engines, RL can, for instance, directly explore the behavior of untested renewable fuels in a testbench environment by leveraging previously trained policies. This eliminates the need for entirely new extensive datasets, accelerating and refining assessments of renewable fuels directly within a real-world setting. This adaptability surpasses the capabilities of traditional control methods, presenting novel opportunities for research and development.

To fully realize these potential benefits of RL, safe exploration within the real-world environment is required. This work introduces a safe RL approach designed to ensure safe interaction within such environments. First, we outline the RL fundamentals and the experimental setup, followed by the development of a toolchain based on the Learning and Experiencing Cyclic Interface (LExCI), a free and open-source tool,  developed in \citep{Badalian2024}, which enables RL with embedded hardware. This toolchain is integrated into the HCCI testbench to facilitate RL in a real-world setting. We then detail the methodology employed for safety monitoring to ensure operational safety. Finally, we validate the toolchain by comparing it to an ANN-based reference strategy developed in~\citep{Bedei.2023}. Additionally, we demonstrate the transfer learning abilities by adaptation of the agent's policy  to increase the proportion of renewable fuels, specifically ethanol, substituting part of the gasoline --~highlighting potential future directions for RL research in the context of real-world engine control.

%% file: Chapters/RLFundamentals.tex
\label{chap:Fundamentals}
\section{Reinforcement Learning Fundamentals}

RL is based on the Markov Decision Process (MDP), which models the interaction between an agent --~a decision-making entity that selects actions to maximize a cumulative reward~-- and its environment. 
The environment is represented by a state space \gls{rX}, an action space \gls{rURaum}, transition probabilities $\gls{rP}\left(\gls{rxStateVec}_{\gls{iZyklus}} \mid \gls{rxStateVec}_{\gls{iZyklus}-1}, \gls{ruCtrlVec}_{\gls{iZyklus}}\right)$ and rewards $\gls{rRReward}(\gls{rxStateVec}_{\gls{iZyklus}-1}, \gls{ruCtrlVec}_{\gls{iZyklus}},\gls{rxStateVec}_{\gls{iZyklus}})$.
At each discrete time step, in case of HCCI control each combustion cycle $\gls{iZyklus}$, the agent observes the current state~$\gls{rxStateVec}_{\gls{iZyklus}-1}$, selects actions $\gls{ruCtrlVec}_{\gls{iZyklus}}$, resulting in state $\gls{rxStateVec}_{\gls{iZyklus}}$ and receives a reward $\gls{rRReward}_{\gls{iZyklus}}$, which evaluates the quality of the chosen actions. From this, an experience tuple $\gls{rTRL}_{\gls{iZyklus}}=\left(\gls{rxStateVec}_{\gls{iZyklus}-1},\,\gls{ruCtrlVec}_{\gls{iZyklus}},\,\gls{rxStateVec}_{\gls{iZyklus}},\,\gls{rRReward}_{\gls{iZyklus}},\,\gls{rdRL}_{\gls{iZyklus}}\right)$ is formed, where $\gls{rdRL}_{\gls{iZyklus}}$ is a binary termination indicator marking the end of an episode $\gls{rERL}=(\gls{rTRL}_{1},\gls{rTRL}_{2},...,\gls{rTRL}_{\gls{rnumber}})$, which is a sequence of consecutive experiences~\gls{rTRL}.

The agent's goal is to find an optimal policy $\gls{gmRL}^*$, which maximizes its return $\gls{rG}$ over time. The return is the cumulative reward, computed using a discount factor $\gls{ggamma}\leq1$, which weights future rewards compared to immediate ones: 
\begin{equation}
	\gls{rG}_{\gls{iZyklus}}=\sum _{k=0}^{\infty }{\gls{ggamma} }^{k}{\gls{rRReward}}_{\gls{iZyklus}+k}
\end{equation}
To iteratively update the policy in order to maximize the return, typically an evaluation function, such as the action-value function (Q-function), is used. The Q-function describes the expected return when taking a specific action $\gls{ruCtrlVec}_{\gls{iZyklus}}$ in a given state $\gls{rxStateVec}_{\gls{iZyklus}-1}$ and then following the policy $\gls{gmRL}$:
\begin{equation}
	\gls{rQfunc}(\gls{rxStateVec}_{\gls{iZyklus}-1}, \gls{ruCtrlVec}_{\gls{iZyklus}})= \mathbb{E}_{\gls{gmRL}} \left[ \gls{rG}_{\gls{iZyklus}}     \mid \gls{rX}_{0} = \gls{rxStateVec}_{\gls{iZyklus}-1}, \gls{rURaum}_{0} = \gls{ruCtrlVec}_{\gls{iZyklus}} \right] = \gls{rRReward}_{\gls{iZyklus}}( \gls{rxStateVec}_{\gls{iZyklus}-1}, \gls{ruCtrlVec}_{\gls{iZyklus}})+\gls{ggamma}\mathbb{E}_{\gls{gmRL}} \left[ \sum_{k=0}^{\infty} \gls{ggamma}^k \gls{rRReward}_{\gls{iZyklus}+k} \mid \gls{rX}_{0} = \gls{rxStateVec}_{\gls{iZyklus}} \right]	
\end{equation}
To apply RL to HCCI control, the specific problem requirements lead to the following considerations that must be taken into account when selecting an RL algorithm:
\begin{enumerate}
	\item Accuracy of existing process models insufficient: Model-free approach required.
	\item Capability to leverage existing data for offline learning.
	\item High data efficiency for reduced training time in a real-world environment.
	\item Stability and robustness of the learning process.
	\item Suitability for continuous state and action spaces.
\end{enumerate}
The DDPG algorithm is a model-free, off-policy, actor-critic algorithm that satisfies the requirements outlined above. Specifically, it is model-free, meaning it does not require a process model and can learn from direct interactions with the environment, making it ideal for the control of HCCI engines. Additionally, as an off-policy algorithm, DDPG is capable of leveraging existing data through its experience replay buffer, enabling it to learn also from data that have not been generated with the agent's policy itself. The replay buffer also ensures high data efficiency and improved convergence behavior~\citep{Lin.1992}, allowing DDPG to learn from relatively few interactions with the environment, which is crucial for reducing training time, especially in real-world settings. Moreover, DDPG is designed to work in continuous state and action spaces, making it particularly suited for real-time control of processes like HCCI, where both states and actions are continuous. Therefore, DDPG is employed for HCCI control in the following. The key features of the DDPG algorithm and its mathematical foundations are discussed in detail in~\citep{Lillicrap.2015}.

DDPG is using an actor-critic-architecture where both  the deterministic policy $\gls{gmRL}_{\gls{gthetaParam}_{\gls{gmRL}}}$ and the approximation of the Q-function $\hat{\gls{rQfunc}}_{\gls{gthetaParam}_{\gls{rQ}}}$ are represented by ANNs with parameter sets $\gls{gthetaParam}_{\gls{gmRL}}$ and $\gls{gthetaParam}_{\gls{rQ}}$. Alongside these, \gls{kDDPG} employs target networks $\gls{gmRL}'_{\gls{gthetaParam}_{\gls{gmRL}'}}$, $\hat{\gls{rQfunc}'}_{\gls{gthetaParam}_{\gls{rQfunc}'}}$, providing target values for the training. These are updated significantly slower than the actor and the critic in order to increase the numerical stability and improve the convergence behavior of the training~\citep{Lillicrap.2015}.

The parameters $\gls{gthetaParam}_{\gls{rQfunc}}$ of the approximated Q-function~$\hat{\gls{rQfunc}}_{\gls{gthetaParam}_{\gls{rQfunc}}}$ are updated by minimizing the following loss function $\gls{rErr}$, incorporating the Bellman equation:
\begin{equation}
	\label{eq:DDPG1}
	\gls{rErr}_{\gls{gthetaParam}_{\gls{rQfunc}},\gls{gthetaParam}'_{\gls{rQfunc}},\gls{gthetaParam}'_{\gls{gmRL}}}=
	\biggl( \hat{\gls{rQfunc}}_{\gls{gthetaParam}_{\gls{rQfunc}}}(\gls{rxStateVec}_{\gls{iZyklus}-1},\gls{ruCtrlVec}_{\gls{iZyklus}})
	- \biggl[\gls{rRReward}_{\gls{iZyklus}}+\gls{ggamma}\cdot(1-\gls{rdRL}_{\gls{iZyklus}})\cdot\hat{\gls{rQfunc}}'_{\gls{gthetaParam}'_{\gls{rQfunc}}}(\gls{rxStateVec}_{\gls{iZyklus}},\gls{gmRL}'_{\gls{gthetaParam}'_{\gls{gmRL}}}(\gls{rxStateVec}_{\gls{iZyklus}}))\biggr] \biggr)^2
\end{equation}
Typically, gradient descent with learning rate~$\gls{gxi}_{\gls{rQ}}$ is used to train the critic $\hat{\gls{rQfunc}}$ in order to minimize the loss \gls{rErr} of the Q-value approximation: 
\begin{equation}
	\label{eq:DDPG2}
	\gls{gthetaParam}_{\gls{rQfunc}}\gets \gls{gthetaParam}_{\gls{rQfunc}} -\gls{gxi}_{\gls{rQfunc}}  \cdot	\gls{aNabla}_{\gls{gthetaParam}_{\gls{rQfunc}}}	\gls{rErr}_{\gls{gthetaParam}_{\gls{rQfunc}},\gls{gthetaParam}'_{\gls{rQfunc}},\gls{gthetaParam}'_{\gls{gmRL}}}
\end{equation}
The parameters $\gls{gthetaParam}_{\gls{gmRL}}$ of the actor network are updated via gradient ascent using the critic network $\hat{\gls{rQfunc}}$ to maximize the Q function:
\begin{equation}
	\label{eq:DDPG3}
	\gls{gthetaParam}_{\gls{gmRL}}\gets \gls{gthetaParam}_{\gls{gmRL}} +\gls{gxi}_{\gls{gmRL}}  \cdot	\gls{aNabla}_{\gls{gthetaParam}_{\gls{gmRL}}}\hat{\gls{rQfunc}}_{\gls{gthetaParam}_{\gls{rQfunc}}}\left(\gls{rxStateVec},\gls{gmRL}_{\gls{gthetaParam}_{\gls{gmRL}}}(\gls{rxStateVec}) \right)
\end{equation}
The parameters of the target networks $\gls{gthetaParam}'_{\gls{rQfunc}},\gls{gthetaParam}'_{\gls{gmRL}}$, are updated significantly slower with Polyak averaging using the factor $\gls{grhoPoly}\ll1$:

\begin{equation}
	\label{eq:DDPG4}
	\gls{gthetaParam}'_{\gls{rQfunc}}\gets \gls{grhoPoly}\gls{gthetaParam}_{\gls{rQfunc}} + (1-\gls{grhoPoly})\cdot \gls{gthetaParam}'_{\gls{rQfunc}}
\end{equation}
\begin{equation}
	\label{eq:DDPG5}
	\gls{gthetaParam}'_{\gls{gmRL}}\gets \gls{grhoPoly}\gls{gthetaParam}_{\gls{gmRL}} + (1-\gls{grhoPoly})\cdot \gls{gthetaParam}'_{\gls{gmRL}}	
\end{equation}
The deterministic policy $\gls{gmRL}_{\gls{gthetaParam}_{\gls{gmRL}}}$ always acts greedily to maximize the approximated Q-function $\hat{\gls{rQfunc}}$, without exploring the action space. However, exploration is crucial to gain new and potentially higher value experiences. Thus, exploratory noise \gls{rNG} is added to the policy:
\begin{equation}
	\label{eq:Policy}
	\gls{ruCtrlVec}_{\gls{iZyklus}}=\gls{gmRL}_{\gls{gthetaParam}_{\gls{gmRL}}}\left(\gls{rxStateVec}_{\gls{iZyklus}-1}\right)+\gls{rNG}(0,\gls{gsigma}^2) 
\end{equation}
Gaussian noise \gls{rNG} with a standard deviation \gls{gsigma} is used, which results in actions that deviate from the policy~\gls{gmRL} also being tested in the environment. Typically, the standard deviation is reduced over time using a decay factor $\gls{gLambdaRL}<1$, which is updated for example once per episode ($\gls{gsigma}\gets\gls{gsigma}\cdot\gls{gLambdaRL}$), in order to reduce exploration through noise and increasingly follow the policy $\gls{gmRL}_{\gls{gthetaParam}_{\gls{gmRL}}}$ itself.

%% file: Chapters/ExperimentalSetup.tex
\section{Experimental Setup and Toolchain Integration}
\label{chap:ExpSetup}
This study utilizes a single-cylinder research engine (SCRE) with a displacement of $\gls{rVH}=0.5\,\si{\liter}$ and a compression ratio of 12. The SCRE is equipped with two direct injectors for fuel and ethanol, respectively. Additionally, the SCRE features a fully variable electromechanical valve train (EMVT), where the opening and closing of the valves is achieved by alternating energization of two solenoid coils. This enables HCCI operation with NVO to leverage internal exhaust gas recirculation, contributing to elevated mixture temperatures that support auto-ignition during the compression phase. It also enables throttle free operation, reducing gas exchange losses significantly. Both the fuel injections and NVO can be adjusted on a cycle-to-cycle basis, making them suitable variables for process control. 

An overview of the SCRE parameters including conditioning parameters is given in Table~\ref{tab:SCRE_param}, while fuel properties are listed in Table~\ref{tab:fuelData}. 

\begin{table}[!htpb]
	\caption{Single-Cylinder Research Engine and Conditioning Parameters.}
	\label{tab:SCRE_param}
	\begin{center}
		\begin{tabular}{lll}
			\toprule
			& Parameter           		& Value \\
			\midrule
			Geometry  			&Displaced Volume&499\,$\text{cm}^3$\\ 
			&Stroke&90\,mm \\
			&Bore&84\,mm\\
			&Compression Ratio&12:1\\
			\midrule
			Conditioning   &Intake Pressure&1013\,mbar\\
					   	   &Exhaust Pressure&1013\,mbar\\
			&Oil Temperature&105\,\si{\degreeCelsius}\\
			&Coolant Temperature&90\,\si{\degreeCelsius}\\
			&Fuel Rail Pressure&100\,bar\\
			&Ethanol Rail Pressure&60\,bar\\
			&Intake Temperature&50\,\si{\degreeCelsius}\\
			\midrule

		\end{tabular}
	\end{center}
\end{table}

\begin{table}[!htpb]
	\caption{Fuel Properties: Gasoline Values from Internal Fuel Analysis and Ethanol Values from \citep{Qi.2016}.}
	\label{tab:fuelData}
	\begin{center}
		\begin{tabular}{lll}
			\toprule
			Parameter							& Gasoline           				& Ethanol \\	
			\midrule
			Research octane number  				& 96 									& 106\\
			Motor octane number 					& 85									& 89\\
			Lower calorific value			& $44.3\,\si{\mega\joule\per\kilo\gram}$	& $26.8\,\si{\mega\joule\per\kilo\gram}$\\
			Ethanol mass fraction 				& $10.4\,\%$ 							& $100\,\%$\\
			Water content				& \SI{360}{\milli\gram\per\kilo\gram} 	& \SI{0}{\milli\gram\per\kilo\gram}\\
			Density (\SI{20}{\degreeCelsius})	& $745.8\,\si{\kilogram\per\cubic\meter}$	& \SI{790}{\kilogram\per\cubic\meter}\\
			\bottomrule
		\end{tabular}
	\end{center}
\end{table}

The SCRE is controlled using a dSPACE Microautobox (\gls{kMABX})~III (1403/\allowbreak 1513/\allowbreak 1514) with the Multi I/O Board DS1552B1~\citep{dSPACEGmbH.2024}. In addition to a quad-core ARM Cortex-A15 real-time processor running at $1.4$ GHz, this control unit features a Xilinx Kintex-7 XC7K325T Field-Programmable Gate Array~(\gls{kFPGA}) with a task rate of $12.5\,\si{\nano\second}$. Furthermore, a Raspberry Pi 400~(\gls{kRPI})~\citep{RaspberryPiFoundation.2024}, equipped with a quad-core ARM Cortex-A72 processor with a base clock speed of $1.8$~GHz, is integrated into the testbench.

The algorithms implemented in this research are allocated between the FPGA, the processor and the \gls{kRPI}, depending on the specific requirements of each calculation. 

Cylinder pressure indication, based on the work of~\citep{Pfluger.2012}, is employed on the FPGA. This enables the calculation of cycle integral parameters such as \gls{rpmi}, maximum pressure rise rate $\gls{rDPMAX}$, determined through numerical integration and differentiation, respectively. Additionally, heat release $\gls{rQ}$ and combustion phasing~$\gls{gAlpha50}$, which describe the thermodynamic state of the mixture, are computed using the first law of thermodynamics and a real-time gas exchange model based on~\citep{Gordon.2020b}, allowing for the calculation of the residual gas fraction. Moreover, ion current signal analysis provides chemical information on the current mixture state by analyzing the maximum $\gls{rIonMax}$ and the integral $\gls{rIonIntegral}$ of the signal. The complementary use of pressure and ion current sensors for process control has shown significant benefits~\citep{Bedei.2023b}, which is why both sensors are employed in this study. 
Additionally, the signals to actuate the EMVT and injectors are generated using Transistor-Transistor Logic on the FPGA, delivering precise short pulses to control valve and injector opening durations, and are transmitted to the corresponding power electronics.


The higher-level engine control is handled by the processor, which operates at slower task rates, with the smallest being 1 ms. This is sufficient for controlling certain conditioning parameters, such as  rail and exhaust pressure. Additionally, several algorithms in this study are executed on the processor. These include an ANN used as a reference control  strategy~\citep{Bedei.2023}, a dynamic measurement algorithm based on \citep{Wick.2020} and the safety monitoring developed in this work, which is an enabler for applying RL in real-world environments.

Finally, the RL-specific algorithms, including policy execution\footnote{Initially, an MABX~II was used for this project, which does not support compilation of TensorFlow Lite, making it impossible to run the policy directly on this hardware. After switching to the newer MABX~III, which resolves this limitation, the policy execution on the RPI was nevertheless retained. Direct execution of the policy on the newer hardware is feasible and will be implemented in future projects to minimize latencies.} and the training process, are executed on the RPI. Communication with the primary control unit is carried out via an Ethernet interface. 

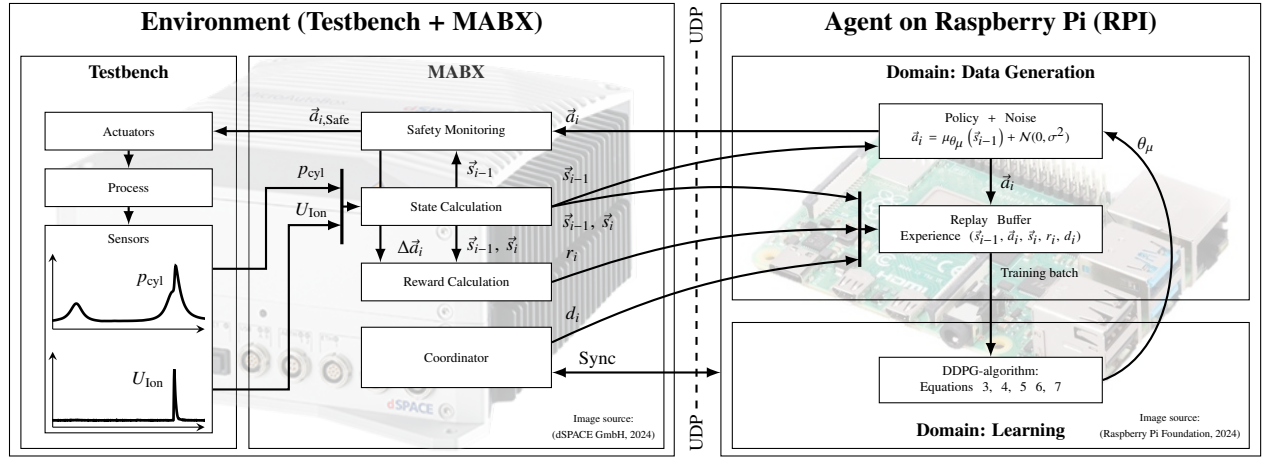
\begin{figure*}[!htbp]
	\centering
	\input{Figures/RLEinbindugPST}
	\caption{Integration of \gls{kDDPG} into the Testbench Environment Using \gls{kLEXCI}~\citep{Badalian2024}.}
	\label{Fig:RLEinbindugPST}    
\end{figure*}

Figure~\ref{Fig:RLEinbindugPST} presents an overview of the relevant functions and data flows, illustrating the integration of the DDPG algorithm into the testbench environment.

The integration is based on the LExCI framework~\citep{Badalian2024}, which facilitates RL on embedded systems by leveraging the Python libraries Ray/RLlib~\citep{Moritz.2018, Liang.2018} and TensorFlow~\citep{Abadi.2015}. RLlib provides high-level abstractions for the implementation, training, and testing of RL algorithms. It is optimized for distributed systems and is thus unsuitable for prototype control units like the MABX due to high computational and memory demands. As a solution, the RL algorithms are offloaded to an additional processing unit, the RPI, where the training, experience replay buffer, and execution of the policy are managed. RLlib uses TensorFlow in the background for model training and policy execution.

Communication with the MABX, which is considered part of the environment here, is conducted via an Ethernet interface using the User Datagram Protocol (UDP). For each combustion cycle, the current state~$\gls{rxStateVec}_{\gls{iZyklus}-1}$ is determined on the FPGA and transmitted to the RPI. Upon receipt, the policy is executed with added exploratory Gaussian noise, and actions~$\gls{ruCtrlVec}_{\gls{iZyklus}}$ are sent back to the MABX, where they are checked using the safety monitoring function, described in detail in Section~\ref{chap:Safety Monitoring}. Verified safe actions $\gls{ruCtrlVec}_{\gls{iZyklus},\gls{iSicher}}$ are then applied to the process, with the resulting cylinder pressure $\pzyl$ and ion current $\gls{rUIon}$ measured to update the state $\gls{rxStateVec}_{\gls{iZyklus}}$. Additionally, reward calculation $\gls{rRReward}_{\gls{iZyklus}}$ is executed on the MABX, using information from both the state determination and safety monitoring. This reward, along with the state and the boolean termination indicator $\gls{rdRL}_{\gls{iZyklus}}$, is returned to the RPI, where it is stored in the experience replay buffer. Moreover, a coordinator is implemented on the MABX, acting as a supervisory module that manages all interactions between the processing units. It coordinates events such as the start and end of episodes, synchronizing operations on both the MABX and RPI to ensure real-time capability and data consistency.

Training of the actor, critic and corresponding target networks is performed on the RPI following each episode, with a training batch randomly sampled from the most recent episode. Additionally, replay trainings are performed using random samples from the experience replay buffer, reducing sequential dependency and enhancing training stability~\citep{Zhang.2017}. Replay training also helps prevent catastrophic forgetting, where ANNs may lose previously learned knowledge when exposed to new data~\citep{McCloskey.1989}. To validate the learned policy, validation episodes are periodically conducted without exploratory noise.

%% file: Figures/RLEinbindugPST.tex
\begin{tikzpicture}[>=latex]
	\tikzstyle{arrow} = [thick,->,>=latex]
	\tikzstyle{arrowb} = [thick,<->,>=latex]
	\def\boxheight{6cm}
	\def\totalwidth{\textwidth}
	\def\boxgap{0.6cm}
	\def\innerboxgap{0.15cm}
	\def\boxwidthleft{(0.55*\totalwidth - 0.5*\boxgap)}
	\def\boxwidthright{(0.45*\totalwidth - 0.5*\boxgap)}
	\def\innerBoxHeight{5.2cm}  

	\def\widthPruefstand{0.35*\boxwidthleft-\innerboxgap*3/2}
	\def\widthMABX{0.65*\boxwidthleft-\innerboxgap*3/2}
	
	\def\labelshift{0.0cm}
	\def\linelabelshift{-0.25cm}
	\def\innerlabelshift{0.0cm}
	
	\def\xPosRachel{0cm}
	\def\yPosRachel{-5cm}
	\def\imagewidth{2.5cm}
	
	\node[draw, minimum width=\boxwidthleft, minimum height=\boxheight, anchor=north west] (umgebung) at (0,0) {};
	\node[anchor=north, yshift=\labelshift] at (umgebung.north) {\textbf{Environment (Testbench + \gls{kMABX})}};
	
	\node[draw, minimum width=\widthPruefstand, minimum height=\innerBoxHeight, anchor=north west] 
	(pruefstand) at (\innerboxgap,-0.7cm) {};
	\node[anchor=north, yshift=\labelshift] at (pruefstand.north) {\scriptsize \textbf{Testbench}};
	
	\node[draw, minimum width=\widthMABX, minimum height=\innerBoxHeight, anchor=north west] 
	(mabx) at ($(pruefstand.north east) + (\innerboxgap,0)$) {};
	\node[anchor=north, yshift=\labelshift] at (mabx.north) {\scriptsize \textbf{\gls{kMABX}}};
	
	\node[inner sep=0pt, opacity=0.25, anchor=center] at ([xshift=0, yshift=-0.3cm] umgebung.center) 
	{\includegraphics[clip, width=0.5\textwidth, trim=0 0 0 0]{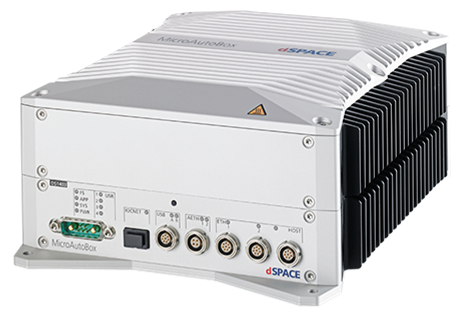}};
	\node[above left, font=\tiny] at (mabx.south east) {\scalebox{0.8}{\shortstack{Image source:\\ \citep{dSPACEGmbH.2024}}}};
	
	\node[draw, minimum width=\boxwidthright, minimum height=\boxheight, anchor=north west] 
	(agent) at ($(umgebung.north east) + (\boxgap,0)$) {};
	\node[anchor=north, yshift=\labelshift] at (agent.north) {\textbf{Agent on Raspberry~Pi (\gls{kRPI})}};
	
	\node[inner sep=0pt, opacity=0.25, anchor=center] at ([xshift=0, yshift=-0.3cm] agent.center) 
	{\includegraphics[clip, width=0.4\textwidth, trim=0 0 0 0]{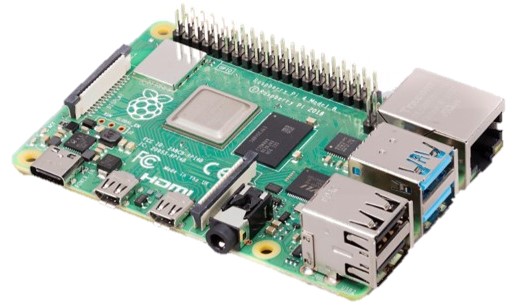}};

	\node[draw, minimum width=\boxwidthright-2*\innerboxgap, minimum height=0.65*\innerBoxHeight-\innerboxgap, anchor=north west] 
	(Learner) at ($(umgebung.north east) + (\boxgap+\innerboxgap,-0.7cm)$) {};
	\node[anchor=north, yshift=\labelshift] at (Learner.north) {\scriptsize \textbf{Domain: Data Generation}};

	\node[draw, minimum width=\boxwidthright-2*\innerboxgap, minimum height=0.35*\innerBoxHeight-\innerboxgap, anchor=north west] 
	(Generation) at ($(umgebung.north east) + (\boxgap+\innerboxgap,-0.7cm-0.65*\innerBoxHeight-\innerboxgap)$) {};
	\node[anchor=south, yshift=-\labelshift] at (Generation.south) {\scriptsize \textbf{Domain: Learning}};
	
	\node[above left, font=\tiny] at (Generation.south east){\scalebox{0.8}{\shortstack{Image source:\\ \citep{RaspberryPiFoundation.2024}}}}; %
	
	
	   
	\node[draw, fill=white, text centered, minimum width=2.5cm, minimum height=0.5cm,yshift=-1.0cm] 
	(stellgroessen) at (mabx.north) {\tiny Safety Monitoring};
	\node[anchor=center, draw, fill=white, text centered, minimum width=2.5cm, minimum height=0.5cm, yshift=-3cm] (reward) at (mabx.north) {\tiny Reward Calculation};
	\node[anchor=center, draw, fill=white, text centered, minimum width=2.5cm, minimum height=0.8cm, yshift=-4.0cm] (Koordinator) at (mabx.north) {\tiny Coordinator};
	\draw[arrow] ([xshift=-1cm] stellgroessen.south)-- ([xshift=-1cm]reward.north)node[pos=0.8,xshift=0.4cm,yshift=-0.1cm]{\scriptsize $\Delta \gls{ruCtrlVec}_{\gls{iZyklus}}$};
	\node[draw, fill=white, text centered, minimum width=2.5cm, minimum height=0.5cm,yshift=-2cm] 
	(Zustand) at (mabx.north) {\tiny State Calculation}; 
	
	\node[draw, fill=white, text centered, minimum width=2.2cm, minimum height=0.5cm,yshift=-1.0cm] 
	(Aktorik) at (pruefstand.north) {\tiny Actuators};
	\node[draw, fill=white, text centered, minimum width=2.2cm, minimum height=0.5cm,yshift=-1.75cm] 
	(Prozess) at (pruefstand.north) {\tiny Process};
	\node[draw, fill=white, text centered, minimum width=2.2cm, minimum height=2.75cm,yshift=-3.625cm] 
	(Sensorik) at (pruefstand.north){};
	\node[anchor=north, yshift=\labelshift] at (Sensorik.north) {\tiny Sensors};

	\node[draw, fill=white, text centered,text width=2.7cm, minimum height=0.5cm,yshift=-3cm] 
	(Replay) at (agent.north) {\baselineskip=10pt \tiny Replay Buffer \\ \tiny Experience ($\gls{rxStateVec}_{\gls{iZyklus}-1},\,\gls{ruCtrlVec}_{\gls{iZyklus}},\,\gls{rxStateVec}_{\gls{iZyklus}},\,\gls{rRReward}_{\gls{iZyklus}},\,\gls{rdRL}_{\gls{iZyklus}}$)\par};
	\node[draw, fill=white, text centered,text width=2.7cm, minimum height=0.5cm,yshift=-1.7cm] 
	(Policy) at (agent.north) {\baselineskip=10pt \tiny Policy + Noise \\\tiny  $\gls{ruCtrlVec}_{\gls{iZyklus}}=\gls{gmRL}_{\gls{gthetaParam}_{\gls{gmRL}}}\left(\gls{rxStateVec}_{\gls{iZyklus}-1}\right)+\gls{rNG}(0,\gls{gsigma}^2)$\par};
	\node[draw, fill=white, text centered,text width=2.7cm, minimum height=0.5cm,yshift=-5cm,xshift=0cm] 
	(KNNUpdate) at (agent.north) {\baselineskip=10pt \tiny \gls{kDDPG}-algorithm: \\ \tiny  Equations~\ref{eq:DDPG1},~\ref{eq:DDPG2},~\ref{eq:DDPG3}~\ref{eq:DDPG4},~\ref{eq:DDPG5}\par};

	
	
	\path [arrow, bend left=10] (Zustand.east) edge node[pos=0,right,xshift=0cm,yshift=-0.2cm] {\scriptsize $\gls{rxStateVec}_{\gls{iZyklus}-1},\,\gls{rxStateVec}_{\gls{iZyklus}}$} ([xshift=-0.25cm,yshift=0.4cm] Replay.west);
	\path [arrow, bend left=10] (Zustand.east) edge node[pos=0,right,xshift=0cm,yshift=0.42cm] {\scriptsize $\gls{rxStateVec}_{\gls{iZyklus}-1}$} ([xshift=0cm,yshift=-0.2cm] Policy.west);

	\path [arrow, bend left=10] (reward.east) edge node[pos=0,right,xshift=0.05cm,yshift=0.35cm] {\scriptsize $\gls{rRReward}_{\gls{iZyklus}}$} ([xshift=-0.25cm,yshift=0cm] Replay.west);
	
	\path [arrow, bend left=10] ([yshift=0.2cm]Koordinator.east) edge node[pos=0,right,xshift=0.05cm,yshift=0.37cm] {\scriptsize $\gls{rdRL}_{\gls{iZyklus}}$} ([xshift=-0.25cm,yshift=-0.4cm] Replay.west);

	\draw[arrow] (Policy.west) -- (stellgroessen.east)node[pos=1,right,xshift=0.05cm,yshift=0.2cm] {\scriptsize $\gls{ruCtrlVec}_{\gls{iZyklus}}$};

	\draw[arrow] (Zustand.north) -- (stellgroessen.south)node[midway,right,xshift=0cm,yshift=0cm] {\scriptsize $\gls{rxStateVec}_{\gls{iZyklus}-1}$};
	\draw[arrow] (Zustand.south) -- (reward.north)node[midway,right,xshift=0cm,yshift=0cm] {\scriptsize $\gls{rxStateVec}_{\gls{iZyklus}-1},\,\gls{rxStateVec}_{\gls{iZyklus}}$};

	\path [arrow, bend right=70] (KNNUpdate.east) edge node[pos=0.9,right]{\scriptsize $\gls{gthetaParam}_{\gls{gmRL}}$} (Policy.east);
	
	\draw[arrow](stellgroessen.west) -- (Aktorik.east) node[pos=0,left,xshift=0cm,yshift=0.2cm] {\scriptsize $\gls{ruCtrlVec}_{\gls{iZyklus},\gls{iSicher}}$};

	\draw[arrow] (Policy.south) -- (Replay.north)node[midway,right,xshift=0cm,yshift=0cm] {\scriptsize $\gls{ruCtrlVec}_{\gls{iZyklus}}$};

	\draw[arrow] (Replay.south) -- (KNNUpdate.north)node[pos=0.2,right,xshift=0cm,yshift=0cm] {\tiny Training batch};
	\newlength{\LinePosX}
	\pgfmathsetlength{\LinePosX}{(0.55*\totalwidth)}

	\draw[thick,dashed] (\LinePosX,0cm) -- (\LinePosX,-6cm) node[rotate=90,pos=0.05, draw=none, fill=white, rectangle, inner sep=2pt] {\scriptsize \gls{kUDP}}node[rotate=90,pos=0.95, draw=none, fill=white, rectangle, inner sep=2pt] {\scriptsize \gls{kUDP}};

	\draw[arrow] 
	([yshift=-0.8cm]Sensorik.east) -- 
	([yshift=-0.8cm, xshift=1.0cm]Sensorik.east) 
	|- 
	([xshift=-0.25cm,yshift=-0.7em]Zustand.west) 
	node[pos=1, sloped, xshift=-0.4cm, yshift=0.2cm] {\scriptsize $\gls{rUIon}$};
	
	\draw[arrow] 
	([yshift=0.8cm]Sensorik.east) -- 
	([yshift=0.8cm, xshift=0.70cm]Sensorik.east) 
	|- 
	([xshift=-0.25cm,yshift=0.7em]Zustand.west) 
	node[pos=1, sloped, xshift=-0.4cm, yshift=0.2cm] {\scriptsize $\pzyl$};


	\draw[very thick]([xshift=-0.25cm,yshift=0.5cm] Zustand.west)-- ([xshift=-0.25cm,yshift=-0.5cm] Zustand.west);
	\draw[arrow] ([xshift=-0.25cm] Zustand.west)-- (Zustand.west);

	\draw[very thick]([xshift=-0.25cm,yshift=0.5cm] Replay.west)-- ([xshift=-0.25cm,yshift=-0.5cm] Replay.west);
	\draw[arrow] ([xshift=-0.25cm] Replay.west)-- (Replay.west);

	\draw[arrow] (Aktorik.south)-- (Prozess.north);
	\draw[arrow] (Prozess.south)-- (Sensorik.north);
	
	\draw[arrowb] ([yshift=-0.2cm]Koordinator.east)-- ([yshift=-0.2cm, xshift=2.22cm]Koordinator.east)node[pos=0.1, right, xshift=0cm,yshift=0.17cm]{\scriptsize Sync};


    \begin{axis}[%
	width=2.0cm,
	height=1cm,
	at={(Sensorik.center)},
	xshift=-1cm,              
	yshift=0cm,
	scale only axis,
	xmin=-450,
	xmax=120,
	xtick=\empty,
	ytick=\empty,
	xticklabels=\empty,
	yticklabels=\empty,
	ymin=-5,
	ymax=50,
	axis background/.style={fill=white},
	axis lines=left,  
	]
	\addplot[color=mmpBlack,line width=1pt,forget plot] table [x index=0, y index=1, col sep=comma] {Figures/Data/VerlaufeRLSchema/VerlaufeRLSchema.csv};
	\node[anchor=east,draw=none,fill=none] at (axis cs: 5,30) {\scriptsize$\pzyl$};
	
	\end{axis}
	
	\begin{axis}[%
		width=2.0cm,
		height=1cm,
		at={(Sensorik.center)},
		xshift=-1cm,              
		yshift=-1.3 cm,
		scale only axis,
		xmin=-450,
		xmax=120,
		xtick=\empty,
		ytick=\empty,
		xticklabels=\empty,
		yticklabels=\empty,
		ymin=-1,
		ymax=10,
		axis background/.style={fill=white},
		axis lines=left,  
		]
		\addplot[color=mmpBlack,line width=1pt,forget plot] table [x index=0, y index=2, col sep=comma] {Figures/Data/VerlaufeRLSchema/VerlaufeRLSchema.csv};
		\node[anchor=east,draw=none,fill=none] at (axis cs: 5 ,6) {\scriptsize$\gls{rUIon}$};
	\end{axis}

\end{tikzpicture}

%% file: Chapters/Fundamentals.tex
\section{Problem Formulation}
\subsection{Definition of the State-Action-Space}
\label{cha:SASpace}
A requirement for applying an~MDP is fulfilling the Markov property, which states that transition probabilities $\gls{rP}\left(\gls{rxStateVec}_{\gls{iZyklus}} \mid \gls{rxStateVec}_{\gls{iZyklus}-1}, \gls{ruCtrlVec}_{\gls{iZyklus}}\right)$ depend solely on the current state $\gls{rxStateVec}_{\gls{iZyklus}-1}$ and not on previous ones. Consequently, the current state must capture all relevant information for predicting future states. Prior research has demonstrated, via partial autocorrelation, that the HCCI process memory in stable operation spans only one combustion cycle~\citep{StuartDaw.2007, Andert.2018}. Therefore, it is assumed that \gls{kHCCI} fulfills the Markov property in stabilized, closed-loop operation. Thus, for state description, it is sufficient to use cycle~$\gls{iZyklus}-1$ to determine actions for cycle~$\gls{iZyklus}$.

Prior studies indicate that the combustion phasing~$\gls{gAlpha50i1}$, $\gls{rpmii1}$ and heat release~$\gls{rQ}_{\gls{iZyklus}-1}$ adequately represent the current thermodynamic mixture state for the purpose of combustion control~\citep{Wick.2019, Nuss.2019, Bedei.2023}. Additionally, the maximum pressure gradient~$\gls{rDPMAXi1}$ is included, as it must be constrained to mitigate mechanical stress on the engine and improve acoustic behavior. In addition to these pressure-based variables, features of the ion current --~the maximum $\gls{rIonMax}$ and integral $\gls{rIonIntegral}$~-- are used, as they have been shown to enhance control performance in the literature~\citep{Bedei.2023b}. Finally, the load setpoint for cycle~$\gls{iZyklus}$, $\gls{rpmiiSoll}$, is provided to the agent to address transient load control, while the previous cycle’s target load~$\gls{rpmii1Soll}$ supplies information on any load steps in the current cycle.

The action space includes adjusting the NVO duration through the angle interval  $\gls{gAlpha}_{\gls{iNVO},\gls{iZyklus}}$, which specifically controls the amount of fresh air and residual gas fraction. Additionally, both gasoline~$\gls{rt}_{\gls{iKr},\gls{iInj},\gls{iZyklus}}$ and ethanol $\gls{rt}_{\gls{iEth},\gls{iInj},\gls{iZyklus}}$ injection durations are applied, allowing the engine's power output to be distributed between the two fuels:
\begin{equation}
	\begin{array}{cc}
		\gls{rxStateVec}_{\gls{iZyklus}-1}=\begin{pmatrix}
			\gls{gAlpha50i1}\\
			\gls{rQ}_{\gls{iZyklus}-1}\\
			\gls{rpmii1}\\
			\gls{rDPMAXi1}\\
			\gls{rIonMaxi1}\\
			\gls{rIonIntegrali1}\\
			\gls{rpmii1Soll}\\
			\gls{rpmiiSoll}\\
		\end{pmatrix}
	\end{array}
	\quad\quad 
	\begin{array}{cc}
		\gls{ruCtrlVec}_{\gls{iZyklus}}=\begin{pmatrix}
			\gls{gAlpha}_{\gls{iNVO},\gls{iZyklus}}\\
			\gls{rt}_{\gls{iKr},\gls{iInj},\gls{iZyklus}}\\
			\gls{rt}_{\gls{iEth},\gls{iInj},\gls{iZyklus}}
		\end{pmatrix} 
	\end{array}
\end{equation}
\subsection{Reward Function}
\label{cha:RewardFunction}
Defining an effective reward function is a key challenge in \gls{kRL}, as it provides feedback on the quality of the agent’s actions and guides it toward an optimal policy. Thus, a well-designed reward function can improve training efficiency and accelerate convergence~\citep{Hu.2020}. Beyond the primary goal of precise load tracking, additional objectives like safety, efficiency and minimizing process fluctuations are incorporated. 

Quadratic terms, commonly used in \gls{kMPC}~\citep{Gordon.2024}, were initially considered for the evaluation of the objectives. However, they proved unsuitable for RL in the HCCI environment. Specifically, quadratic functions create large error gradients when the agent is far from the target, potentially destabilizing training. Additionally, close to the target, the small reward gradients provide only minimal motivation for the agent to further optimize its policy, potentially leading to suboptimal solutions.

To address these issues, a modified reward function with reward clipping, which can stabilize training and improve policy performance~\citep{Mnih.2015, Schaul.2021}, is employed. Specifically, we use the hyperbolic tangent function to limit output values to the interval $\left[-1,1\right]$. However, this leads to saturation and small gradients far from the target, which may prevent the agent from further improving its policy. To mitigate this, a moderate linear term is introduced to prevent zero gradients while avoiding excessive rewards. The resulting function $r_{f}$  is applied to all reward components:
\begin{equation}
	\label{eq:Belohnungsterm}
	r_{f} = \min(\tanh(C_1 \cdot f + C_2) \cdot C_3 + C_4 \cdot f + C_5, 0)
\end{equation}
Here, $f$ represents an evaluation metric for each objective. Constants $C_1$ to $C_5$ allowing prioritization among objectives. The parameters are manually tuned through iterative adjustments during testbench experiments to achieve the desired agent behavior. Table~\ref{tab:Rewardwerte} provides the evaluation metric $f$ and the corresponding parameters used for each objective.
\begin{table*}[htbp]
	\centering
	\caption{Objectives and Parameterization of the Reward Function.}
	\label{tab:Rewardwerte}
	\begin{tabular}{lcccccc}
		\toprule
		Objective & \(f\)& \(C_1\) & \(C_2\) & \(C_3\) & \(C_4\) & \(C_5\) \\
		\midrule
		Load tracking & $\left(\gls{rpmi}-\gls{rpmiSoll}\right)^2$   &$3$  &$0$  &$-1.5$  & $-0.1$ &0\\
		Stability & $\left(\Delta \gls{gAlpha50}\right)^2$  &$0.015$  &$0$  &$-0.5$  &$-5 \cdot 10^{-4}$ &$0$\\
		Pressure gradient limitation&$\Delta\gls{rDPMAX}$ &$20$&$-2$&$-0.25$&$-1$&$-0.241$\\
		Safe actions & $\Delta \gls{rr}_{\gls{iSafetyFilter}}$  &$-7$  &$-2$  &$-0.25$  &$0.4$ &$-0.241$\\
		Efficiency & $\gls{geta}$ & $0$  & $0$  &$0$  &$-5 \cdot 10^{-3}$  &$-0.2$\\
		Ethanol energy share & $\left(\Delta \gls{rxAnteil}_{\gls{rEEnergie}_{\gls{iEth}}} \right)^2$ & $100$ & $0$ &$-0.75$ &$ -10$ &$0$\\
		\bottomrule
	\end{tabular}
\end{table*}
In the following the objectives are introduced in detail.

\textbf{Load Tracking:} To achieve the primary objective of load tracking, the control error is explicitly incorporated into the reward function.  The evaluation metric for the load setpoint is the quadratic control deviation: $f_{\text{Load}}=\left(\Delta \gls{rpmi}\right)^2=\left(\gls{rpmi}-\gls{rpmiSoll}\right)^2$. 

\textbf{Stability:} To enhance process stability combustion phasing~\gls{gAlpha50} variance need to be minimized. No explicit target is set for the phasing, instead, stability is ensured by minimizing the change of the phasing from cycle to cycle, defined as $f_{\text{Stability}}=\left(\Delta \gls{gAlpha50}\right)^2=\left(\gls{gAlpha50i}-\gls{gAlpha50i1}\right)^2$.

\textbf{Safety Aspects:} Two safety criteria are included in the reward function. First, the pressure gradient is constrained by a limit of $\gls{rDPMAXLim}=5\,\si{\bar\per\degreeKW}$, incorporated in the reward using $f_{\text{Safe, Gradient}}=\Delta\gls{rDPMAX}=\gls{rDPMAX}-\gls{rDPMAXLim}$. Second, state-dependent action space limitations are considered. A safety monitoring method introduced in Section~\ref{chap:Safety Monitoring}, determines the distance $f_{\text{Safe, Monitor}}=\Delta \gls{rr}_{\gls{iSafetyFilter}}$ from the safe action space. No penalty applies if the actions taken by the agent are within safe limits; otherwise, penalties increase with the distance of the chosen actions from the safe range. Unsafe actions are replaced by the safety monitoring to prevent potentially harmful actions from being applied to the testbench environment. These adjustments are penalized with a magnitude comparable to that of pressure gradient violations, discouraging the agent from taking potentially harmful actions.

\textbf{Efficiency:} The system's thermal efficiency, $f_{\text{Efficiency}}=\gls{geta}$, is evaluated by considering the contributions from injected masses of both gasoline ($\gls{rm}_{\gls{iKr},\gls{iInj}}$) and ethanol ($\gls{rm}_{\gls{iEth},\gls{iInj}}$) and using the lower calorific values (LCV) of both fuels:
\begin{equation}
	\gls{geta}=\frac{\gls{rpmi}\cdot\gls{rVH}}{\gls{rm}_{\gls{iKr},\gls{iInj}}\cdot\gls{rHUKr}+\gls{rm}_{\gls{iEth},\gls{iInj}}\cdot\gls{rHUEth}}
\end{equation}

\textbf{Ethanol Energy Share:} For the online adaptation of the agent's policy performed in Section~\ref{cha:OnlineAdaption}, a target ethanol energy share is defined. The squared deviation $f=\left(\Delta \gls{rxAnteil}_{\gls{rEEnergie}_{\gls{iEth}}} \right)^2$ is then incorporated into the reward. The ethanol energy share is calculated as:
\begin{equation}
	\gls{rxAnteil}_{\gls{rEEnergie}_{\gls{iEth}}}=\frac{\gls{rm}_{\gls{iEth},\gls{iInj}}\cdot\gls{rHUEth}}{\gls{rm}_{\gls{iKr},\gls{iInj}}\cdot\gls{rHUKr}+\gls{rm}_{\gls{iEth},\gls{iInj}}\cdot\gls{rHUEth}}
\end{equation}

\textbf{Total Reward:} The total reward is the sum of all reward components~$r_{f}$ :
\begin{equation}
	\label{eq:RewardOverall}
	\gls{rRReward}=\gls{rRReward}_{\gls{rpmiSoll}}+\gls{rRReward}_{\Delta\gls{gAlpha50}}+\gls{rRReward}_{\gls{rDPMAX}}+\gls{rRReward}_{\Delta\gls{rr}_{\gls{iSafetyFilter}}}+\gls{rRReward}_{\gls{geta}}+\gls{rRReward}_{\Delta \gls{rxAnteil}_{\gls{rEEnergie}_{\gls{iEth}}}}
\end{equation}

The reward terms (Table~\ref{tab:Rewardwerte}) are weighted as follows: For the safety criteria ($\gls{rRReward}_{\gls{rDPMAX}},\gls{rRReward}_{\Delta\gls{rr}_{\gls{iSafetyFilter}}}$), the highest weights are assigned to minimize the risk of damage in the testbench environment. The deviation from the IMEP setpoint ($\gls{rRReward}_{\gls{rpmiSoll}}$) is weighted relatively high to ensure accurate load tracking, followed by stability ($\gls{rRReward}_{\Delta\gls{gAlpha50}}$) with a slightly lower weight, while system efficiency ($\gls{rRReward}_{\gls{geta}}$) receives the least weight. The reward for the ethanol share is used only during the online adaptation of the agent's policy in Section~\ref{cha:OnlineAdaption} and is set relatively high to encourage the agent to modify its policy.
\subsection{Safety Monitoring}
\label{chap:Safety Monitoring}
One major challenge in implementing RL in real-world environments is ensuring safety. For HCCI engines, exceeding defined pressure rise rate limits risks damaging mechanical parts or degrading acoustic behavior. Additionally, frequent misfires, while not directly harmful, disrupt continuous testbench operation and must be avoided. To reduce the likelihood of misfires, an IMEP deviation of up to $0.3\,\si{\bar}$ below the target load is tolerated. Deviations beyond this threshold are far off the load tracking objective and are thus considered out of bounds to minimize misfiring.

It is crucial to prevent unsafe actions, that could lead to high pressure rise rates or misfires, from being applied to the real-world environment. Instead, these actions must be replaced with safe ones and the agent receives a penalty to guide it toward the safe region. Since the experimental space limitations are unknown beforehand, RL cannot be safely applied directly to the real engine without first identifying these boundaries.

\begin{figure}[!htpb]
	\centering
	\input{Figures/FlowChartLimitation.tex}
	\caption{Dynamic Measurement Algorithm with Self-Learning of Experimental Space Limitations.}
	\label{fig:AlgoLimits}
\end{figure}
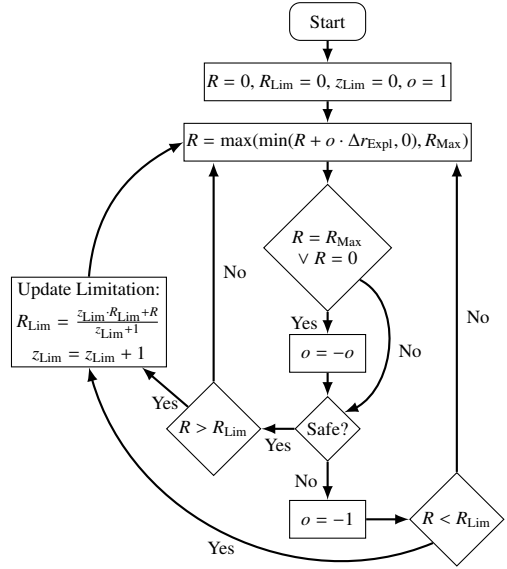

To achieve this, a dynamic measurement algorithm, initially described in~\citep{Wick.2020}, is extended to automatically learn the experimental space limitations required for safety monitoring. This algorithm is designed to generate highly dynamic data with substantial variance while ensuring that the actions chosen by the measurement algorithm remain safe. The goal is to prevent the exploration of unsafe regions while maximizing the coverage of the experimental space. Due to the autoregressive nature of HCCI, the limitations are highly state-dependent, which is addressed by classifying similar combustion cycles based on cycle integral parameters, such as the combustion phasing~\gls{gAlpha50i1}. After classification the algorithm is applied separately for each class~\gls{iKlasse}. Figure~\ref{fig:AlgoLimits} shows the algorithmic approach. 

Starting from a stable, safe starting point~$\gls{ruCtrlVec}_{\gls{iStart}}$ --~chosen dependent on the current load setpoint~-- the algorithm gradually increases the variance by exploring the action space along predefined direction vectors~$\gls{rRichtungsvektor}_{\gls{iRichtung}}$, extending the distance from the starting point in increments of $\gls{rrDeltaExpl}$. The algorithm is executed within a normalized action space, where each action is mapped from $[\gls{ruCtrl}_{\gls{imin}},\gls{ruCtrl}_{\gls{imax}}]$ to $[-1,1]$ using min-max normalization. Thus, cycle individual actions are given by: $\gls{ruCtrlVec}_{\gls{iNorm}}=\gls{ruCtrlVec}_{\gls{iStart}}+\rkl\cdot\gls{rRichtungsvektor}_{\gls{iRichtung}}$. The algorithm uses a total of four matrices to store progress separately for each direction \( l \) and class \( k \):

\begin{enumerate}
	\item The position matrix \( R \) records the current distance from the starting point.
	\item The limitation matrix \( R_{\text{Lim}} \) stores the maximum allowed distance from the starting point.  
	\item The counter matrix \( Z_{\text{Lim}} \) contains the counter \( z_{k,l,\text{Lim}} \) that is incremented with each limit update. Thus, the larger the value, the more reliable is the determined limitation.
	\item The orientation matrix \( O \) indicates the direction of exploration, where \( o_{k,l} = 1 \) means moving away from and \( o_{k,l} = -1 \) moving toward the starting point. Upon reaching a safe limit or a limit of the action range, the algorithm reverses the orientation.
\end{enumerate}

Throughout exploration, the algorithm continuously monitors for violations of boundaries, i.e. exceeding the pressure gradient limit~\gls{rDPMAXLim} or misfiring. When violations occur, the limitation matrix is updated. However, given the stochastic nature of the process, limits cannot be precisely set after a single violation. Instead, the matrix entries \mySymbol{\gls{rr}}{\gls{iKlasse},\gls{iRichtung},\gls{iLim}} are iteratively refined, gradually converging to positions where the likelihood of limitation violations is minimized. Upon completion of the algorithm, the limitation matrix~\mySymbol{\gls{rRVermessung}}{\gls{iLim}} defines a safe action space, which serves as the basis for the safety monitoring function. The data generated during this process is not discarded but can be potentially utilized for subsequent offline training phases --~for example, by loading it into the experience replay buffer before online training. 

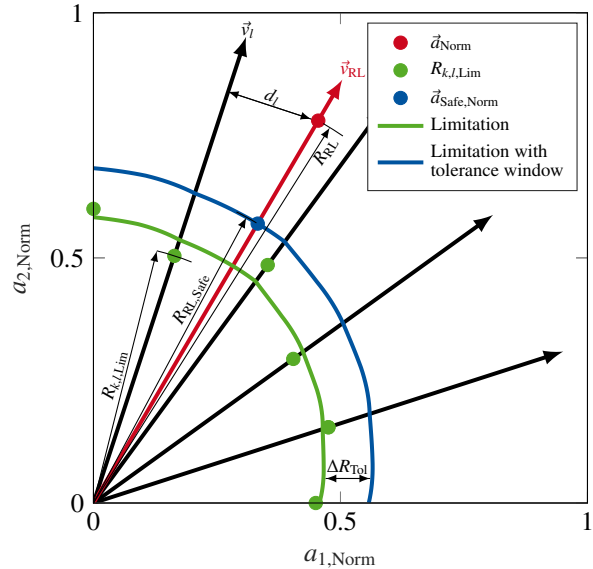
\begin{figure}[!htpb]
	\centering
	\input{Figures/SafetyFilterPlot}
	\caption{K-Nearest-Neighbor Based Safety Monitoring Principle with Replacement of the Actions Taken by the Agent $u_\text{A,Norm}$ with a Safe Point $u_\text{A,Safe,Norm}$.}
	\label{fig:SafeteyFilter}
\end{figure}

Due to hard real-time requirements and limited time for safety monitoring, a method with minimal computational effort is needed. The k-nearest neighbors algorithm is ideal, requiring little programming effort and computation time.
The algorithm allows new points to be evaluated relative to previously known ones. In this way, the actions selected by the agent can be verified using the limitation matrix.
The principle of the developed k-nearest neighbors based safety monitoring, is depicted in Figure~\ref{fig:SafeteyFilter} using a two-dimensional action space, though the methodology, including all equations, is applicable to higher dimensional spaces as well.

First, the real-valued action vector of the agent, $\mySymbol{\gls{ruCtrlVec}}{\gls{iRoh}}$, is mapped to the relevant range for each individual action $j$ using the hyperbolic tangent function:
\begin{equation}
	\mySymbol{\gls{ruCtrl}}{\gls{ij}}=\mySymbol{\gls{ruCtrl}}{\gls{imin},\gls{ij}}+\frac{\tanh(\mySymbol{\gls{ruCtrl}}{\gls{iRoh},\gls{ij}})+1}{2}\cdot(\mySymbol{\gls{ruCtrl}}{\gls{imax},\gls{ij}}-\mySymbol{\gls{ruCtrl}}{\gls{imin},\gls{ij}})
\end{equation}  

To compare these actions $\gls{ruCtrlVec}$ to the limitation matrices, the vector is normalized using the load-dependent start point $\gls{ruCtrl}_{\gls{iStart}}$ and the allowed action range $[\gls{ruCtrl}_{\gls{imin}}, \gls{ruCtrl}_{\gls{imax}}]$:
\begin{equation}
	\label{eq:URLNorm}
	\gls{ruCtrl}_{\gls{iNorm},{\gls{ij}}}=
	\begin{cases}
		\,\,\,\,\frac{\gls{ruCtrl}_{\gls{ij}}-\gls{ruCtrl}_{{\gls{iStart}, \gls{ij}}}}{\gls{ruCtrl}_{\gls{imax},{\gls{ij}}}-\gls{ruCtrl}_{{\gls{iStart}, \gls{ij}}}}&\text{if}\quad \gls{ruCtrl}_{\gls{ij}} \geq \gls{ruCtrl}_{{\gls{iStart}, \gls{ij}}}\\
		-\frac{\gls{ruCtrl}_{\gls{ij}}-\gls{ruCtrl}_{{\gls{iStart}, \gls{ij}}}}{{\gls{ruCtrl}_{\gls{imin},{\gls{ij}}}}-\gls{ruCtrl}_{{\gls{iStart}, \gls{ij}}}}&\text{if}\quad \gls{ruCtrl}_{\gls{ij}} < \gls{ruCtrl}_{{\gls{iStart}, \gls{ij}}}
	\end{cases}
\end{equation}

From this, the distance $\mySymbol{\gls{rd}}{\gls{iRichtung}}$ from the   direction~$\mySymbol{\gls{rRichtungsvektor}}{\gls{iRL}}$ of the agent's actions $\gls{ruCtrlVec}_{\gls{iNorm}}$ to each of the predefined directions  is calculated.
\begin{equation}
	\mySymbol{\gls{rd}}{\gls{iRichtung}} = \left\| \mySymbol{\gls{ruCtrlVec}}{\gls{iNorm}} - \frac{\mySymbol{\gls{ruCtrlVec}}{\gls{iNorm}} \cdot \mySymbol{\gls{rRichtungsvektor}}{\gls{iRichtung}}}{{\|\mySymbol{\gls{rRichtungsvektor}}{\gls{iRichtung}}\|^2}} \cdot \mySymbol{\gls{rRichtungsvektor}}{\gls{iRichtung}} \right\|
\end{equation}

A weight $\mySymbol{\gls{rw}}{\gls{iRichtung}}$ for each of the $\gls{rnumber}$ nearest neighbors is defined as:
\begin{equation}
	\mySymbol{\gls{rw}}{\gls{iRichtung}} = \frac{\frac{\mySymbol{\gls{rd}}{\gls{imax}}- \mySymbol{\gls{rd}}{\gls{iRichtung}}}{\mySymbol{\gls{rd}}{\gls{imax}}-\mySymbol{\gls{rd}}{\gls{imin}}}\cdot\mySymbol{\gls{rz}}{\gls{iKlasse},\gls{iRichtung},\gls{iLim}}}{\sum_{j=1}^{\gls{rnumber}}\frac{\mySymbol{\gls{rd}}{\gls{imax}}-\mySymbol{\gls{rd}}{j}}{\mySymbol{\gls{rd}}{\gls{imax}}-\mySymbol{\gls{rd}}{\gls{imin}}}\cdot \gls{rz}_{\gls{iKlasse},j,\gls{iLim}}}
\end{equation}

where $\mySymbol{\gls{rd}}{\gls{imax}}$ is the largest and $\mySymbol{\gls{rd}}{\gls{imin}}$ the smallest distance of $\gls{ruCtrlVec}_{\gls{iNorm}}$ to the $\gls{rnumber}$ closest directions. In addition, the weight is also influenced by the counter $\mySymbol{\gls{rz}}{\gls{iKlasse},\gls{iRichtung},\gls{iLim}}$, giving greater consideration to limits that have been more accurately determined by the measurement algorithm.

From those weights and the limitation matrix entries $\mySymbol{\gls{rr}}{\gls{iKlasse},\gls{iRichtung},\gls{iLim}}$  a maximum allowed distance to the starting point $\mySymbol{\gls{rr}}{\gls{iSicher}}$ in the agent's direction is calculated:
\begin{equation}	
	\mySymbol{\gls{rr}}{\gls{iSicher}}=\Delta \mySymbol{\gls{rr}}{\gls{iTol}}+ \sum_{\gls{iRichtung}=1}^{\gls{rnumber}}\mySymbol{\gls{rw}}{\gls{iRichtung}} \cdot \mySymbol{\gls{rr}}{\gls{iKlasse},\gls{iRichtung},\gls{iLim}}
\end{equation}

Hereby, a tolerance window $\Delta \gls{rRVermessung}_{\gls{iTol}}$ is used to account for the process stochasticity, uncertainties in the limitation matrices and the approximation of the limits by interpolation.
The tolerance window allows minor limitation violations, which is acceptable as the limits were set conservatively. This enables the agent to learn the boundaries itself, while large, potentially harmful violations are prevented by the safety monitoring.
It was heuristically found that $\Delta \gls{rRVermessung}_{\gls{iTol}}=0.15$ leads to a good compromise between exploration capability of the agent and safety for the SCRE.

Finally, safe normalized actions $\mySymbol{\gls{ruCtrlVec}}{\gls{iSicher},\gls{iNorm}}$ are calculated as $ \mySymbol{\gls{ruCtrlVec}}{\gls{iSicher},\gls{iNorm}}=\mySymbol{\gls{rr}}{\gls{iSicher}} \cdot \mySymbol{\gls{rRichtungsvektor}}{\gls{iRL}}$. In case of a violation i.e. $\left\|\gls{ruCtrlVec}_{\gls{iNorm}}\right\|>\left\|\gls{ruCtrlVec}_{\gls{iSicher},\gls{iNorm}}\right\|$  the agent's actions are replaced with the safe actions $\mySymbol{\gls{ruCtrlVec}}{\gls{iSicher},\gls{iNorm}}$, which are then denormalized and applied to the engine. Otherwise the actions selected by the agent are applied.

In case of  a violation, for the safety monitoring reward $\gls{rRReward}_{\Delta\gls{rr}_{\gls{iSafetyFilter}}}$, a penalty is applied based on the distance of the agent's action from the safe region: 
\begin{equation}
	\label{eq:fSftFlt}
	\Delta \mySymbol{\gls{rr}}{\gls{iSafetyFilter}}=\min\left(\left\| \mySymbol{\gls{ruCtrlVec}}{\gls{iNorm}}\right\|-\left\| \mySymbol{\gls{ruCtrlVec}}{\gls{iSicher},\gls{iNorm}}\right\|,0\right)
\end{equation}

The term increases proportionally with the distance of the actions taken by the agent to the tolerated safe limit. It is incorporated into the total reward (Equation~\ref{eq:RewardOverall}) using Equation~\ref{eq:Belohnungsterm} and the parameters from Table~\ref{tab:Rewardwerte}. 

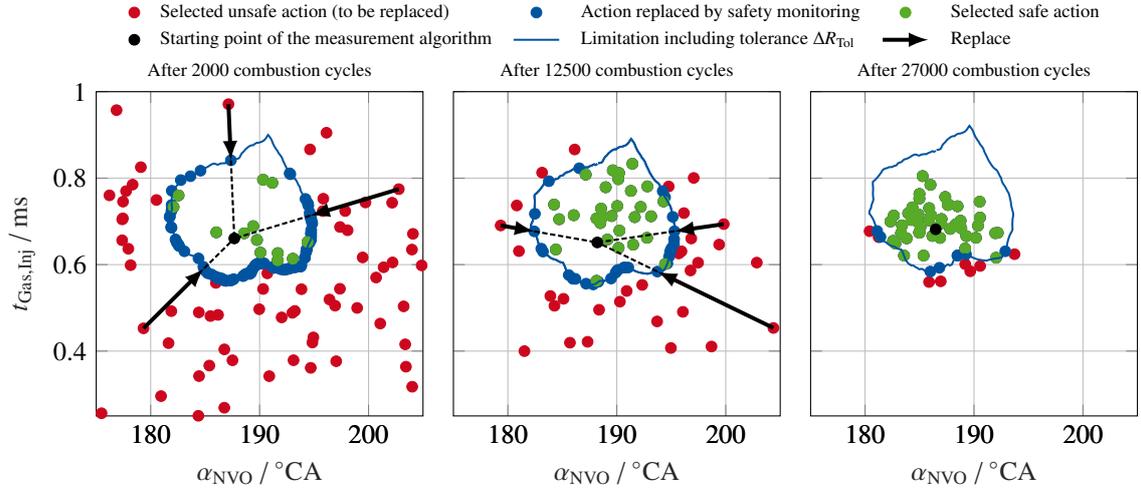
\begin{figure*}[!htpb]
	\centering
	\input{Figures/LimitierungRL.tex}
	\caption{Replacing the Actions Selected by the Agent by Using the Safety Monitoring at Various Stages of the Training.}
	\label{fig:FilterRL}
\end{figure*}

Figure~\ref{fig:FilterRL} illustrates the impact of safety monitoring and unsafe action replacement during an \gls{kRL} training in the real-world HCCI testbench environment. The two actions considered are the  \gls{kNVO} duration~$\gls{gAlpha}_{\gls{iNVO}}$ and the gasoline injection duration~$\gls{rt}_{\gls{iKr},\gls{iInj}}$.
The cycles from a class with $\SI{3}{\degreeKW}<\gls{gAlpha50i1}\leq\SI{9}{\degreeKW}$ are shown for a load setpoint of  $\gls{rpmiSoll}=\SI{3}{\bar}$ at various training stages. Initially (left), $88{.}5\,\%$ of the agent's selected points (red) fall outside the safe area. These are replaced and set to the boundary of the safe action space (blue points). The replacement takes place in the direction of the starting point of the measurement algorithm, as shown by the arrows. Meanwhile, $11{.}5\,\%$ of the agent’s actions (green) already fall within the safe limits and are applied without changes. Through penalization for exceeding safe boundaries, the agent implicitly learns to stay within the safe region. Thus, after $12,500$ training combustion cycles (middle), the percentage of unsafe points decreases significantly, though $53.2\,\%$ still exceed the limits. By $27,000$ cycles (right), only $12{.}9\,\%$  remain outside the limits, with smaller distances to the safe region. This demonstrates the effectiveness of the safety monitoring in conjunction with the penalization of boundary violations. Additionally, the agent no longer selects actions in the upper region of the safe space, likely due to the penalization of other objectives in that region.
\subsection{Boundary Conditions for Real-Time Execution}
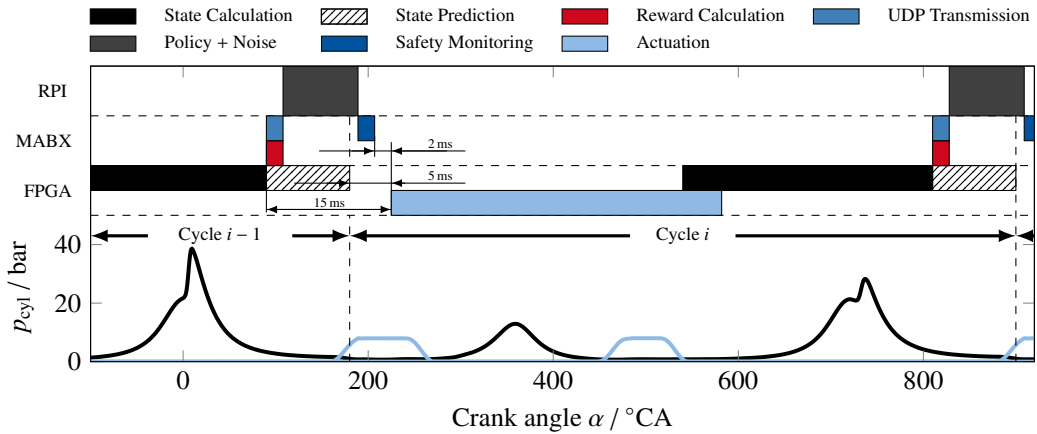
\begin{figure*}[!htpb]
	\centering
	\input{Figures/BerechnugAblaufRLWide.tex}
	\caption{Real-Time Boundary Conditions for the RL Toolchain in the Testbench Environment at $\gls{rNMot}=1,500~\si{\per\minute}$.}
	\label{fig:RTRandbedingungenRL}
\end{figure*}

To ensure real-time execution, the latencies in data transfer between the \gls{kFPGA}, \gls{kMABX} processor and \gls{kRPI} must be considered. Additionally, as the \gls{kRPI} is not a real-time system, completion of calculations within a fixed time frame cannot be guaranteed. It was determined that computation of the policy on the \gls{kRPI} and data transmission to the \gls{kMABX} typically fall within a \SI{3}{\milli\second} time window, achieved through overclocking the RPI to $2.2$ GHz. Due to the \gls{kRPI}'s non-real-time nature, a safety factor of $3$ is applied, extending the maximum available window to \SI{9}{\milli\second}, which covers over $99.9\,\%$ of combustion cycles.  In the rare case that the policy computation is not completed on time, the safety monitoring implemented on the real-time system acts as the final fallback. Figure~\ref{fig:RTRandbedingungenRL} shows the execution times for the computations across the three units.

After determining the state, the reward function is calculated on the \gls{kMABX}, with the state and reward then sent to the \gls{kRPI} using a task with \SI{1}{\milli\second} clock rate. The \gls{kRPI} calculates the actions (Equation~\ref{eq:Policy})  and sends them to the \gls{kMABX} for safety monitoring. The safe actions $\gls{ruCtrlVec}_{\gls{iSicher}}$ are then transmitted to the \gls{kFPGA} for actuation. 

Altogether, this process requires a time window of \SI{13}{\milli\second} to accommodate calculations, latencies and safety checks, meaning that the state calculation must be completed at approximately \SI{90}{\degreeKW}. However, the IMEP is typically integrated through the end of the expansion phase (\SI{180}{\degreeKW}), which would result in finishing the calculation too late to be real-time capable:
\begin{equation}
	\label{eq:IMEP}
	\gls{rpmi}=\frac{1}{\gls{rVH}}\cdot\int_\text{\SI{-540}{\degreeKW}}^{\SI{180}{\degreeKW}}\mySymbol{\gls{rp}}{\gls{iZyl}}\cdot\mathrm{d}\mySymbol{\gls{rV}}{\gls{iZyl}}
\end{equation}

To meet the real-time requirements, it is assumed that no further fuel conversion occurs after \SI{50}{\degreeKW} and that the subsequent pressure trace follows an isentropic process $\left(\mySymbol{\gls{rp}}{\gls{iZyl}} \cdot \mySymbol{\gls{rV}}{\gls{iZyl}}^{\gls{gKappa}} = \text{constant}\right)$. This assumption allows the integral during the expansion phase to be predicted already at \SI{50}{\degreeKW}, so the state calculation can be completed once the cylinder pressure at \SI{50}{\degreeKW} is known. This ensures to provide the current state on time.

%% file: Figures/FlowChartLimitation.tex
\begin{tikzpicture}[>=latex, node distance=2cm]
	
	\tikzstyle{startstop} = [rectangle, rounded corners, minimum width=1cm, minimum height=0.5cm,text centered, draw=black]
	\tikzstyle{process} = [rectangle, minimum width=1cm, minimum height=0.5cm, text centered, draw=black, align=center, ,inner sep=1pt]
	\tikzstyle{decision} = [diamond, minimum width=0.5cm, minimum height=0.5cm, text centered, draw=black,inner sep=1pt]
	
	\tikzstyle{arrow} = [thick,->,>=latex]
	
	\node (start) [startstop] at (0,0) {\scriptsize Start};
	\node (init) [process, below of=start, yshift=1.2cm] {\scriptsize $\gls{rr}=0$, $\mySymbol{\gls{rr}}{\gls{iLim}}=0$, $\mySymbol{\gls{rz}}{\gls{iLim}}=0$, $\gls{ro}=1$};
	\node (updateRadius) [process, below of=init, yshift=1.2cm] {\scriptsize $\gls{rr}=\max(\min(\gls{rr}+\gls{ro}\cdot\gls{rrDeltaExpl},0),\mySymbol{\gls{rr}}{\gls{imax}})$};
	\node (decision1) [decision, below of=updateRadius, yshift=0.6cm] {\begin{minipage}{1.1cm}\centering\scriptsize $\gls{rr}=\mySymbol{\gls{rr}}{\gls{imax}}$ \\ \scriptsize {$\vee~\gls{rr}=0$} \end{minipage}};
	\node (DirUpdate) [process, below of=decision1, yshift=0.6cm] {\scriptsize $\gls{ro}=-\gls{ro}$};
	\node (decision2) [decision, below of=DirUpdate, yshift=1cm] {\scriptsize Safe?};
	\node (to_be_filled3) [process, below of=decision2, yshift=0.8cm] {\scriptsize $\gls{ro}=-1$};	
	\node (decision3) [decision, right of=to_be_filled3, xshift=-0.3cm] {\scriptsize $\gls{rr}<\mySymbol{\gls{rr}}{\gls{iLim}}$};
	\node (decision4) [decision, left of=decision2, xshift=0.5cm] {\scriptsize $\gls{rr}>\mySymbol{\gls{rr}}{\gls{iLim}}$};
	\node (to_be_filled4) [process, above left=0.5cm and 0.3cm of decision4, inner sep=2pt]{\scriptsize Update Limitation: \\[0ex]\scriptsize $\gls{rr}_{\gls{iLim}}=\frac{\gls{rz}_{\gls{iLim}}\cdot\gls{rr}_{\gls{iLim}}+\gls{rr}}{\gls{rz}_{\gls{iLim}}+1}$\\[-0ex]\scriptsize$\gls{rz}_{\gls{iLim}}=\gls{rz}_{\gls{iLim}}+1$};

	
	\draw [arrow] (start) -- (init);
	\draw [arrow] (init) -- (updateRadius);
	\draw [arrow] (updateRadius) -- (decision1);
	\draw [arrow] (decision1) -- node[anchor=east] {\scriptsize Yes} (DirUpdate);
	\draw [arrow] (DirUpdate) -- (decision2);
	\draw [arrow] (decision2) -- node[anchor=east] {\scriptsize No} (to_be_filled3);
	\draw [arrow] (to_be_filled3) -- (decision3);
	\draw [arrow] (decision2) -- node[anchor=north] {\scriptsize Yes}(decision4);
	\draw [arrow] (decision4) -- node[anchor=north] {\scriptsize Yes}(to_be_filled4);
	\draw [arrow] (decision4.north)-- ++(0,2.9cm)node[midway, anchor=west] {\scriptsize No};
	\draw [arrow] (decision3.north)-- ++(0,4.1cm)node[midway, anchor=west] {\scriptsize No};
	
	\path [arrow, bend left=50] (decision3.south west) edge node[anchor=north] {\scriptsize Yes} (to_be_filled4.south);
	\path [arrow, bend left=70] (decision1.south east) edge node[anchor=west] {\scriptsize No} (decision2.north east);
	\path [arrow, bend left=30] (to_be_filled4.north) edge node[anchor=west] {\scriptsize} (updateRadius.west);
\end{tikzpicture}

%% file: Figures/SafetyFilterPlot.tex
%
%
\definecolor{mycolor1}{rgb}{0.00000,0.32941,0.62353}%
\definecolor{mycolor2}{rgb}{0.85490,0.12157,0.23922}%
\definecolor{mycolor3}{rgb}{0.55686,0.72941,0.89804}%
\begin{tikzpicture}

	\begin{axis}[%
		width=6.5cm,
		height=6.5cm,
		at={(3.384in,0.445in)},
		scale only axis,
		xmin=0,
		xmax=1,
		xtick={ -1,   0, 0.5,   1, 1.5},
		xticklabels={{$-1$},{$0$},{$0.5$},{$1$},{$1.5$}},
		xlabel style={font=\color{white!15!black}},
		xlabel={\mySymbol{\gls{ruCtrl}}{1,\gls{iNorm}}},
		ymin=0,
		ymax=1,
		ytick={ -1,   0, 0.5,   1, 1.5},
		yticklabels={{$-1$},{$0$},{$0.5$},{$1$},{$1.5$}},
		ylabel style={font=\color{white!15!black}},
		ylabel={\mySymbol{\gls{ruCtrl}}{2,\gls{iNorm}}},
		axis background/.style={fill=white},
		legend style={legend cell align=left, align=left, draw=white!15!black}
		]

		\draw[line width=1.5pt,->,>=latex,color=black](0,0)to(0.309017,0.951057);
		\draw[line width=1.5pt,->,>=latex,color=black](0,0)to(0.587785,0.809017);
		\draw[line width=1.5pt,->,>=latex,color=black](0,0)to(0.809017,0.587785);
		\draw[line width=1.5pt,->,>=latex,color=black](0,0)to(0.951057,0.309017);

		\addplot[only marks, mark=*, mark options={}, mark size=2.5000pt, draw=mmpRed, fill=mmpRed] table[row sep=crcr]{%
			x	y\\
			0.455	0.78\\
		};
		\addlegendentry{\scriptsize\mySymbol{\gls{ruCtrlVec}}{\gls{iNorm}}}
		
		\draw[line width=1.5pt,->,>=latex,color=mmpRed](0,0)to(0.503871025524086,0.863778900898433);

		\node[above right, align=left, font=\color{black}]
		at (axis cs:0.28,0.93) {\scriptsize \mySymbol{\gls{rRichtungsvektor}}{\gls{iRichtung}}};

		\node[above right, align=left, font=\color{mmpRed}]
		at (axis cs:0.48,0.85) {\scriptsize \mySymbol{\gls{rRichtungsvektor}}{\gls{iRL}}};

		\addplot[only marks, mark=*, mark options={}, mark size=2.5000pt, draw=mmpGreen, fill=mmpGreen] table[row sep=crcr]{%
			x	y\\
			0.352671151375484	0.485410196624968\\
		};
		\addlegendentry{\scriptsize  \mySymbol{\gls{rr}}{\gls{iKlasse},\gls{iRichtung},\gls{iLim}}}
		
		\addplot[only marks, mark=*, mark options={}, mark size=2.5000pt, draw=mmpGreen, fill=mmpGreen, forget plot] table[row sep=crcr]{%
			x	y\\
			0.163779007018722	0.504059953636431\\
		};
		\addplot[only marks, mark=*, mark options={}, mark size=2.5000pt, draw=mmpGreen, fill=mmpGreen, forget plot] table[row sep=crcr]{%
			x	y\\
			0.404508497187474	0.293892626146237\\
		};
		\addplot[only marks, mark=*, mark options={}, mark size=2.5000pt, draw=mmpGreen, fill=mmpGreen, forget plot] table[row sep=crcr]{%
			x	y\\
			0	0.6\\
		};
		
		\addplot[only marks, mark=*, mark options={}, mark size=2.5000pt, draw=mmpGreen, fill=mmpGreen, forget plot] table[row sep=crcr]{%
			x	y\\
			0.475528258147577	0.154508497187474\\
		};
		
		\addplot[only marks, mark=*, mark options={}, mark size=2.5000pt, draw=mmpGreen, fill=mmpGreen, forget plot] table[row sep=crcr]{%
			x	y\\
			0.475528258147577	0.154508497187474\\
		};
		
		\addplot[only marks, mark=*, mark options={}, mark size=2.5000pt, draw=mmpGreen, fill=mmpGreen, forget plot] table[row sep=crcr]{%
			x	y\\
			0.45	0\\
		};
		
		\addplot[only marks, mark=*, mark options={}, mark size=2.5000pt, draw=mmpDarkBlue, fill=mmpDarkBlue] table[row sep=crcr]{%
			x	y\\
			0.332454188830202	0.56992146656606\\
		};
		\addlegendentry{\scriptsize \mySymbol{\gls{ruCtrlVec}}{\gls{iSicher},\gls{iNorm}}}

		
		\addplot [color=mmpGreen, line width=1.5pt]
		table[row sep=crcr]{%
			0	0.582962187048859\\
			0.0101559637192847	0.581834771801206\\
			0.0202756404633102	0.580618375780683\\
			0.0303599671436776	0.579302682903451\\
			0.0404092449765217	0.57787912609495\\
			0.050423264979718	0.576340555994171\\
			0.0604014053792016	0.574680984324964\\
			0.0703427080451518	0.572895383001592\\
			0.0802459390877643	0.570979525329366\\
			0.0901096373564214	0.568929859345924\\
			0.0998549768528546	0.566305714823942\\
			0.109523861807483	0.563451423106265\\
			0.119108601148455	0.56036191126088\\
			0.12860080066523	0.557031265495071\\
			0.137991218738263	0.5534525491108\\
			0.14726958516302	0.549617574237981\\
			0.156424371644615	0.545516613042168\\
			0.165442498310828	0.541138028808529\\
			0.17430895451649	0.536467799697632\\
			0.183459935608121	0.53280634061081\\
			0.192552680624814	0.529034142072134\\
			0.201584704686158	0.525146109787439\\
			0.210553495593315	0.52113818888386\\
			0.219456523494708	0.517007170070465\\
			0.228291248555272	0.51275053941565\\
			0.237055127133692	0.508366360680476\\
			0.245745616830937	0.503853182239744\\
			0.254360180675181	0.499209962775299\\
			0.26297017126527	0.49457496069525\\
			0.271491530075696	0.489783685408263\\
			0.279920782249339	0.48483701695028\\
			0.288254345980567	0.479735793713826\\
			0.296488506406451	0.474480794266052\\
			0.304619382653904	0.469072714647575\\
			0.312642885926305	0.463512139757521\\
			0.320554665729823	0.457799506913931\\
			0.328350040214345	0.451935058937007\\
			0.334417979860097	0.4437876484301\\
			0.340454864990344	0.435762355583815\\
			0.346450281866641	0.427830467957882\\
			0.352395208502565	0.419968255745918\\
			0.3582817494891	0.412156005496133\\
			0.364102930735552	0.404377271560891\\
			0.369852538934304	0.396618290076334\\
			0.375524994805247	0.38886751570688\\
			0.381115252132327	0.381115252132327\\
			0.386844958459526	0.373571833975073\\
			0.392369667705543	0.365890634478241\\
			0.397678711999418	0.358071520615432\\
			0.402760109429159	0.350114021648181\\
			0.40760028971275	0.342017252765721\\
			0.412183749993128	0.333779819486878\\
			0.416492619129699	0.32539969687217\\
			0.420506100853351	0.316874075390977\\
			0.42419975465098	0.308199162123377\\
			0.427495064277665	0.299335266554666\\
			0.43078326081743	0.290566978334509\\
			0.434048435452934	0.281874349798635\\
			0.437277083953506	0.273241048054828\\
			0.440457645010089	0.264653653236183\\
			0.443580142736919	0.256101114816333\\
			0.446635906929695	0.247574325914947\\
			0.449617352054204	0.23906578672588\\
			0.452517801079399	0.230569335999211\\
			0.454995454120765	0.22191611062402\\
			0.457210867287005	0.213200928807619\\
			0.459150619822591	0.204427026822212\\
			0.460799417678328	0.195597748128501\\
			0.462139703835183	0.186716560349106\\
			0.463151169506438	0.177787076758113\\
			0.463810135538058	0.168813083686826\\
			0.464088761970663	0.159798575763914\\
			0.463954027411433	0.150747801652556\\
			0.464227188644225	0.141928494735837\\
			0.46452856343284	0.133201422117535\\
			0.464837520091087	0.124552838120093\\
			0.465136728120222	0.115971611471227\\
			0.465411521715037	0.107448716147346\\
			0.46564940679985	0.0989768368530616\\
			0.465839675205132	0.090550060074988\\
			0.465973099769887	0.0821636297737885\\
			0.46604169123707	0.0738137524344837\\
			0.466077316496076	0.0655028950962092\\
			0.465886874060046	0.0572037152618896\\
			0.465461752837336	0.0489220015773895\\
			0.464792081994928	0.040664038071168\\
			0.463866465759382	0.0324367031180571\\
			0.46267165079612	0.0242475937554434\\
			0.461192105300102	0.0161051832005988\\
			0.459409481206861	0.00801902232310455\\
			0.4575280074997	0.000798537825924491\\
		};
		\addlegendentry{\scriptsize Limitation}
		\addplot [color=mmpDarkBlue, line width=1.5pt]
		table[row sep=crcr]{%
			0	0.682962187048859\\
			0.0119012043630131	0.681819541316845\\
			0.0237655901335603	0.680557458482592\\
			0.035593562767972	0.679165636378908\\
			0.0473848923509342	0.677635531120932\\
			0.0591388392544838	0.675960025803346\\
			0.0708542517059669	0.674133173861792\\
			0.0825296423856665	0.672149998165724\\
			0.0941632491837709	0.670006332203523\\
			0.105753083860444	0.667698693405438\\
			0.117219794619548	0.664786490125163\\
			0.128604761345138	0.661614141451031\\
			0.139899770230231	0.65817667133426\\
			0.151095906099617	0.654468271973595\\
			0.16218340829823	0.650482121738399\\
			0.173151489673272	0.646210156866888\\
			0.183988107226315	0.641642782636\\
			0.194679668783101	0.636768504404832\\
			0.205210653953985	0.631573451327148\\
			0.216016751053836	0.627358198170741\\
			0.226754694957381	0.623003404150725\\
			0.237421499640688	0.618504152437159\\
			0.248014154934906	0.613856574340539\\
			0.258529636343636	0.609057655415709\\
			0.268964912862852	0.60410508517991\\
			0.279316953307762	0.598997139384141\\
			0.289582731509845	0.593732586869661\\
			0.299759230649136	0.588310615194136\\
			0.309917327543859	0.582869719981143\\
			0.319972492100329	0.577245656122202\\
			0.329920782249339	0.571439557328724\\
			0.339758153471573	0.565452523784037\\
			0.349480432829771	0.559285603881694\\
			0.359083286155407	0.552939771442117\\
			0.368562176273379	0.546415897013025\\
			0.377912309364927	0.53971471134283\\
			0.387128565443592	0.532836758374502\\
			0.394599482175301	0.523651199434829\\
			0.40202101252291	0.514563430944487\\
			0.409382320971625	0.505545064103579\\
			0.416673969471219	0.496572700057816\\
			0.423887652388151	0.48762696351841\\
			0.431015991371438	0.47869175410863\\
			0.438052374940554	0.469753660238251\\
			0.444990831851147	0.460801495740745\\
			0.451825930250981	0.451825930250982\\
			0.458778938493391	0.443037671020973\\
			0.465505037867461	0.43409047048449\\
			0.471993194547157	0.424984581251317\\
			0.478231067451437	0.415719924547231\\
			0.484204734024647	0.406296013734375\\
			0.489898346138825	0.396711858591861\\
			0.495293694490371	0.386965844404736\\
			0.50036965185808	0.377055577706181\\
			0.505101454088475	0.366977687352624\\
			0.509410268706564	0.35669291018977\\
			0.513687018072934	0.346486268681584\\
			0.517915492247476	0.336338253300138\\
			0.522081893569149	0.326232974478149\\
			0.5261743750803	0.316157460727189\\
			0.530182683115363	0.306101114816333\\
			0.534097877643635	0.29605528793958\\
			0.537912111340097	0.286012943004469\\
			0.541618453498235	0.275968385973166\\
			0.544874858750682	0.265753225302928\\
			0.54784164599067	0.255462754981689\\
			0.550505165586851	0.245100691129792\\
			0.552849903023572	0.234670860977429\\
			0.554858089291862	0.224177219690697\\
			0.556509212156158	0.213623871712643\\
			0.557779397616648	0.203015098019393\\
			0.558640619530595	0.19235539120963\\
			0.559059679040949	0.181649501090051\\
			0.559857664240528	0.171165665208111\\
			0.560654733026672	0.160765157699235\\
			0.561430102719994	0.150434742630345\\
			0.562166300747822	0.140163801031193\\
			0.56284852819356	0.129943821581733\\
			0.563464166873231	0.119768005934837\\
			0.564002393549899	0.109630959612643\\
			0.564453875071107	0.0995284475404815\\
			0.564810525296584	0.0894571989385068\\
			0.565104123370233	0.0794202051922158\\
			0.565141489224178	0.0693906496024043\\
			0.564913942374163	0.0593748479041548\\
			0.564411551804103	0.0493796123459339\\
			0.563622870785364	0.0394123504924696\\
			0.562534604271577	0.0294811893797378\\
			0.561131188002012	0.0195951328708489\\
			0.5593942507225	0.00976426296683292\\
			0.557527855191029	0.000973070662514317\\
		};
		\addlegendentry{\scriptsize {Limitation with} \\[-0.5em] \scriptsize{tolerance window}}
	\end{axis}
	
	\begin{axis}[%
		width=6.5cm,
		height=6.5cm,
		at={(3.384in,0.445in)},
		scale only axis,
		xmin=0,
		xmax=1,
		xtick={ -1,   0, 0.5,   1, 1.5},
		xticklabels={},
		xlabel style={font=\color{white!15!black}},
		xlabel={},
		ymin=0,
		ymax=1,
		ytick={ -1,   0, 0.5,   1, 1.5},
		yticklabels={},
		ylabel style={font=\color{white!15!black}},
		ylabel={},
		axis background/.style={fill=none},
		legend style={legend cell align=left, align=left, draw=white!15!black}
		]
		
		\draw[-latex] (0.272, 0.839) -- node[midway, sloped,above,, yshift=-3pt] {\scriptsize \mySymbol{\gls{rd}}{\gls{iRichtung}}} (0.441,0.783);
		\draw[-latex] (0.441,0.783)--(0.272, 0.839);		
		\draw[-latex] (0.47,0.05) -- node[midway, sloped,above, yshift=-2pt] {\scriptsize $\Delta \mySymbol{\gls{rr}}{\gls{iTol}}$} (0.56,0.05);
		\draw[-latex]  (0.56,0.05)-- (0.47,0.05);		

				\draw[-latex] (0,0) -- node[midway,sloped, above, yshift=-3pt] {\scriptsize \mySymbol{\gls{rr}}{\gls{iKlasse},\gls{iRichtung},\gls{iLim}}} (76:0.5301); 
		\draw (68:0.5301) arc (68:76:0.5301); 

		\draw[-latex] (0,0) -- node[pos=0.95, sloped, below, yshift=3pt] {\scriptsize \mySymbol{\gls{rr}}{\gls{iRL}}} (58:0.9030); 
		\draw (56:0.9030) arc (56:59:0.9030); 

		\draw[-latex] (0,0) -- node[pos=0.7, sloped, above, yshift=-3pt] {\scriptsize \mySymbol{\gls{rr}}{\gls{iRL},\gls{iSicher}}} (62:0.6598); 
		\draw (60:0.6598) arc (59:63:0.6598); 
		
	\end{axis}
\end{tikzpicture}%

%% file: Figures/LimitierungRL.tex
\setlength{\PlotWidth}{4.3cm} 
\setlength{\PlotHeight}{4.3cm} 
\setlength{\DeltaHeight}{1cm} 
\setlength{\DeltaWidth}{0.4cm} 

\begin{tikzpicture}

\newcommand{\NumPlotX}{1}
\newcommand{\NumPlotY}{1}
\pgfmathsetlength{\Ypos}{-1*\NumPlotY*\PlotHeight - 1*\NumPlotY*\DeltaHeight}
\pgfmathsetlength{\Xpos}{1*\NumPlotX*\PlotWidth + 1*\NumPlotX*\DeltaWidth}
\begin{axis}[%
	width=\PlotWidth,
	height=\PlotHeight,
	title={\scriptsize After 2000 combustion cycles},
	title style={yshift=-5pt},
	at={(\Xpos, \Ypos)},
	scale only axis,
	xmin=175,
	xmax=205,
	xlabel style={font=\color{white!15!black}},
	xlabel={$\gls{gAlpha}_{\gls{iNVO}}\,/\,\si{\degreeKW}$},
	ymin=0.25,
	ymax=1,
	ylabel style={font=\color{white!15!black}},
	ylabel={$\gls{rt}_{\gls{iKr},\gls{iInj}}\,/\,\si{\milli\second}$},
	axis background/.style={fill=white},
	xmajorgrids,
	ymajorgrids,
	]	

	\addplot[only marks, mark=*, mark options={}, draw=mmpRed, fill=mmpRed] table [x index=0, y index=1, col sep=comma]{Figures/Data/LimiterungPunktwolke/VSR30089/episode4_class17_Last2_PointsOutsideBorder.csv};

	\addplot[only marks, mark=*, mark options={}, draw=mmpDarkBlue, fill=mmpDarkBlue] table [x index=0, y index=1, col sep=comma]{Figures/Data/LimiterungPunktwolke/VSR30089/episode4_class17_Last2_PointsOnBorder.csv};
	
	\addplot[only marks, mark=*, mark options={}, draw=mmpGreen, fill=mmpGreen] table [x index=0, y index=1, col sep=comma]{Figures/Data/LimiterungPunktwolke/VSR30089/episode4_class17_Last2_PointsInsideBorder.csv};
	\addplot[only marks, mark=*, mark options={fill=mmpBlack}, draw=mmpBlack] coordinates {(187.6793, 0.6606)};
	
	\addplot[color=mmpDarkBlue,line width=1.5pt] table [x expr=\thisrowno{1} + 160, y expr=\thisrowno{2} + 0.5, col sep=comma] {Figures/Data/LimiterungPunktwolke/VSR30080/SafeteyFilterBorder_VSR30080_Class17.csv};
\end{axis}

\begin{axis}[%
	width=\PlotWidth,
	height=\PlotHeight,
	at={(\Xpos, \Ypos)},
	scale only axis,
	xmin=175,
	xmax=205,
	ymin=0.25,
	ymax=1,
	axis line style={opacity=0}, 
	tick style={opacity=0}, 
	xlabel style={font=\color{white!15!black}},
	ylabel style={font=\color{white!15!black}},
	yticklabels={},
	axis background/.style={fill=none},
	]
	\pgfmathsetmacro{\biasX}{187.6793}
	\pgfmathsetmacro{\biasY}{0.6606}
	
	\addplot[color=mmpDarkBlue,line width=0.75pt,forget plot] table [x expr=\thisrowno{1} + \biasX, y expr=\thisrowno{2} + \biasY, col sep=comma] {Figures/Data/LimiterungPunktwolke/VSR30080/SafeteyFilterBorder_VSR30080_Class17.csv};

	\draw[line width=1.5pt,->,>=latex,color=mmpBlack] (axis cs: 187.141303,	0.971069) -- (axis cs:187.366264,	0.841246);
	\draw[line width=1.5pt,->,>=latex,color=mmpBlack] (axis cs:179.340875,	0.452606) -- (axis cs:184.898396,	0.591242);
	\draw[line width=1.5pt,->,>=latex,color=mmpBlack] (axis cs:202.776867,	0.774557) -- (axis cs:194.641784, 0.713160);
	
	\addplot[color=black, thick, dash pattern=on 2pt off 1pt] coordinates {(187.366264,	0.841246) (187.6793, 0.6606)};
	\addplot[color=black, thick, dash pattern=on 2pt off 1pt] coordinates {(184.898396,	0.591242) (187.6793, 0.6606)};
	\addplot[color=black, thick, dash pattern=on 2pt off 1pt] coordinates {(194.641784, 0.713160) (187.6793, 0.6606)};
\end{axis}


\renewcommand{\NumPlotX}{2}
\renewcommand{\NumPlotY}{1}
\pgfmathsetlength{\Ypos}{-1*\NumPlotY*\PlotHeight - 1*\NumPlotY*\DeltaHeight}
\pgfmathsetlength{\Xpos}{1*\NumPlotX*\PlotWidth + 1*\NumPlotX*\DeltaWidth}
\begin{axis}[%
	width=\PlotWidth,
	height=\PlotHeight,
	title={\scriptsize After 12500 combustion cycles},
	title style={yshift=-5pt},
	at={(\Xpos, \Ypos)},
	scale only axis,
	xmin=175,
	xmax=205,
	xlabel style={font=\color{white!15!black}},
	xlabel={$\gls{gAlpha}_{\gls{iNVO}}\,/\,\si{\degreeKW}$},
	ymin=0.25,
	ymax=1,
	ylabel style={font=\color{white!15!black}},
	ylabel={},
	yticklabels={},
	axis background/.style={fill=white},
	xmajorgrids,
	ymajorgrids,
	legend style={at={(0.5,1.1)},anchor=south, legend cell align=left, align=left, draw=none,legend columns=3,column sep=0.2cm}
	]	
	
	\addplot[only marks, mark=*, mark options={}, draw=mmpRed, fill=mmpRed] table [x index=0, y index=1, col sep=comma]{Figures/Data/LimiterungPunktwolke/VSR30089/episode19_class17_Last2_PointsOutsideBorder.csv};
	
	\addplot[only marks, mark=*, mark options={}, draw=mmpDarkBlue, fill=mmpDarkBlue] table [x index=0, y index=1, col sep=comma]{Figures/Data/LimiterungPunktwolke/VSR30089/episode19_class17_Last2_PointsOnBorder.csv};
	
	\addplot[only marks, mark=*, mark options={}, draw=mmpGreen, fill=mmpGreen] table [x index=0, y index=1, col sep=comma]{Figures/Data/LimiterungPunktwolke/VSR30089/episode19_class17_Last2_PointsInsideBorder.csv};

	\addplot[only marks, mark=*, mark options={fill=mmpBlack}, draw=mmpBlack] coordinates {(188.2119, 0.6513)};
	
	\pgfmathsetmacro{\biasX}{188.2119}
	\pgfmathsetmacro{\biasY}{0.6513}
	\addplot[color=mmpDarkBlue,line width=0.75pt] table [x expr=\thisrowno{1} + \biasX, y expr=\thisrowno{2} + \biasY, col sep=comma] {Figures/Data/LimiterungPunktwolke/VSR30080/SafeteyFilterBorder_VSR30080_Class17.csv};

	\addlegendentry{\scriptsize {Selected unsafe action (to be replaced)}};
	\addlegendentry{\scriptsize {Action replaced by safety monitoring}};
	\addlegendentry{\scriptsize {Selected safe action}};
	\addlegendentry{\scriptsize {Starting point of the measurement algorithm}};
	\addlegendentry{\scriptsize {Limitation including tolerance $\Delta\mySymbol{\gls{rr}}{\gls{iTol}}$}};
	
	\addlegendimage{line width=1.5pt,->,>=latex,black}
	\addlegendentry{\scriptsize{Replace}}

\end{axis}

\begin{axis}[
	width=\PlotWidth,
	height=\PlotHeight,
	at={(\Xpos, \Ypos)},
	scale only axis,
	xmin=175,
	xmax=205,
	ymin=0.25,
	ymax=1,
	axis line style={opacity=0}, 
	tick style={opacity=0}, 
	xlabel style={font=\color{white!15!black}},
	ylabel style={font=\color{white!15!black}},
	yticklabels={},
	axis background/.style={fill=none},
	legend style={at={(-0.2,1.2)},anchor=south west,legend cell align=left, align=left, draw=none,legend columns=2,column sep=0.2cm}
	]
	\pgfmathsetmacro{\biasX}{188.2119}
	\pgfmathsetmacro{\biasY}{0.6513}
	\addplot[color=mmpDarkBlue,line width=0.75pt] table [x expr=\thisrowno{1} + \biasX, y expr=\thisrowno{2} + \biasY, col sep=comma] {Figures/Data/LimiterungPunktwolke/VSR30080/SafeteyFilterBorder_VSR30080_Class17.csv};
	\draw[line width=1.5pt,->,>=latex,color=mmpBlack] (axis cs:179.338803,	0.690526) -- (axis cs:182.424515, 0.676882);
	\draw[line width=1.5pt,->,>=latex,color=mmpBlack] (axis cs:204.375555,	0.453490) -- (axis cs:193.724079,	0.583837);
	\draw[line width=1.5pt,->,>=latex,color=mmpBlack] (axis cs:199.813747,	0.693624) -- (axis cs:195.289360,	0.677116);
	
	\addplot[color=black, thick, dash pattern=on 2pt off 1pt] coordinates {(182.424515, 0.676882) (188.2119, 0.6513)};
	\addplot[color=black, thick, dash pattern=on 2pt off 1pt] coordinates {(193.724079,	0.583837) (188.2119, 0.6513)};
	\addplot[color=black, thick, dash pattern=on 2pt off 1pt] coordinates {(195.289360,	0.677116) (188.2119, 0.6513)};
\end{axis}

\renewcommand{\NumPlotX}{3}
\renewcommand{\NumPlotY}{1}
\pgfmathsetlength{\Ypos}{-1*\NumPlotY*\PlotHeight - 1*\NumPlotY*\DeltaHeight}
\pgfmathsetlength{\Xpos}{1*\NumPlotX*\PlotWidth + 1*\NumPlotX*\DeltaWidth}
\begin{axis}[%
	width=\PlotWidth,
	height=\PlotHeight,
	title={\scriptsize After 27000 combustion cycles},
	title style={yshift=-5pt},
	at={(\Xpos, \Ypos)},
	scale only axis,
	xmin=175,
	xmax=205,
	xlabel style={font=\color{white!15!black}},
	xlabel={$\gls{gAlpha}_{\gls{iNVO}}\,/\,\si{\degreeKW}$},
	ymin=0.25,
	ymax=1,
	ylabel style={font=\color{white!15!black}},
	ylabel={},
	yticklabels={},
	axis background/.style={fill=white},
	xmajorgrids,
	ymajorgrids,
	legend style={at={(-0.2,1.2)},anchor=south west,legend cell align=left, align=left, draw=none,legend columns=2,column sep=0.2cm}
	]	
	
	\addplot[only marks, mark=*, mark options={}, draw=mmpRed, fill=mmpRed] table [x index=0, y index=1, col sep=comma]{Figures/Data/LimiterungPunktwolke/VSR30089/episode35_class17_Last1_PointsOutsideBorder.csv};
	
	\addplot[only marks, mark=*, mark options={}, draw=mmpDarkBlue, fill=mmpDarkBlue] table [x index=0, y index=1, col sep=comma]{Figures/Data/LimiterungPunktwolke/VSR30089/episode35_class17_Last1_PointsOnBorder.csv};
	
	\addplot[only marks, mark=*, mark options={}, draw=mmpGreen, fill=mmpGreen] table [x index=0, y index=1, col sep=comma]{Figures/Data/LimiterungPunktwolke/VSR30089/episode35_class17_Last1_PointsInsideBorder.csv};
	
	\addplot[only marks, mark=*, mark options={fill=mmpBlack}, draw=mmpBlack] coordinates {(186.4766, 0.6817)};
\end{axis}

\begin{axis}[
	width=\PlotWidth,
	height=\PlotHeight,
	at={(\Xpos, \Ypos)},
	scale only axis,
	xmin=175,
	xmax=205,
	ymin=0.25,
	ymax=1,
	axis line style={opacity=0}, 
	tick style={opacity=0}, 
	xlabel style={font=\color{white!15!black}},
	ylabel style={font=\color{white!15!black}},
	yticklabels={},
	axis background/.style={fill=none},
	legend style={at={(-0.2,1.2)},anchor=south west,legend cell align=left, align=left, draw=none,legend columns=2,column sep=0.2cm}
	]
	\pgfmathsetmacro{\biasX}{186.4766}
	\pgfmathsetmacro{\biasY}{0.6817}
	\addplot[color=mmpDarkBlue,line width=0.75pt] table [x expr=\thisrowno{1} + \biasX, y expr=\thisrowno{2} + \biasY, col sep=comma] {Figures/Data/LimiterungPunktwolke/VSR30080/SafeteyFilterBorder_VSR30080_Class17.csv};
\end{axis}

\end{tikzpicture}

%% file: Figures/BerechnugAblaufRLWide.tex
\begin{tikzpicture}

\newcommand{\angleHDWindowStart}{-180}
\newcommand{\angleHDWindowMid}{90}

\newcommand{\angleHDWindowEnd}{180}

\newcommand{\angleZKWindowStart}{300}	
\newcommand{\angleZKWindowEnd}{420}

\newcommand{\DataTransferCStart}{400}
\newcommand{\DataTransferCEnd}{409}

\newcommand{\RewardStart}{90}
\newcommand{\RewardEnd}{108}
\newcommand{\PolicyStart}{108}
\newcommand{\PolicyEnd}{189}
\newcommand{\CalcUAStart}{189}
\newcommand{\CalcUAEnd}{207}

\newcommand{\ActuationHDNVOStart}{225}
\newcommand{\ActuationHDNVOEnd}{582}

\newcommand{\YFPGAA}{50}
\newcommand{\YFPGAB}{58.5}
\newcommand{\YFPGAC}{67}

\newcommand{\YFPGAD}{75.5}
\newcommand{\YFPGAE}{84}

\newcommand{\YFPGAF}{92.5}
\newcommand{\YFPGAG}{101}

\begin{axis}[
ymin=0, ymax=101, 
xmin=-100, xmax=920, 
axis on top, 
width=14cm,
height=5.5cm,
xtick=\empty,
ytick=\empty,
legend style={at={(0.5,1.01)},anchor=south,legend cell align=left, align=left, draw=none,legend columns=4,column sep=0.3cm,font=\scriptsize},
clip=false
]
\node[align=left,anchor=east,fill=white] at (axis cs:-110,\YFPGAF) {\scriptsize \gls{kRPI} };
\node[align=left,anchor=east,fill=white] at (axis cs:-110,\YFPGAD) {\scriptsize\gls{kMABX}};
\node[align=left,anchor=east,fill=white] at (axis cs:-110,\YFPGAB) {\scriptsize \gls{kFPGA}};
\end{axis}

\begin{axis}[
			ymin=0, ymax=101, 
			xmin=-100, xmax=920, 
			xlabel={Crank angle \gls{gAlpha} / \si{\degreeKW}},
			ylabel={$\pzyl$ / \si{\bar}},
			ylabel style={at={(-0.05,0.25)}},
			axis on top, 
			width=14cm,
			height=5.5cm,
			xtick={-200,0,200,400,600,800,1000},
			ytick={0,20,40},
			legend style={at={(0.5,1.01)},anchor=south,legend cell align=left, align=left, draw=none,legend columns=4,column sep=0.3cm,font=\scriptsize},
			]
\addplot[color=mmpBlack,line width=1.5pt,forget plot] table [x index=0, y index=1, col sep=comma] {Figures/Data/DreiZyklen/3PressureCycles.csv};

\addplot[color=mmpLightBlue,line width=1.5pt,forget plot] table [x index=0, y index=2, col sep=comma] {Figures/Data/DreiZyklen/3PressureCycles.csv};
\addplot[color=mmpLightBlue,line width=1.5pt,forget plot] table [x index=0, y index=3, col sep=comma] {Figures/Data/DreiZyklen/3PressureCycles.csv};
		
\draw[dashed] (axis cs:-360,\YFPGAA) -- (axis cs:1300,\YFPGAA);
\draw[dashed] (axis cs:-360,\YFPGAC) -- (axis cs:1300,\YFPGAC);
\draw[dashed] (axis cs:-360,\YFPGAE) -- (axis cs:1300,\YFPGAE);
\draw[dashed] (axis cs:180,0) -- (axis cs:180,50);
\draw[dashed] (axis cs:180,70) -- (axis cs:180,100);
\draw[dashed] (axis cs:900,0) -- (axis cs:900,100);

\draw[very thick,-latex] (axis cs: 118,43) -- (axis cs:180,43);
\draw[very thick,-latex] (axis cs:-38,43) -- (axis cs:-100,43);

\draw[very thick,-latex] (axis cs:483,43) -- (axis cs:180,43);
\draw[very thick,-latex] (axis cs:597,43) -- (axis cs:900,43);
\draw[very thick,-latex] (axis cs: 1030,43) -- (axis cs:900,43);

\draw[] (axis cs:\ActuationHDNVOStart,75) -- (axis cs:\ActuationHDNVOStart,50);
\draw[] (axis cs:\angleHDWindowMid,75) -- (axis cs:\angleHDWindowMid,51);

\draw[] (axis cs:\CalcUAEnd,77) -- (axis cs:\CalcUAEnd,70);
\draw[] (axis cs:\CalcUAEnd-25,72) -- (axis cs:\ActuationHDNVOStart+60,72);
\draw[-latex] (axis cs:\CalcUAEnd-60,72)  -- (axis cs:\CalcUAEnd,72);
\draw[-latex] (axis cs:\ActuationHDNVOStart+80,72)  -- (axis cs:\ActuationHDNVOStart,72);
\node[align=center] at (axis cs:\ActuationHDNVOStart+55,74.5) {\tiny \SI{2}{\milli\second} };

\draw[] (axis cs:\angleHDWindowEnd-25,61) -- (axis cs:\angleHDWindowEnd+60,61);
\draw[-latex] (axis cs:\angleHDWindowEnd-60,61)  -- (axis cs:\angleHDWindowEnd,61);
\draw[-latex] (axis cs:\ActuationHDNVOStart+80,61)  -- (axis cs:\ActuationHDNVOStart,61);
\node[align=center] at (axis cs:\ActuationHDNVOStart+55,63.5) {\tiny \SI{5}{\milli\second} };


\draw[-latex] (axis cs:\ActuationHDNVOStart,52)  -- (axis cs:90,52);
\draw[-latex] (axis cs:90,52)  -- (axis cs:\ActuationHDNVOStart,52);
\node[align=center] at (axis cs:160,54.5) {\tiny \SI{15}{\milli\second} };

\path[name path=T](rel axis cs:1.1,1.1)--(rel axis cs:1.2,1.1); 
\path[name path=B](rel axis cs:1.1,1.05)--(rel axis cs:1.2,1.05);; 

\node[align=center,anchor=center] at (axis cs:40,43) {\scriptsize Cycle~$\gls{iZyklus}-1$};
\node[align=center,anchor=center] at (axis cs:540,43) {\scriptsize Cycle~$\gls{iZyklus}$};

\addplot[draw=black, fill=mmpBlack, opacity=1]
fill between [%
of=T and B,
soft clip={domain=315:405}
];
\addlegendentry{State Calculation}
\addplot[draw=none,fill=none, opacity=1]
fill between [%
of=T and B,
soft clip={domain=315:405}
];
\addlegendentry{State Prediction}

\addplot[draw=black, fill=mmpRed, opacity=1]
fill between [%
of=T and B,
soft clip={domain=315:405}
];
\addlegendentry{Reward Calculation}

\addplot[draw=black, fill=mmpMiddleBlue, opacity=1]
fill between [%
of=T and B,
soft clip={domain=315:405}
];
\addlegendentry{UDP Transmission}

\addplot[draw=black, fill=mmpMiddleGray, opacity=1]
fill between [%
of=T and B,
soft clip={domain=315:405}
];
\addlegendentry{Policy + Noise}

\addplot[draw=black, fill=mmpDarkBlue, opacity=1]
fill between [%
of=T and B,
soft clip={domain=315:405}
];
\addlegendentry{Safety Monitoring}

\addplot[draw=black, fill=mmpLightBlue, opacity=1]
fill between [%
of=T and B,
soft clip={domain=315:405}
];
\addlegendentry{Actuation}

\addplot [
fill=mmpBlack,
opacity=1,
draw=black,
forget plot
] coordinates {({\angleHDWindowStart},\YFPGAB) ({\angleHDWindowMid},\YFPGAB) ({\angleHDWindowMid},\YFPGAC) ({\angleHDWindowStart},\YFPGAC)} -- cycle;

\addplot [
pattern=north east lines,
pattern color=black,
opacity=1,
draw=black,
forget plot
] coordinates {({\angleHDWindowMid},\YFPGAB) ({\angleHDWindowEnd},\YFPGAB) ({\angleHDWindowEnd},\YFPGAC) ({\angleHDWindowMid},\YFPGAC)} -- cycle;

\addplot [
fill=mmpBlack,
opacity=1,
draw=black,
forget plot
] coordinates {({\angleHDWindowStart+720},\YFPGAB) ({\angleHDWindowMid+720},\YFPGAB) ({\angleHDWindowMid+720},\YFPGAC) ({\angleHDWindowStart+720},\YFPGAC)} -- cycle;

\addplot [
pattern=north east lines,
pattern color=black,
opacity=1,
draw=black,
forget plot
] coordinates {({\angleHDWindowMid+720},\YFPGAB) ({\angleHDWindowEnd+720},\YFPGAB) ({\angleHDWindowEnd+720},\YFPGAC) ({\angleHDWindowMid+720},\YFPGAC)} -- cycle;

\addplot [
fill=mmpDarkBlue,
opacity=1,
draw=black,
forget plot
] coordinates {(\CalcUAStart,\YFPGAD) (\CalcUAEnd,\YFPGAD) (\CalcUAEnd,\YFPGAE) (\CalcUAStart,\YFPGAE)} -- cycle;

\addplot [
fill=mmpDarkBlue,
opacity=1,
draw=black,
forget plot
] coordinates {(\CalcUAStart+720,\YFPGAD) (\CalcUAEnd+720,\YFPGAD) (\CalcUAEnd+720,\YFPGAE) (\CalcUAStart+720,\YFPGAE)} -- cycle;

\addplot [
fill=mmpMiddleGray,
opacity=1,
draw=black,
forget plot
] coordinates {(\PolicyStart,\YFPGAE) (\PolicyEnd,\YFPGAE) (\PolicyEnd,\YFPGAG) (\PolicyStart,\YFPGAG)} -- cycle;

\addplot [
fill=mmpMiddleGray,
opacity=1,
draw=black,
forget plot
] coordinates {(\PolicyStart+720,\YFPGAE) (\PolicyEnd+720,\YFPGAE) (\PolicyEnd+720,\YFPGAG) (\PolicyStart+720,\YFPGAG)} -- cycle;

\addplot [
fill=mmpRed,
opacity=1,
draw=black,
forget plot
] coordinates {(\RewardStart,\YFPGAC) (\RewardEnd,\YFPGAC) (\RewardEnd,\YFPGAD) (\RewardStart,\YFPGAD)} -- cycle;

\addplot [
fill=mmpRed,
opacity=1,
draw=black,
forget plot
] coordinates {(\RewardStart+720,\YFPGAC) (\RewardEnd+720,\YFPGAC) (\RewardEnd+720,\YFPGAD) (\RewardStart+720,\YFPGAD)} -- cycle;

\addplot [
fill=mmpMiddleBlue,
opacity=1,
draw=black,
forget plot
] coordinates {(\RewardStart,\YFPGAD) (\RewardEnd,\YFPGAD) (\RewardEnd,\YFPGAE) (\RewardStart,\YFPGAE)} -- cycle;

\addplot [
fill=mmpMiddleBlue,
opacity=1,
draw=black,
forget plot
] coordinates {(\RewardStart+720,\YFPGAD) (\RewardEnd+720,\YFPGAD) (\RewardEnd+720,\YFPGAE) (\RewardStart+720,\YFPGAE)} -- cycle;

\addplot [
fill=mmpLightBlue,
opacity=1,
draw=black,
forget plot
] coordinates {({\ActuationHDNVOStart-720},\YFPGAA) ({\ActuationHDNVOEnd-720},\YFPGAA) ({\ActuationHDNVOEnd-720},\YFPGAB) ({\ActuationHDNVOStart-720},\YFPGAB)} -- cycle;

\addplot [
fill=mmpLightBlue,
opacity=1,
draw=black,
forget plot
] coordinates {({\ActuationHDNVOStart},\YFPGAA) ({\ActuationHDNVOEnd},\YFPGAA) ({\ActuationHDNVOEnd},\YFPGAB) ({\ActuationHDNVOStart},\YFPGAB)} -- cycle;				

\addplot [
fill=mmpLightBlue,
opacity=1,
draw=black,
forget plot
] coordinates {({\ActuationHDNVOStart+720},\YFPGAA) ({\ActuationHDNVOEnd+720},\YFPGAA) ({\ActuationHDNVOEnd+720},\YFPGAB) ({\ActuationHDNVOStart+720},\YFPGAB)} -- cycle;

\end{axis}

 Schraffierter Bereich mit Mini-Axis
\node at (3.336cm,4.575cm) {
	\begin{tikzpicture}
		\begin{axis}[
			width=0.722cm, height=0.238cm, 
			axis lines=none, 
			scale only axis, 
			clip=false,
			]
			\addplot [
			fill=none, 
			pattern=north east lines, 
			pattern color=mmpBlack, 
			draw=mmpBlack
			] coordinates {(0,0) (1,0) (1,1) (0,1)} -- cycle;
		\end{axis}
	\end{tikzpicture}
};

\end{tikzpicture}

%% file: Chapters/Validation.tex
\section{Results and Discussion}
\label{chap:Validation}
To validate our safe RL approach, first, an initially untrained agent is trained purely through direct interaction with the real-world testbench environment. Secondly, the agent's adaptability to changing objectives is demonstrated.
\subsection{Policy Training in the Real-World Environment}
\label{cha:PolicyTrainingReal}
In this feasibility study, the agent learns exclusively through experiences gathered from direct interaction with the real-world testbench environment. Table~\ref{tab:RLHyperParam} contains the hyperparameters used for this training. Those hyperparameters were selected iteratively by manual tuning until the agent's performance was satisfactory, ensuring they effectively supported the tracking, stability, and safety objectives defined in Section~\ref{cha:RewardFunction}.
\begin{table}[!htpb]
	\caption{Hyperparameters for the Training of the Agent's Policy in the Real-World Testbench Environment.}
	\label{tab:RLHyperParam}
	\begin{center}
		\begin{tabular}{ll}
			Parameter & Specification\\
			\toprule
			Initial standard deviation~\gls{gsigma} &$0.5$\\
			Decay factor~\gls{gLambdaRL} &$0.95$\\
			Discount factor~\gls{ggamma}&$0{.}9$\\
			Training batch size & $64$\\
			Net topology critic & [64 64]\\
			Activation function critic & \gls{kRelu}\\
			Learning rate critic~$\gls{gxi}_{\gls{rQfunc}}$ & $1 \cdot 10^{-3}$\\
			Net topology actor & [64 64]\\
			Activation function actor & \gls{kRelu}\\
			Learning rate actor $\gls{gxi}_{\gls{gmRL}}$ & $1 \cdot 10^{-3}$\\
			Size replay buffer & $50,000$\\
			Polyiak averaging $\gls{grhoPoly}$&$1 \cdot 10^{-3}$\\
			\bottomrule
		\end{tabular}
	\end{center}
\end{table}

For validation, an ANN-based inverse process model, as described in~\citep{Bedei.2023}, serves as the reference strategy. The training database~\citep{bedei2024safe} required for the ANN is generated using the dynamic measurement method (Section~\ref{chap:Safety Monitoring}) and includes $68,000$ combustion cycles, which are also used to automatically parameterize the limitation matrix $R_{\text{Lim}}$ during measurement.

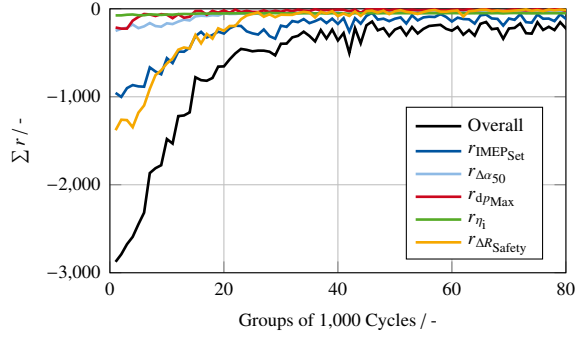
\begin{figure}[!htpb]
	\centering
	\input{Figures/PlotRewardUntrainiert.tex}
	\caption{Evolution of the Cumulative Reward $\sum\gls{rRReward}$ Including All Components of the Reward Function During the Agent's Training in the Real-World Testbench Environment.}
	\label{fig:PlotRewardUntrainiert}
\end{figure}

Figure~\ref{fig:PlotRewardUntrainiert} illustrates the evolution of the total reward and its components over time, with the non-discounted cumulative reward $\sum \gls{rRReward}$ plotted for groups of $1,000$ consecutive training combustion cycles. 

As described in Section~\ref{cha:RewardFunction}, the focus of the weightings is on the reward components related to safety monitoring $\gls{rRReward}_{\Delta\gls{rr}_{\gls{iSafetyFilter}}}$ and pressure gradient limitation $\gls{rRReward}_{\gls{rDPMAX}}$. At the beginning of the training, the cumulative penalty for exceeding the pressure gradient (shown in red) is already relatively low, primarily due to the action monitoring, which reduces both the number of cycles with excessive pressure gradients and the magnitude of remaining overshoots. 

\begin{figure}[!htpb]
	\centering
	\input{Figures/PlotDPMAXDistUntrainiert.tex}
	\caption{Distribution of the Maximum Pressure Gradient~\gls{rDPMAX} During the Agent's Training in the Real-World Testbench Environment.}
	\label{fig:PlotDPMAXDistUntrainiert}
\end{figure}
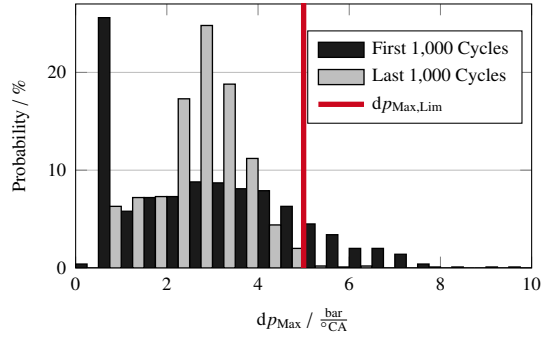

This is also reflected by the distribution of the pressure gradient $\gls{rDPMAX}$ shown for the first and last $1,000$ training combustion cycles in Figure~\ref{fig:PlotDPMAXDistUntrainiert}.

For the first $1,000$ cycles, a significant reduction in the probability distribution is shown when exceeding $\gls{rDPMAX}=\SI{5}{\bar\per\degreeKW}$. Remaining violations are minor and are mainly attributed to process stochasticity and the tolerance window $\Delta \mySymbol{\gls{rr}}{\gls{iTol}}$ of the monitoring function. This allows the agent to learn smaller limit violations by itself, while preventing larger, potentially harmful ones. Additionally, the tolerance window applied extends the allowed action space toward regions where misfires are more likely. This results in a higher probability of cycles with low pressure gradients $\left(\gls{rDPMAX}\leq\SI{1}{\bar\per\degreeKW}\right)$ during the first $1,000$ cycles. It is important to note that not all cycles within this group correspond to misfires; some low-load cycles may also exhibit low pressure gradients while still maintaining proper combustion. Over the course of training, the agent improves its adherence to both the misfiring and the pressure gradient limit, leading to a significant reduction in limit violations, as shown by the distribution across the last $1,000$ training combustion cycles.

Thus, safety monitoring serves as a key enabler for the safe deployment of RL algorithms in real-world environments. By effectively reducing the risk of excessive constraint violations, our approach enables RL in settings like HCCI testbenches, where safety is critical.

Thus, in contrast to the pressure gradient limitation, the penalty for safety monitoring, shown in orange in Figure~\ref{fig:PlotRewardUntrainiert}, is relatively high at the beginning. This is due to the initial selection of actions based on Gaussian noise with high standard deviation (see Figure~\ref{fig:FilterRL}), leading to larger deviations from the IMEP setpoint and a relatively high cumulative penalty $\gls{rRReward}_{\gls{rpmiSoll}}$. Process stability is also lower in this phase, but its impact on the total reward is minimal due to the lower weighting of $\gls{rRReward}_{\Delta\gls{gAlpha50}}$.

With progression of the training, all reward components increase. The total reward converges after approximately $50,000$ cycles, which corresponds to an engine runtime of around $1.1\,\si{\hour}$. However, due to pauses to execute the training on the \gls{kRPI} after each episode, as well as occasional valve malfunctions or combustion misfires, the total time to reach convergence extends to approximately $\SI{2}{\hour}$.

\begin{figure}[!htpb]
	\centering
	\input{Figures/PlotEpisodeUntrainiert.tex}
	\caption{Validation Episode of the Converged Policy, Trained Exclusively in the Real-World Testbench, in Comparison to an ANN-Based Inverse Process Model.}
	\label{fig:PlotEpisodeUntrainiert}
\end{figure}
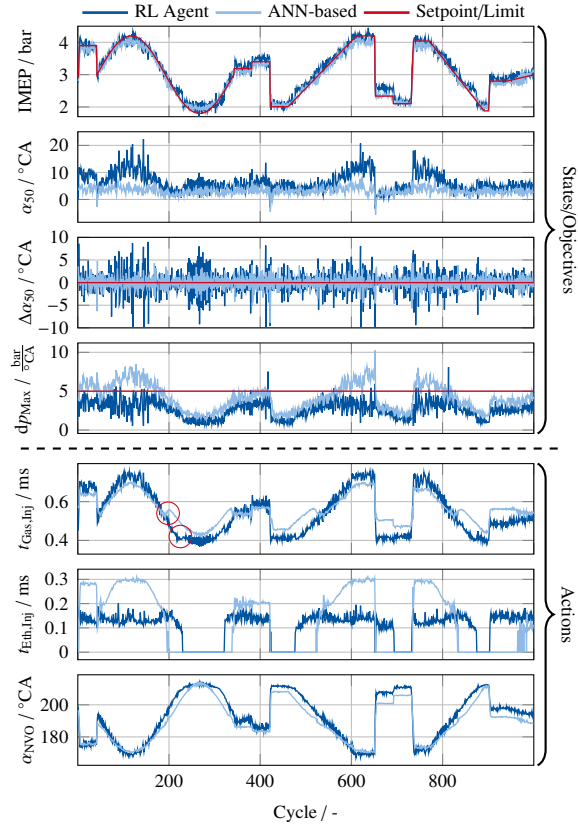

Figure~\ref{fig:PlotEpisodeUntrainiert} compares the learned policy with a control strategy using an ANN-based inverse process model, as described in \citep{Bedei.2023}.

The \gls{kRL} agent tracks the IMEP setpoint with a small deviation. The root mean square error (\gls{rRMSE}) for $\gls{rpmi}$ is $0.137\,\si{\bar}$, slightly higher than that of the inverse process model ($\text{\gls{rRMSE}}=0.107\,\si{\bar}$). This difference is partly due to discrete steps in the injected ethanol mass when the injector's minimum opening time of $0.08\,\si{\milli\second}$ is not reached. Below this duration, no ethanol is injected, while at $0.08\,\si{\milli\second}$, the injected mass increases discretely to the minimum possible value. This behavior creates a pronounced, step-like change in the ethanol mass at this threshold. As a result, the gasoline $\gls{rt}_{\gls{iKr},\gls{iInj}}$ and ethanol $\gls{rt}_{\gls{iEth},\gls{iInj}}$ injections show that the inverse model significantly increases the gasoline injection duration at points where the ethanol injection falls below the threshold as indicated by the red circle. This compensation is less pronounced with the \gls{kRL} agent, indicating a challenge for the agent in learning such discrete steps of the actions. This behavior leads to significant static offsets in the control deviation of the \gls{kRL} agent at certain loads, for example, during cycles 653 to 691. For higher loads, the control deviations of both approaches are of the same order of magnitude.

One disadvantage of the inverse process model is its inability to meet boundary conditions, such as pressure gradient limitations, which results in frequent overshoots under high load requirements. A total of 330 cycles violate the pressure gradient limit, with a mean overshoot of $1.18\,\si{\bar}$. In contrast, the \gls{kRL} policy consistently adheres to this limit, with only 19 violations and a mean overshoot of $0.61\,\si{\bar}$. At higher loads, the two control approaches primarily differ in their injection strategies. Notably, the inverse model utilizes longer ethanol injection durations at higher target loads compared to the \gls{kRL} policy. Since \gls{kNVO} durations $\gls{gAlpha}_{\gls{iNVO}}$ are similar for both approaches, this implies that the richer mixture resulting from the larger ethanol mass injected  in case of the inverse process model increases the likelihood of exceeding the pressure gradient limit.

The \gls{kRL} agent’s policy also retards the combustion phasing~$\gls{gAlpha50}$ under high load demands, effectively limiting the pressure gradient. This adjustment is enabled by the reward function, which avoids a fixed target for the phasing but encourages minimizing cycle-to-cycle fluctuations, measured using the stability objective (i.e. $\sqrt{\sum\left(\Delta \gls{gAlpha50}\right)^2}$), yielding a value of $2.66\,\si{\degreeKW}$. In contrast, the inverse process model targets a fixed phasing of $\gls{gAlpha50Soll}=\SI{6}{\degreeKW}$ to increase efficiency. An \gls{rRMSE} of $3.11\,\si{\degreeKW}$ is achieved, which is not directly comparable to the RL objective.

Regarding the efficiency objective the RL agent achieves a mean thermal efficiency of $30.2\,\%$ outperforming the ANN-based approach with $28.8\,\%$.

\subsection{Online Adaptation of the Agent's Policy}
\label{cha:OnlineAdaption}
In this scenario, we investigate the adaptability of the converged policy from the experiment presented in Section~\ref{cha:PolicyTrainingReal}. Specifically, the agent’s ability to adjust its policy to achieve a higher ethanol energy share --~thus supporting a greater share of renewable, carbon-neutral fuels~-- is examined, while adhering to safety criteria. Starting from the previously learned safe policy, the agent needs to explore new regions of the experimental space that lack prior dynamic measurement data. Consequently, safety monitoring, which is restricted to areas covered by the measurement algorithm beforehand, cannot be applied and the corresponding reward component $\gls{rRReward}_{\Delta\gls{rr}{\gls{iSafetyFilter}}}$ is set to zero. 

To ensure safety while exploring uncharted areas of the action space, the agent’s policy is updated slowly, avoiding abrupt transitions into prohibited regions. For this purpose, the standard deviation of exploratory noise is reduced to $\gls{gsigma}=0.3$, compared to the higher value of $\gls{gsigma}=0.5$ used in pure online training with action monitoring enabled. This reduction favors safety by keeping the agent’s actions closer to known safe policies. Otherwise the same hyperparameters as for the first scenario, listed in Table~\ref{tab:RLHyperParam}, are used.

To support policy adaptation, an additional reward component $\gls{rRReward}_{\Delta \gls{rxAnteil}_{\gls{rEEnergie}_{\gls{iEth}}}}$ is introduced into the reward function to evaluate the deviation from a target ethanol energy share $\gls{rxAnteil}_{\gls{rEEnergie}_{\gls{iEth}},\gls{iSoll}}$, set to $50\,\%$. The additional term is given substantial weight to strongly encourage the agent to adapt its policy.

\begin{figure}[!htpb]
	\centering
	\input{Figures/PlotRewardAdaption.tex}
	\caption{Evolution of the Cumulative Reward $\sum\gls{rRReward}$ Including All Parts of the Reward Function for Adaptation of the Agent's Policy in the Real-World Testbench Environment.}
	\label{fig:PlotRewardAdaption}
\end{figure}
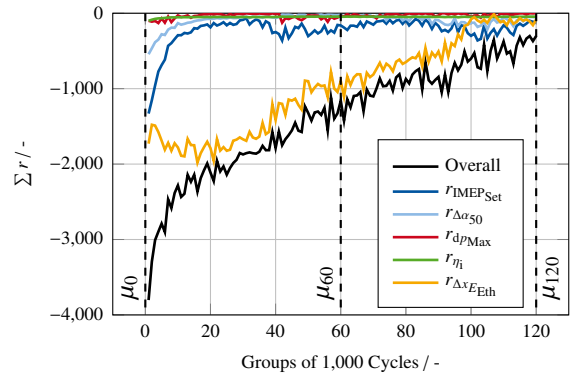

Figure~\ref{fig:PlotRewardAdaption} illustrates the evolution of the non-discounted cumulative reward during real-world policy adaptation in groups of $1,000$ training combustion cycles.

As shown by the reward evolution, at the beginning, the total reward is mainly influenced by the deviation from the target ethanol energy share $\Delta\gls{rxAnteil}_{\gls{rEEnergie}_{\gls{iEth}}}$ through large weights shown in orange. As a result, the agent increases the ethanol energy share toward the target over time, resulting in continuous increase of the reward component $\gls{rr}_\Delta\gls{rxAnteil}_{\gls{rEEnergie}_{\gls{iEth}}}$ and the total reward.

After approximately $90,000$ training cycles, while further reducing the penalty for the ethanol share deviations~$\gls{rRReward}_{\Delta\gls{rxAnteil}_{\gls{rEEnergie}_{\gls{iEth}}}}$, a slight increase in penalties for stability $\gls{rRReward}_{\Delta\gls{gAlpha50}}$ and IMEP setpoint $\gls{rRReward}_{\gls{rpmiSoll}}$ is observed.  This suggests that not all objectives defined in the reward function can be fully achieved simultaneously. Ultimately, the agent finds a balance among safety, stability, efficiency, and setpoint tracking based on the reward function weights. This is confirmed by the resulting key objectives and their evolution given in table~\ref{tab:ObjectivesAdaptation}, showing an increase for the load tracking and stability objectives while the \gls{rRMSE} for the ethanol energy share is reduced significantly.

\begin{table*}[htbp]
	\centering
	\caption{Key Objectives and Their Evolution During the Online Adaptation of the Agent's Policy.}
	\label{tab:ObjectivesAdaptation}
	\begin{tabular}{lcccc}
		\toprule
		Objective & Metric & $\gls{gmRL}_0$& $\gls{gmRL}_{60}$& $\gls{gmRL}_{120}$ \\
		\midrule
		Load tracking & $\sqrt{\sum\left(\gls{rpmi}-\gls{rpmiSoll}\right)^2}$    &$0.139\,\si{\bar}$  &$0.224\,\si{\bar}$  &$0.237\,\si{\bar}$ \\
		Stability & $\sqrt{\sum\left(\Delta \gls{gAlpha50}\right)^2}$  &$2.52\,\si{\degreeKW}$&$1.79\,\si{\degreeKW}$  &$2.68\,\si{\degreeKW}$  \\
		Pressure gradient limitation& Number of violations &159 &64 &4 \\
		Pressure gradient limitation& Mean overshoot &$0.74\,\si{\bar}$ &$0.59\,\si{\bar}$ &$0.23\,\si{\bar}$\\
		Efficiency & $\gls{geta}$ &$30.41\,\%$   &$30.89\,\%$   &$32.59\,\%$ \\
		Ethanol energy share & $\sqrt{\sum\left(\Delta \gls{rxAnteil}_{\gls{rEEnergie}_{\gls{iEth}}} \right)^2}$ & $0.3501$ & $0.2115$ &$0.0416$\\
		\bottomrule
	\end{tabular}
\end{table*}

\begin{figure}[!htpb]
	\centering
	\input{Figures/PlotEpisodeStrategieadaption.tex}
	\caption{Adaptation of the Agent's Policy in the Real-World Testbench Environment Shown by Three Validation Episodes at Different Stages of the Training.}
	\label{fig:PlotEpisodeStrategieadaption}
\end{figure}
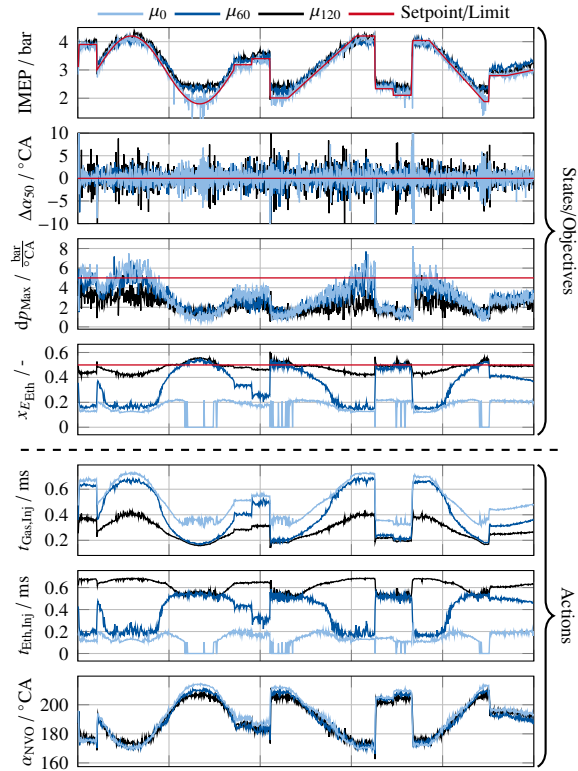

Thus, after $120,000$ training cycles small static IMEP deviations are observed in favor of increased stability for low load setpoints, which can be seen in Figure~\ref{fig:PlotEpisodeStrategieadaption}.

Figure~\ref{fig:PlotEpisodeStrategieadaption} shows three validation episodes: at the beginning $\left(\gls{gmRL}_0\right)$, after $60,000$ training combustion cycles $\left(\gls{gmRL}_{60}\right)$ and after $120,000$ cycles $\left(\gls{gmRL}_{120}\right)$. The specific time instances are highlighted with dashed lines in Figure~\ref{fig:PlotRewardAdaption}.

As illustrated, the ethanol energy share, $\gls{rxAnteil}_{\gls{rEEnergie}_{\gls{iEth}}}$, for the initial policy $\gls{gmRL}_0$ consistently remains below $25\,\%$. At this stage, the deviation is $\gls{rRMSE}\left(\gls{rxAnteil}_{\gls{rEEnergie}_{\gls{iEth}},\gls{gmRL}_0}\right)=0.3501$. After $60,000$ training cycles where the validation episode with $\gls{gmRL}_{60}$ is measured, the target ethanol energy share is already met for low load requirements, though it remains too low for higher loads. As can be seen from the injection durations $\gls{rt}_{\gls{iKr},\gls{iInj}},\gls{rt}_{\gls{iEth},\gls{iInj}}$, showing the reduced amount of gasoline and increased amount of ethanol, the policy is already significantly adapted at this stage. The deviation is reduced to $\gls{rRMSE}\left(\gls{rxAnteil}_{\gls{rEEnergie}_{\gls{iEth}},\gls{gmRL}_{60}}\right)=0.2115$. The final policy, $\gls{gmRL}_{120}$, is much closer to the target ethanol share, achieving an  $\gls{rRMSE}\left(\gls{rxAnteil}_{\gls{rEEnergie}_{\gls{iEth}},\gls{gmRL}_{120}}\right)$ of $0.0416$. 

Regarding safety during adaptation, even without safety monitoring, no increase in the reward component associated with pressure gradient limitations $\gls{rRReward}_{\gls{rDPMAX}}$ is observed as shown by the red line in Figure~\ref{fig:PlotRewardAdaption}. The \gls{kRL} algorithm’s safety is preserved, primarily due to slow policy adaptation and lower exploratory noise. Disabling safety monitoring is feasible only because a safe policy was learned before through its use. Thus, our safety monitoring remains an essential component, ensuring a safe initial policy for adaptation, even when it is later disabled. Consequently, the toolchain’s ability to safely enable RL showcases its crucial role in bridging safety gaps for real-world applications.

%% file: Figures/PlotRewardUntrainiert.tex
\begin{tikzpicture}
\setlength{\PlotWidth}{6cm} 
\setlength{\PlotHeight}{3.5cm} 
\setlength{\DeltaHeight}{1cm}
\newcommand{\NumPlot}{1}
\pgfmathsetlength{\Ypos}{-1*\NumPlot*\PlotHeight - 1*\NumPlot*\DeltaHeight}
\begin{axis}[%
	width=\PlotWidth,
	height=\PlotHeight,
	at={(0cm, \Ypos)},
	scale only axis,
	xmin=0,
	xmax=80,
	ymin=-3000,
	ymax=0,
	xticklabel style={/pgf/number format/fixed},
	xlabel style={font=\color{white!15!black}},
	xlabel={Groups of $1{,}000$ Cycles / -},
	ylabel={$\sum\gls{rRReward}$ / -},
    ytick={-3000,-2000,-1000,0},
	yticklabels={$-3{,}000$,$-2{,}000$,$-1{,}000$,$0$},
	axis background/.style={fill=white},
	xmajorgrids,
	ymajorgrids,
	legend style={at={(0.95,0.05)},legend columns=1, legend cell align=left, align=left, anchor=south east,  draw=black,fill=white},
	label style={font=\scriptsize},
	tick label style={font=\scriptsize},
	legend style={font=\scriptsize}
	]
	\addplot[ color=mmpBlack,line width=1pt] table [x index=0, y index=1, col sep=comma] {Figures/Data/PlotEpisode/RewardVSR1000Groups30099.csv};
	\addlegendentry{Overall}
	
	\addplot[ color=mmpDarkBlue,line width=1pt] table [x index=0, y index=8, col sep=comma] {Figures/Data/PlotEpisode/RewardVSR1000Groups30099.csv};
	\addlegendentry{$\gls{rRReward}_{\gls{rpmiSoll}}$}
	
	\addplot[ color=mmpLightBlue,line width=1pt] table [x index=0, y index=5, col sep=comma] {Figures/Data/PlotEpisode/RewardVSR1000Groups30099.csv};
	\addlegendentry{$\gls{rRReward}_{\Delta\gls{gAlpha50}}$}
	
	\addplot[ color=mmpRed,line width=1pt] table [x index=0, y index=6, col sep=comma] {Figures/Data/PlotEpisode/RewardVSR1000Groups30099.csv};
	\addlegendentry{$\gls{rRReward}_{\gls{rDPMAX}}$}
	
	\addplot[ color=mmpGreen,line width=1pt] table [x index=0, y index=7, col sep=comma] {Figures/Data/PlotEpisode/RewardVSR1000Groups30099.csv};
	\addlegendentry{$\gls{rRReward}_{\gls{geta}}$}
	
	\addplot[ color=mmpOrange,line width=1pt] table [x index=0, y index=9, col sep=comma] {Figures/Data/PlotEpisode/RewardVSR1000Groups30099.csv};
	\addlegendentry{$\gls{rRReward}_{\Delta\gls{rr}_{\gls{iSafetyFilter}}}$}
	
\end{axis}

\end{tikzpicture}

%% file: Figures/PlotDPMAXDistUntrainiert.tex
\begin{tikzpicture}
\setlength{\PlotWidth}{6cm} 
\setlength{\PlotHeight}{3.5cm} 
\setlength{\DeltaHeight}{1cm}
\newcommand{\NumPlot}{1}
\pgfmathsetlength{\Ypos}{-1*\NumPlot*\PlotHeight - 1*\NumPlot*\DeltaHeight}
\begin{axis}[%
	width=\PlotWidth,
	height=\PlotHeight,
	at={(0cm, \Ypos)},
	scale only axis,
	xmin=0,
	xmax=10,
	xlabel style={font=\color{white!15!mmpBlack}},
	xlabel={\gls{rDPMAX} / $\frac{\si{\bar}}{\si{\degreeKW}}$},
	ymin=0,
	ymax=27,
	ylabel style={font=\color{white!15!mmpBlack}},
	ylabel={Probability / \%},
	axis background/.style={fill=white},
	ymajorgrids,
	legend style={at={(0.98,0.9)},legend columns=1, legend cell align=left, align=left, anchor=north east,  draw=black,fill=white},
	label style={font=\scriptsize},
	tick label style={font=\scriptsize},
	legend style={font=\scriptsize}
	]

	\addplot[ybar, bar width=0.25, fill=mmpLightGray, draw=mmpBlack, area legend, line width = 0.5] table[row sep=crcr] {%
        0.125   0.4\\
		0.625   25.6\\
		1.125   5.8\\
		1.625   7.2\\
		2.125   7.3\\
		2.625   8.8\\
		3.125   8.7\\
		3.625   8.1\\
		4.125   7.9\\
		4.625   6.3\\
		5.125   4.5\\
		5.625   3.4\\
		6.125   2\\
		6.625   2\\
		7.125   1.4\\
		7.625   0.4\\
		8.125   0\\
		8.625   0\\
		9.125   0.1\\
		9.625   0.1\\
	};
	\addlegendentry{First $1{,}000$ Cycles}
	
	
	\addplot[ybar, bar width=0.25, fill=mmpDarkGray, draw=mmpBlack, area legend, line width = 0.5] table[row sep=crcr] {%
	0.375 0\\
	0.875 6.3\\
	1.375 7.2\\
	1.875 	7.3\\
	2.375 17.3\\
	2.875 24.8\\
	3.375 18.8\\
	3.875 11.2\\
	4.375 4.4\\
	4.875 2\\
	5.375 0.2\\
	5.875  0.1\\
	6.375 0.2\\
	6.875 0\\
	7.375 0\\
	7.875 0.1\\
	8.375 0.1\\
	8.875 0\\
	9.375 0\\
	9.875 0\\
	};
	\addlegendentry{Last $1{,}000$ Cycles}

	\addplot[mmpRed, line width=2pt, domain=0:30, samples=2] coordinates {(5,0) (5,30)};
	\addlegendentry{$\gls{rDPMAXLim}$}
	
	
\end{axis}

\end{tikzpicture}

%% file: Figures/PlotEpisodeUntrainiert.tex
\begin{tikzpicture}
\setlength{\PlotWidth}{6cm} 
\setlength{\PlotHeight}{1.2cm} 
\setlength{\DeltaHeight}{0.2cm}
\setlength{\DeltaHeightDashed}{0.2cm}

\newcommand{\NumPlot}{1}
\pgfmathsetlength{\Ypos}{-1*\NumPlot*\PlotHeight - 1*\NumPlot*\DeltaHeight}
\begin{axis}[%
	width=\PlotWidth,
	height=\PlotHeight,
	at={(0cm, \Ypos)},
	scale only axis,
	xmin=0,
	xmax=1000,
	xticklabel style={/pgf/number format/fixed},
	xticklabels = {},
	xlabel style={font=\color{white!15!black}},
	xlabel={},
	ymin=1.7,
	ymax=4.5,
	ylabel style={at={(-0.075,0.5)}},
	ylabel={\gls{rpmi} / \si{\bar}},
	axis background/.style={fill=white},
	xmajorgrids,
	ymajorgrids,
	legend style={at={(0.5,1.15)},legend columns=3, legend cell align=left, align=left, anchor=center,  draw=none,fill=none},
	label style={font=\scriptsize},
	tick label style={font=\scriptsize},
	legend style={font=\scriptsize}
	]
	\addplot[ color=mmpDarkBlue,line width=1pt] table {
		x y 
		0 1
		1 1
	};
	\addplot[ color=mmpLightBlue,line width=1pt] table {
		x y 
		0 1
		1 1
	};
	\addplot[ color=mmpRed,line width=1pt] table {
		x y 
		0 1
		1 1
	};

	\addplot[ color=mmpDarkBlue,line width=0.5pt,forget plot] table [x index=0, y index=1, col sep=comma] {Figures/Data/PlotEpisode/DataRLVSR30099_episode622.csv};
	\addplot[ color=mmpLightBlue,line width=0.5pt,forget plot] table [x index=0, y index=2, col sep=comma] {Figures/Data/PlotEpisode/RLVSR_30091MP51.csv};

	\addplot[ color=mmpRed,line width=0.5pt,forget plot] table [x index=0, y index=2, col sep=comma] {Figures/Data/PlotEpisode/DataRLVSR30099_episode622.csv};

	\addlegendentry{RL Agent}
	\addlegendentry{ANN-based}
	\addlegendentry{Setpoint/Limit}
	
\end{axis}

\renewcommand{\NumPlot}{2}
\pgfmathsetlength{\Ypos}{-1*\NumPlot*\PlotHeight - 1*\NumPlot*\DeltaHeight}
\begin{axis}[%
	width=\PlotWidth,
	height=\PlotHeight,
	at={(0cm, \Ypos)},
	scale only axis,
	xmin=0,
	xmax=1000,
	xticklabel style={/pgf/number format/fixed},
	xticklabels = {},
	xlabel style={font=\color{white!15!black}},
	xlabel={},
	ylabel style={at={(-0.075,0.5)}},
	ylabel={\gls{gAlpha50} / \si{\degreeKW}},
	axis background/.style={fill=white},
	xmajorgrids,
	ymajorgrids,
	legend style={at={(0.02,2.6)},legend columns=3, legend cell align=left, align=left, anchor=north west, draw=none},
	label style={font=\scriptsize},
	tick label style={font=\scriptsize},
	]
	\addplot[ color=mmpDarkBlue,line width=0.5pt, forget plot] table [x index=0, y index=3, col sep=comma] {Figures/Data/PlotEpisode/DataRLVSR30099_episode622.csv};
	
	\addplot[ color=mmpLightBlue,line width=0.5pt, forget plot] table [x index=0, y index=3, col sep=comma] {Figures/Data/PlotEpisode/RLVSR_30091MP51.csv};
	label style={font=\scriptsize},
	tick label style={font=\scriptsize},
\end{axis}

\renewcommand{\NumPlot}{3}
\pgfmathsetlength{\Ypos}{-1*\NumPlot*\PlotHeight - 1*\NumPlot*\DeltaHeight}
\begin{axis}[%
	width=\PlotWidth,
	height=\PlotHeight,
	at={(0cm, \Ypos)},
	scale only axis,
	xmin=0,
	xmax=1000,
	ymin=-10,
	ymax=10,
	xticklabel style={/pgf/number format/fixed},
	xticklabels = {},
	xlabel style={font=\color{white!15!black}},
	xlabel={},
	ylabel style={at={(-0.075,0.5)}},
	ylabel={$\Delta$\gls{gAlpha50} / \si{\degreeKW}},
	axis background/.style={fill=white},
	xmajorgrids,
	ymajorgrids,
	legend style={at={(0.02,2.6)},legend columns=2, legend cell align=left, align=left, anchor=north west, draw=none},
	label style={font=\scriptsize},
	tick label style={font=\scriptsize},
	]
	\addplot[ color=mmpDarkBlue,line width=0.5pt, forget plot] table [x index=0, y index=14, col sep=comma] {Figures/Data/PlotEpisode/DataRLVSR30099_episode622.csv};
	
	\addplot[ color=mmpLightBlue,line width=0.5pt, forget plot] table [x index=0, y index=10, col sep=comma] {Figures/Data/PlotEpisode/RLVSR_30091MP51.csv};
		\addplot [color=mmpRed,line width=0.5pt, forget plot]
	table[row sep=crcr]{%
		0	0\\
		1000	0\\};
\end{axis}

\renewcommand{\NumPlot}{4}
\pgfmathsetlength{\Ypos}{-1*\NumPlot*\PlotHeight - 1*\NumPlot*\DeltaHeight}
\begin{axis}[%
	width=\PlotWidth,
	height=\PlotHeight,
	at={(0cm, \Ypos)},
	scale only axis,
	xmin=0,
	xmax=1000,
	xticklabel style={/pgf/number format/fixed},
	xticklabels = {},
	xlabel style={font=\color{white!15!black}},
	xlabel={},
	ylabel style={at={(-0.075,0.5)}},
	ylabel={$\gls{rDPMAX}$ / $\frac{\si{\bar}}{\si{\degreeKW}}$},
	axis background/.style={fill=white},
	xmajorgrids,
	ymajorgrids,
	legend style={at={(0.02,0.05)},legend columns=2, legend cell align=left, align=left, anchor=south west, draw=none},
	label style={font=\scriptsize},
	tick label style={font=\scriptsize},
	]
	\addplot[ color=mmpDarkBlue,line width=0.5pt, forget plot] table [x index=0, y index=4, col sep=comma] {Figures/Data/PlotEpisode/DataRLVSR30099_episode622.csv};
	\addplot[ color=mmpLightBlue,line width=0.5pt, forget plot] table [x index=0, y index=4, col sep=comma] {Figures/Data/PlotEpisode/RLVSR_30091MP51.csv};
	\addplot [color=mmpRed,line width=0.5pt, forget plot]
	table[row sep=crcr]{%
		0	5\\
		10000	5\\};

\end{axis}

\pgfmathsetlength{\Ypos}{-1*\NumPlot*\PlotHeight - 1*(\NumPlot)*\DeltaHeight-\DeltaHeightDashed/2-\DeltaHeight/2}
\pgfmathsetlength{\XEndPos}{\PlotWidth+0.5cm}

\pgfmathsetlength{\Ypos}{-1*\NumPlot/2*\PlotHeight - 1*\NumPlot/2*\DeltaHeight-\DeltaHeightDashed/4-\DeltaHeight/4}
\node[rotate=270] at (\XEndPos-0.05cm,\Ypos) {\scriptsize States/Objectives};

\pgfmathsetlength{\Ypos}{-1*6*\PlotHeight - 1*6*\DeltaHeight-\DeltaHeightDashed+\PlotHeight/2}
\node[rotate=270] at (\XEndPos-0.05cm,\Ypos) {\scriptsize Actions};

\pgfmathsetlength{\YposTop}{-\DeltaHeight}
\pgfmathsetlength{\YposButtom}{-1*4*\PlotHeight - 1*4*\DeltaHeight}
\draw[thick,black,decorate,decoration={brace,amplitude=6pt}]  (\XEndPos-0.45cm,\YposTop)-- (\XEndPos-0.45cm,\YposButtom);

\pgfmathsetlength{\YposTop}{{-1*4*\PlotHeight - 1*5*\DeltaHeight-\DeltaHeightDashed}}
\pgfmathsetlength{\YposButtom}{-1*7*\PlotHeight - 1*7*\DeltaHeight-\DeltaHeightDashed}
\draw[thick,black,decorate,decoration={brace,amplitude=6pt}]  (\XEndPos-0.45cm,\YposTop)-- (\XEndPos-0.45cm,\YposButtom);

\pgfmathsetlength{\Ypos}{-1*\NumPlot*\PlotHeight - 1*(\NumPlot)*\DeltaHeight-\DeltaHeightDashed/2-\DeltaHeight/2}
\pgfmathsetlength{\XEndPos}{\PlotWidth+0.35cm}
\draw[dashed,thick] (-0.75cm,\Ypos) -- (\XEndPos,\Ypos);

\renewcommand{\NumPlot}{5}
\pgfmathsetlength{\Ypos}{-1*\NumPlot*\PlotHeight - 1*\NumPlot*\DeltaHeight-\DeltaHeightDashed}
\begin{axis}[%
	width=\PlotWidth,
	height=\PlotHeight,
	at={(0cm, \Ypos)},
	scale only axis,
	xmin=0,
	xmax=1000,
	xticklabel style={/pgf/number format/fixed},
	xticklabels = {},
	xlabel style={font=\color{white!15!black}},
	xlabel={},
	ylabel style={at={(-0.075,0.5)}},
	ylabel={$\gls{rt}_{\gls{iKr},\gls{iInj}}$ / \si{\milli\second}},
	axis background/.style={fill=white},
	xmajorgrids,
	ymajorgrids,
	label style={font=\scriptsize},
	tick label style={font=\scriptsize},
	]
	\addplot[ color=mmpDarkBlue,line width=0.5pt,forget plot] table [x index=0, y index=5, col sep=comma] {Figures/Data/PlotEpisode/DataRLVSR30099_episode622.csv};
	\addplot[ color=mmpLightBlue,line width=0.5pt, forget plot] table [x index=0, y index=5, col sep=comma] {Figures/Data/PlotEpisode/RLVSR_30091MP51.csv};
	
	\node[circle, draw=mmpRed, fill=none, inner sep=3pt] at (axis cs:198,0.54) {};
	
	\node[circle, draw=mmpRed, fill=none, inner sep=3pt] at (axis cs:225,0.42) {};
	
\end{axis}

\renewcommand{\NumPlot}{6}
\pgfmathsetlength{\Ypos}{-1*\NumPlot*\PlotHeight - 1*\NumPlot*\DeltaHeight-\DeltaHeightDashed}
\begin{axis}[%
	width=\PlotWidth,
	height=\PlotHeight,
	at={(0cm, \Ypos)},
	scale only axis,
	xmin=0,
	xmax=1000,
	xticklabel style={/pgf/number format/fixed},
	xticklabels = {},
	xlabel style={font=\color{white!15!black}},
	xlabel={},
	ylabel style={at={(-0.075,0.5)}},
	ylabel={$\gls{rt}_{\gls{iEth},\gls{iInj}}$ / \si{\milli\second}},
	axis background/.style={fill=white},
	xmajorgrids,
	ymajorgrids,
	label style={font=\scriptsize},
	tick label style={font=\scriptsize},
	]
	\addplot[ color=mmpDarkBlue,line width=0.5pt,forget plot] table [x index=0, y index=9, col sep=comma] {Figures/Data/PlotEpisode/DataRLVSR30099_episode622.csv};
	\addplot[ color=mmpLightBlue,line width=0.5pt, forget plot] table [x index=0, y index=8, col sep=comma] {Figures/Data/PlotEpisode/RLVSR_30091MP51.csv};
\end{axis}

\renewcommand{\NumPlot}{7}
\pgfmathsetlength{\Ypos}{-1*\NumPlot*\PlotHeight - 1*\NumPlot*\DeltaHeight-\DeltaHeightDashed}
\begin{axis}[%
	width=\PlotWidth,
	height=\PlotHeight,
	at={(0cm, \Ypos)},
	scale only axis,
	xmin=0,
	xmax=1000,
	xlabel style={font=\color{white!15!black}},
	xlabel={Cycle / -},
	xtick={200, 400, 600, 800},
	ylabel style={at={(-0.075,0.5)}},
	ylabel={$\gls{gAlpha}_{\gls{iNVO}}$ / \si{\degreeKW}},
	axis background/.style={fill=white},
	xmajorgrids,
	ymajorgrids,
	label style={font=\scriptsize},
	tick label style={font=\scriptsize},
	]
	
	\addplot[ color=mmpDarkBlue,line width=0.5pt,forget plot] table [x index=0, y index=7, col sep=comma]		 {Figures/Data/PlotEpisode/DataRLVSR30099_episode622.csv};
	\addplot[ color=mmpLightBlue,line width=0.5pt, forget plot] table [x index=0, y index=6, col sep=comma] {Figures/Data/PlotEpisode/RLVSR_30091MP51.csv};
\end{axis}

\end{tikzpicture}

%% file: Figures/PlotRewardAdaption.tex
\begin{tikzpicture}
\setlength{\PlotWidth}{6cm} 
\setlength{\PlotHeight}{4cm} 

\newcommand{\NumPlot}{1}
\pgfmathsetlength{\Ypos}{-1*\NumPlot*\PlotHeight - 1*\NumPlot*\DeltaHeight}
\begin{axis}[%
	width=\PlotWidth,
	height=\PlotHeight,
	at={(0cm, \Ypos)},
	scale only axis,
	xmin=-10,
	xmax=130,
	ymin=-4000,
	ymax=0,
	xticklabel style={/pgf/number format/fixed},
	xlabel style={font=\color{white!15!black}},
	xtick={0,20,40,60,80,100,120},
	xlabel={Groups of $1{,}000$ Cycles / -},
	ylabel={$\sum\gls{rRReward}$ / -},
	axis background/.style={fill=white},
	xmajorgrids,
	ymajorgrids,
	ytick={-4000,-3000,-2000,-1000,0},
	yticklabels={$-4{,}000$,$-3{,}000$,$-2{,}000$,$-1{,}000$,$0$},
	legend style={at={(0.9,0.025)},legend columns=1, legend cell align=left, align=left, anchor=south east,  draw=black,fill=white, inner sep=0.5em, , outer sep=0pt},
	label style={font=\scriptsize},
	tick label style={font=\scriptsize},
	legend style={font=\scriptsize}
	]
	\addplot[ color=mmpBlack,line width=1pt,restrict expr to domain={\coordindex}{0:119}] table [x index=0, y index=1, col sep=comma] {Figures/Data/PlotEpisode/RewardVSR1000Groups30104.csv};
	\addlegendentry{Overall}
	
	\addplot[ color=mmpDarkBlue,line width=1pt,restrict expr to domain={\coordindex}{0:119}] table [x index=0, y index=8, col sep=comma] {Figures/Data/PlotEpisode/RewardVSR1000Groups30104.csv};
	\addlegendentry{$\gls{rRReward}_{\gls{rpmiSoll}}$}
	
	\addplot[ color=mmpLightBlue,line width=1pt,restrict expr to domain={\coordindex}{0:119}] table [x index=0, y index=5, col sep=comma] {Figures/Data/PlotEpisode/RewardVSR1000Groups30104.csv};
	\addlegendentry{$\gls{rRReward}_{\Delta\gls{gAlpha50}}$}
	
	\addplot[ color=mmpRed,line width=1pt,restrict expr to domain={\coordindex}{0:119}] table [x index=0, y index=6, col sep=comma] {Figures/Data/PlotEpisode/RewardVSR1000Groups30104.csv};
	\addlegendentry{$\gls{rRReward}_{\gls{rDPMAX}}$}
	
	\addplot[ color=mmpGreen,line width=1pt,restrict expr to domain={\coordindex}{0:119}] table [x index=0, y index=7, col sep=comma] {Figures/Data/PlotEpisode/RewardVSR1000Groups30104.csv};
	\addlegendentry{$\gls{rRReward}_{\gls{geta}}$}
	
	\addplot[ color=mmpOrange,line width=1pt,restrict expr to domain={\coordindex}{0:119}] table [x index=0, y index=10, col sep=comma] {Figures/Data/PlotEpisode/RewardVSR1000Groups30104.csv};
	\addlegendentry{$\gls{rRReward}_{\Delta \gls{rxAnteil}_{\gls{rEEnergie}_{\gls{iEth}}}}$}
	
	\draw[dashed,color=mmpBlack, thick] (axis cs:0, 0) -- (axis cs:0, -4000)node[pos=0.98,xshift=-2mm, anchor=west, rotate=90] {$\gls{gmRL}_{0}$};
	\draw[dashed,color=mmpBlack, thick] (axis cs:60, 0) -- (axis cs:60, -4000)node[pos=0.98, anchor=west,xshift=-2mm, rotate=90] {$\gls{gmRL}_{60}$};
	\draw[dashed,color=mmpBlack, thick] (axis cs:120, 0) -- (axis cs:120, -4000)node[pos=0.98, anchor=west,xshift=2mm, rotate=90] {$\gls{gmRL}_{120}$};

\end{axis}

\end{tikzpicture}

%% file: Figures/PlotEpisodeStrategieadaption.tex
\begin{tikzpicture}
	\setlength{\PlotWidth}{6cm} 
	\setlength{\PlotHeight}{1.2cm} 
	\setlength{\DeltaHeight}{0.2cm}
	\setlength{\DeltaHeightDashed}{0.2cm}
	
\newcommand{\NumPlot}{1}
\pgfmathsetlength{\Ypos}{-1*\NumPlot*\PlotHeight - 1*\NumPlot*\DeltaHeight}
\begin{axis}[%
	width=\PlotWidth,
	height=\PlotHeight,
	at={(0cm, \Ypos)},
	scale only axis,
	xmin=0,
	xmax=1000,
	xticklabel style={/pgf/number format/fixed},
	xticklabels = {},
	xlabel style={font=\color{white!15!black}},
	xlabel={},
	ymin=1.3,
	ymax=4.5,
	ylabel style={at={(-0.075,0.5)}},
	ylabel={\gls{rpmi} / \si{\bar}},
	axis background/.style={fill=white},
	xmajorgrids,
	ymajorgrids,
	legend style={at={(0.5,1.15)},legend columns=4, legend cell align=left, align=left, anchor=center,  draw=none,fill=none},
	label style={font=\scriptsize},
	tick label style={font=\scriptsize},
	legend style={font=\scriptsize}
	]

	\addplot[ color=mmpLightBlue,line width=1pt] table {
		x y 
		0 1
		1 1
	};
		\addplot[ color=mmpDarkBlue,line width=1pt] table {
		x y 
		0 1
		1 1
	};
	\addplot[ color=mmpBlack,line width=1pt] table {
		x y 
		0 1
		1 1
	};
		\addplot[ color=mmpRed,line width=1pt] table {
		x y 
		0 1
		1 1
	};

	\addplot[ color=mmpBlack,line width=0.5pt, forget plot] table [x index=0, y index=1, col sep=comma] {Figures/Data/PlotEpisode/DataRLVSR30104_episode568.csv};
	\addplot[ color=mmpDarkBlue,line width=0.5pt, forget plot] table [x index=0, y index=1, col sep=comma] {Figures/Data/PlotEpisode/DataRLVSR30104_episode497.csv};
	\addplot[ color=mmpLightBlue,line width=0.5pt, forget plot] table [x index=0, y index=1, col sep=comma] {Figures/Data/PlotEpisode/DataRLVSR30104_episode1.csv};
	\addplot[ color=mmpRed,line width=0.5pt, forget plot] table [x index=0, y index=2, col sep=comma] {Figures/Data/PlotEpisode/DataRLVSR30104_episode1.csv};

	\addlegendentry{$\gls{gmRL}_{0}$}
	\addlegendentry{$\gls{gmRL}_{60}$}
	\addlegendentry{$\gls{gmRL}_{120}$}
	\addlegendentry{Setpoint/Limit}
	
\end{axis}

\renewcommand{\NumPlot}{2}
\pgfmathsetlength{\Ypos}{-1*\NumPlot*\PlotHeight - 1*\NumPlot*\DeltaHeight}
\begin{axis}[%
	width=\PlotWidth,
	height=\PlotHeight,
	at={(0cm, \Ypos)},
	scale only axis,
	xmin=0,
	xmax=1000,
	ymin=-10,
	ymax=10,
	xticklabel style={/pgf/number format/fixed},
	xticklabels = {},
	xlabel style={font=\color{white!15!black}},
	xlabel={},
	ylabel style={at={(-0.075,0.5)}},
	ylabel={$\Delta$\gls{gAlpha50} / \si{\degreeKW}},
	axis background/.style={fill=white},
	xmajorgrids,
	ymajorgrids,
	legend style={at={(0.02,2.6)},legend columns=2, legend cell align=left, align=left, anchor=north west, draw=none},
	label style={font=\scriptsize},
	tick label style={font=\scriptsize},
	]
	
	\addplot[ color=mmpBlack,line width=0.5pt, forget plot] table [x index=0, y index=14, col sep=comma] {Figures/Data/PlotEpisode/DataRLVSR30104_episode568.csv};
	\addplot[ color=mmpDarkBlue,line width=0.5pt, forget plot] table [x index=0, y index=14, col sep=comma] {Figures/Data/PlotEpisode/DataRLVSR30104_episode497.csv};
	\addplot[ color=mmpLightBlue,line width=0.5pt, forget plot] table [x index=0, y index=14, col sep=comma] {Figures/Data/PlotEpisode/DataRLVSR30104_episode1.csv};
	
	\addplot [color=mmpRed,line width=0.5pt, forget plot]
	table[row sep=crcr]{%
		0	0\\
		1000	0\\};
\end{axis}

\renewcommand{\NumPlot}{3}
\pgfmathsetlength{\Ypos}{-1*\NumPlot*\PlotHeight - 1*\NumPlot*\DeltaHeight}
\begin{axis}[%
	width=\PlotWidth,
	height=\PlotHeight,
	at={(0cm, \Ypos)},
	scale only axis,
	xmin=0,
	xmax=1000,
	xticklabel style={/pgf/number format/fixed},
	xticklabels = {},
	xlabel style={font=\color{white!15!black}},
	xlabel={},
	ylabel style={at={(-0.075,0.5)}},
	ylabel={$\gls{rDPMAX}$ / $\frac{\si{\bar}}{\si{\degreeKW}}$},
	axis background/.style={fill=white},
	xmajorgrids,
	ymajorgrids,
	legend style={at={(0.02,0.05)},legend columns=2, legend cell align=left, align=left, anchor=south west, draw=none},
	label style={font=\scriptsize},
	tick label style={font=\scriptsize},
	]
	\addplot[ color=mmpBlack,line width=0.5pt, forget plot] table [x index=0, y index=4, col sep=comma] {Figures/Data/PlotEpisode/DataRLVSR30104_episode568.csv};
	\addplot[ color=mmpDarkBlue,line width=0.5pt, forget plot] table [x index=0, y index=4, col sep=comma] {Figures/Data/PlotEpisode/DataRLVSR30104_episode497.csv};
	\addplot[ color=mmpLightBlue,line width=0.5pt, forget plot] table [x index=0, y index=4, col sep=comma] {Figures/Data/PlotEpisode/DataRLVSR30104_episode1.csv};

	\addplot [color=mmpRed,line width=0.5pt, forget plot]
	table[row sep=crcr]{%
		0	5\\
		10000	5\\};

\end{axis}

\renewcommand{\NumPlot}{4}
\pgfmathsetlength{\Ypos}{-1*\NumPlot*\PlotHeight - 1*\NumPlot*\DeltaHeight}
\begin{axis}[%
	width=\PlotWidth,
	height=\PlotHeight,
	at={(0cm, \Ypos)},
	scale only axis,
	xmin=0,
	xmax=1000,
	xticklabel style={/pgf/number format/fixed},
	xticklabels = {},
	xlabel style={font=\color{white!15!black}},
	xlabel={},
	ylabel style={at={(-0.075,0.5)}},
	ylabel={$\gls{rxAnteil}_{\gls{rEEnergie}_{\gls{iEth}}}$ / -},
	axis background/.style={fill=white},
	xmajorgrids,
	ymajorgrids,
	label style={font=\scriptsize},
	tick label style={font=\scriptsize},
	]
	\addplot[ color=mmpBlack,line width=0.5pt, forget plot] table [x index=0, y index=15, col sep=comma] {Figures/Data/PlotEpisode/DataRLVSR30104_episode568.csv};
	\addplot[ color=mmpDarkBlue,line width=0.5pt, forget plot] table [x index=0, y index=15, col sep=comma] {Figures/Data/PlotEpisode/DataRLVSR30104_episode497.csv};
	\addplot[ color=mmpLightBlue,line width=0.5pt,forget plot] table [x index=0, y index=15, col sep=comma] {Figures/Data/PlotEpisode/DataRLVSR30104_episode1.csv};

	\addplot [color=mmpRed,line width=0.5pt, forget plot]
	table[row sep=crcr]{%
	0	0.5\\
	1000	0.5\\};
\end{axis}

\pgfmathsetlength{\Ypos}{-1*\NumPlot*\PlotHeight - 1*(\NumPlot)*\DeltaHeight-\DeltaHeightDashed/2-\DeltaHeight/2}
\pgfmathsetlength{\XEndPos}{\PlotWidth+0.5cm}

\pgfmathsetlength{\Ypos}{-1*\NumPlot/2*\PlotHeight - 1*\NumPlot/2*\DeltaHeight-\DeltaHeightDashed/4-\DeltaHeight/4}
\node[rotate=270] at (\XEndPos-0.05cm,\Ypos) {\scriptsize States/Objectives};

\pgfmathsetlength{\Ypos}{-1*6*\PlotHeight - 1*6*\DeltaHeight-\DeltaHeightDashed+\PlotHeight/2}
\node[rotate=270] at (\XEndPos-0.05cm,\Ypos) {\scriptsize Actions};

\pgfmathsetlength{\YposTop}{-\DeltaHeight}
\pgfmathsetlength{\YposButtom}{-1*4*\PlotHeight - 1*4*\DeltaHeight}
\draw[thick,black,decorate,decoration={brace,amplitude=6pt}]  (\XEndPos-0.45cm,\YposTop)-- (\XEndPos-0.45cm,\YposButtom);

\pgfmathsetlength{\YposTop}{{-1*4*\PlotHeight - 1*5*\DeltaHeight-\DeltaHeightDashed}}
\pgfmathsetlength{\YposButtom}{-1*7*\PlotHeight - 1*7*\DeltaHeight-\DeltaHeightDashed}
\draw[thick,black,decorate,decoration={brace,amplitude=6pt}]  (\XEndPos-0.45cm,\YposTop)-- (\XEndPos-0.45cm,\YposButtom);

\pgfmathsetlength{\Ypos}{-1*\NumPlot*\PlotHeight - 1*(\NumPlot)*\DeltaHeight-\DeltaHeightDashed/2-\DeltaHeight/2}
\pgfmathsetlength{\XEndPos}{\PlotWidth+0.35cm}
\draw[dashed,thick] (-0.75cm,\Ypos) -- (\XEndPos,\Ypos);

\renewcommand{\NumPlot}{5}
\pgfmathsetlength{\Ypos}{-1*\NumPlot*\PlotHeight - 1*\NumPlot*\DeltaHeight-\DeltaHeightDashed}
\begin{axis}[%
	width=\PlotWidth,
	height=\PlotHeight,
	at={(0cm, \Ypos)},
	scale only axis,
	xmin=0,
	xmax=1000,
	xticklabel style={/pgf/number format/fixed},
	xticklabels = {},
	xlabel style={font=\color{white!15!black}},
	xlabel={},
	ylabel style={at={(-0.075,0.5)}},
	ylabel={$\gls{rt}_{\gls{iKr},\gls{iInj}}$ / \si{\milli\second}},
	axis background/.style={fill=white},
	xmajorgrids,
	ymajorgrids,
	label style={font=\scriptsize},
	tick label style={font=\scriptsize},
	]
	\addplot[ color=mmpBlack,line width=0.5pt, forget plot] table [x index=0, y index=5, col sep=comma] {Figures/Data/PlotEpisode/DataRLVSR30104_episode568.csv};
	\addplot[ color=mmpDarkBlue,line width=0.5pt, forget plot] table [x index=0, y index=5, col sep=comma] {Figures/Data/PlotEpisode/DataRLVSR30104_episode497.csv};
	\addplot[ color=mmpLightBlue,line width=0.5pt,forget plot] table [x index=0, y index=5, col sep=comma] {Figures/Data/PlotEpisode/DataRLVSR30104_episode1.csv};
	
\end{axis}

\renewcommand{\NumPlot}{6}
\pgfmathsetlength{\Ypos}{-1*\NumPlot*\PlotHeight - 1*\NumPlot*\DeltaHeight-\DeltaHeightDashed}
\begin{axis}[%
	width=\PlotWidth,
	height=\PlotHeight,
	at={(0cm, \Ypos)},
	scale only axis,
	xmin=0,
	xmax=1000,
	xticklabel style={/pgf/number format/fixed},
	xticklabels = {},
	xlabel style={font=\color{white!15!black}},
	xlabel={},
	ylabel style={at={(-0.075,0.5)}},
	ylabel={$\gls{rt}_{\gls{iEth},\gls{iInj}}$ / \si{\milli\second}},
	axis background/.style={fill=white},
	xmajorgrids,
	ymajorgrids,
	label style={font=\scriptsize},
	tick label style={font=\scriptsize},
	]
	\addplot[ color=mmpBlack,line width=0.5pt, forget plot] table [x index=0, y index=9, col sep=comma] {Figures/Data/PlotEpisode/DataRLVSR30104_episode568.csv};
	\addplot[ color=mmpDarkBlue,line width=0.5pt, forget plot] table [x index=0, y index=9, col sep=comma] {Figures/Data/PlotEpisode/DataRLVSR30104_episode497.csv};
	\addplot[ color=mmpLightBlue,line width=0.5pt,forget plot] table [x index=0, y index=9, col sep=comma] {Figures/Data/PlotEpisode/DataRLVSR30104_episode1.csv};

\end{axis}

\renewcommand{\NumPlot}{7}
\pgfmathsetlength{\Ypos}{-1*\NumPlot*\PlotHeight - 1*\NumPlot*\DeltaHeight-\DeltaHeightDashed}
\begin{axis}[%
	width=\PlotWidth,
	height=\PlotHeight,
	at={(0cm, \Ypos)},
	scale only axis,
	xmin=0,
	xmax=1000,
	xticklabel style={/pgf/number format/fixed},
	xticklabels = {},
	xlabel style={font=\color{white!15!black}},
	xlabel={},
	ylabel style={at={(-0.075,0.5)}},
	ylabel={$\gls{gAlpha}_{\gls{iNVO}}$ / \si{\degreeKW}},
	axis background/.style={fill=white},
	xmajorgrids,
	ymajorgrids,
	label style={font=\scriptsize},
	tick label style={font=\scriptsize},
	]
	\addplot[ color=mmpBlack,line width=0.5pt, forget plot] table [x index=0, y index=7, col sep=comma] {Figures/Data/PlotEpisode/DataRLVSR30104_episode568.csv};
	\addplot[ color=mmpDarkBlue,line width=0.5pt, forget plot] table [x index=0, y index=7, col sep=comma] {Figures/Data/PlotEpisode/DataRLVSR30104_episode497.csv};
	\addplot[ color=mmpLightBlue,line width=0.5pt,forget plot] table [x index=0, y index=7, col sep=comma]		 {Figures/Data/PlotEpisode/DataRLVSR30104_episode1.csv};

\end{axis}

%

\end{tikzpicture}

%% file: Chapters/Conclusion.tex
\section{Conclusion and Outlook}
\label{chap:Conclusion}

In this work, a toolchain was developed to enable the use of RL in safety-critical real-world environments, such as engine testbenches, through application of the DDPG algorithm. To ensure safety, a dynamic measurement algorithm was employed to generate data during load-transient operations, along with a novel algorithm to iteratively determine the stochastic limits of the experimental space. Leveraging these limitations, a safety monitoring function based on the k-nearest neighbor algorithm was implemented, enabling the RL agent to interact with the real-world testbench environment under safety-critical constraints, mitigating risks such as excessive pressure rise rates and misfires.

In an initial feasibility study, the RL agent successfully learned a policy through direct interaction with the testbench environment, achieving an RMSE of $0.1374\,\si{\bar}$ for IMEP, which resulted in control quality comparable to that of ANN-based reference strategies from the literature. The potential of the RL toolchain was especially highlighted by adaptation of the agent's policy into an unexplored region of the experimental space with safety monitoring disabled. Through slow exploration, the agent upheld critical safety constraints while successfully adapting its policy by increasing ethanol use. Our RL methodology thus provides a valuable tool for research and development, enabling, for example, the testing of renewable fuels directly in real-world environments and the adaptation of policies to new boundary conditions or objectives.  Given its adaptability, this method could also be employed in a wide range of other applications with safety-critical environments, such as autonomous vehicles, robotics, or aerospace systems.

A key limitation of our method is its reliance on extensive prior measurements obtained through the dynamic measurement algorithm to parameterize the safety monitoring function, which increases the overall testbench time. Future research could address this by integrating the learning of the safety monitoring function into the RL training process. This would enable the control policy to be learned with even less prior knowledge directly in the real-world environment. However, dynamically learning the safety monitoring function during training poses significant challenges, as it alters the environment. Addressing these challenges in future research could pave the way for even more efficient and adaptable RL applications in safety-critical scenarios.

In conclusion, our safe RL approach represents a significant advancement in bridging the critical gap in applying RL effectively and safely within safety-critical real-world environments. By enabling safety-aware policy adaptations --~even in previously unexplored regions of the experimental space~-- our toolchain establishes a foundation for broader, more reliable RL applications across complex, high-risk scenarios. Additionally, the flexibility of the LExCI toolchain facilitates the seamless transfer of this approach to other safety-critical processes or environments.